\documentclass[10pt]{article}
\usepackage[paperwidth=210mm,paperheight=276mm,left=16mm,right=16mm,top=12mm,bottom=18mm,footskip=9mm]{geometry}
\usepackage[T1]{fontenc}
\usepackage{newtxtext,newtxmath}
\usepackage{amsmath,multicol,graphicx,caption,booktabs,longtable,array,calc}
\usepackage{microtype,xurl,titlesec,fancyhdr,enumitem,needspace}
\usepackage[numbers,sort&compress]{natbib}
\usepackage[hidelinks,hypertexnames=false]{hyperref}
\hypersetup{pdftitle={A Language-Guided Multimodal Foundation Model for Zero-Shot and Multi-Task Brain Signal Analysis},pdfauthor={Mingzhi Chen; Yiyu Gui; Guibo Luo; Yuchao Yang}}
\renewcommand{\normalsize}{\fontsize{10}{13.3}\selectfont}
\normalsize
\titleformat{\section}{\fontsize{11}{13}\selectfont\bfseries}{\thesection\enspace|}{.5em}{}
\titleformat{\subsection}{\normalsize\bfseries}{\thesubsection\enspace|}{.5em}{}
\titleformat{\subsubsection}{\normalsize\itshape}{\thesubsubsection\enspace|}{.5em}{}
\titlespacing*{\section}{0pt}{10pt plus 1pt minus 1pt}{5pt}
\titlespacing*{\subsection}{0pt}{9pt plus 1pt minus 1pt}{4pt}
\titlespacing*{\subsubsection}{0pt}{8pt plus 1pt minus 1pt}{4pt}
\DeclareCaptionFont{paper}{\fontsize{8}{10.5}\selectfont}
\newenvironment{textflow}{\begin{multicols}{2}\raggedcolumns}{\end{multicols}}

\newcommand{\decl}[1]{\par\vspace{8pt}\noindent\textbf{#1}\par\vspace{3pt}\noindent}

\begin{document}
{\centering\fontsize{18}{21.5}\selectfont\bfseries A Language-Guided Multimodal Foundation Model for\\ Zero-Shot and Multi-Task Brain Signal Analysis\par}
\vspace{9pt}
{\centering Mingzhi Chen\textsuperscript{1}\enspace|\enspace Yiyu Gui\textsuperscript{1}\enspace|\enspace Guibo Luo\textsuperscript{1}\enspace|\enspace Yuchao Yang\textsuperscript{1,2,3}\par}
\vspace{6pt}
{\fontsize{8}{10}\selectfont\noindent
\textsuperscript{1}New Cornerstone Science Laboratory, Guangdong Provincial Key Laboratory of In-Memory Computing Chips, School of Electronic and Computer Engineering, Shenzhen Graduate School, Peking University, Shenzhen, China\par
\noindent\textsuperscript{2}Center for Brain Inspired Intelligence, Chinese Institute for Brain Research (CIBR), Beijing, China\par
\noindent\textsuperscript{3}New Cornerstone Science Laboratory, Beijing Advanced Innovation Center for Integrated Circuits, School of Integrated Circuits, Peking University, Beijing, China\par
\vspace{4pt}\noindent\textbf{Correspondence:} Guibo Luo (\href{mailto:luogb@pku.edu.cn}{luogb@pku.edu.cn})\enspace|\enspace Yuchao Yang (\href{mailto:yuchaoyang@pku.edu.cn}{yuchaoyang@pku.edu.cn})\par
\vspace{3pt}\noindent\textbf{Keywords:} brain signal analysis; foundation models; language-signal alignment; zero-shot learning\par
\vspace{3pt}\noindent Published in \emph{Advanced Intelligent Systems} (2026), e70486. \href{https://doi.org/10.1002/aisy.70486}{doi:10.1002/aisy.70486}\par}
\vspace{8pt}\hrule\vspace{5pt}
\noindent\textbf{ABSTRACT}\par\vspace{4pt}
{\fontsize{9.5}{12}\selectfont\noindent
Brain signal analysis is essential for both neuroscience research and clinical diagnostics, yet current approaches face critical limitations. End-to-end models require task-specific retraining and exhibit limited generalization, while pre-trained models lack semantic depth and still depend on extensive fine-tuning. Meanwhile, general-purpose multimodal foundation models, though powerful in other domains, struggle to interpret brain signals due to representational misalignment and lack of domain knowledge. This study introduces a \textbf{M}ultimodal foundation mod\textbf{E}l for zero-sho\textbf{T} and mult\textbf{I}-ta\textbf{S}k brain signal analysis (\textbf{METIS}) through a unified language-signal alignment framework. METIS is pretrained on the largest and most diverse brain-signal corpus to date, comprising over 70 000 h of recordings from more than 11 000 subjects across 20 datasets. In a comprehensive zero-shot evaluation across 12 datasets, METIS outperformed the leading generalist model by over 20.9\% in average accuracy. Remarkably, without any fine-tuning, METIS’s performance matches or exceeds that of supervised, task-specific models. Furthermore, METIS demonstrates exceptional data efficiency and strong generalization, achieving an average AUROC advantage of over 16.0\% in few-shot settings and 15.9\% in cross-dataset transfer. This work establishes a new paradigm for general-purpose brain signal analysis, paving the way for next-generation neurotechnology. The code is available at \url{https://github.com/mingzhi-c/metis-brain-signal-foundation-model}.\par}
\vspace{6pt}\hrule\vspace{6pt}

\begin{textflow}
\section{Introduction}
\par
Brain signal analysis, especially using non-invasive electroencephalography (EEG) and invasive intracranial EEG (iEEG), plays a pivotal role in advancing neuroscience research and improving clinical diagnostics \cite{ref1,ref2,ref3,ref4}. In recent years, deep learning has significantly propelled the field forward, demonstrating performance superior to traditional methods in tasks such as sleep stage classification, seizure detection, and the diagnosis of psychiatric and neurodegenerative disorders \cite{ref5,ref6,ref7,ref8}.
\par
However, the heterogeneity of brain signal data—stemming from differences in electrode setups, acquisition hardware, and participant cohorts—has made it difficult for existing models to generalize across tasks and datasets \cite{ref9,ref10}. Domain-specific models, including both end-to-end \cite{ref11,ref12,ref13,ref14,ref15} and pre-trained architectures \cite{ref16,ref17,ref18,ref19,ref20}, are confined to task-specific training, which leads to high development costs, long iteration cycles, and limited adaptability to real-world demands such as zero-shot classification and multi-task processing. In parallel, general-purpose multimodal foundation models \cite{ref21,ref22,ref23,ref24}, while adept at instruction following and reasoning, lack the necessary domain knowledge to effectively interpret the complex dynamics of brain signals.
\par
Current deep learning-based brain signal analysis methods can be broadly divided into end-to-end models and pre-trained models, both of which face significant challenges. End-to-end models are typically trained on single-task datasets and perform well in constrained settings, but they require retraining for each new task, lack zero-shot generalization, and are sensitive to changes in channel configuration and signal duration. For instance, representative architectures such as EEGNet \cite{ref11} and EEGConformer \cite{ref12} are typically designed and configured for specific datasets and single tasks. Although they perform well within their training domains, they fail to generalize when the task, channel montage, or signal duration differs, necessitating full retraining and limiting their adaptability across datasets and clinical scenarios. Pre-trained models, on the other hand, attempt to enhance feature extraction through large-scale pre-training. Representative examples include BIOT \cite{ref16}, which applies contrastive learning with signal perturbations, and LaBraM \cite{ref17} and CBraMod \cite{ref18}, which employ masked reconstruction \cite{ref25} in the temporal or spectral domain. However, these methods primarily target low-level signal reconstruction or local invariance, while neglecting the high-level semantic information embedded in clinical annotations. Consequently, their representations lack sufficient semantic depth for clinical diagnostic tasks, exhibit poor generalization, and still require task-specific fine-tuning of downstream classifiers, limiting their ability to support zero-shot classification.
\par
Meanwhile, the success of general-purpose multimodal foundation models across various domains offers a new perspective for this field \cite{ref26,ref27,ref28,ref29}. These models are capable of understanding natural-language instructions and performing zero-shot reasoning, allowing them to address complex tasks without requiring task-specific training \cite{ref30,ref31,ref32,ref33}. These advancements raise a central question: can such “general intelligence” help overcome the persistent generalization challenges in brain signal analysis? However, experimental results indicate that directly applying existing general-purpose multimodal foundation models to brain signals leads to significant performance degradation (Figure~\ref{fig:3}b). The root cause lies in the models’ lack of domain-specific knowledge: they have scarcely been exposed to brain signal data during training, and their internal representational spaces are inadequate for capturing the complex spatiotemporal dynamics of brain signals. This representational misalignment often results in severe modality confusion and substantial performance collapse.
\par
Here, we propose a \textbf{M}ultimodal foundation mod\textbf{E}l for zero-sho\textbf{T} and mult\textbf{I}-ta\textbf{S}k brain signal analysis (\textbf{METIS}) through a unified language-signal alignment framework. The model formulates neural state assessment and disease identification as question answering tasks based on brain signals and natural language prompts. To support this paradigm, we constructed the largest and most diverse brain signal instruction corpus to date, comprising over 70 000 h of recordings from more than 11 000 subjects across 20 datasets (Figure~\ref{fig:1}a). METIS’s architecture consists of three key components: a universal signal encoder that maps heterogeneous brain signals into a unified token sequence; a multimodal attention \cite{ref34} mechanism that applies full attention to model the complete context of the signal sequence and uses causal attention for conditional language modeling, thereby bridging signal representations and language generation; and a mixture-of-experts \cite{ref35,ref36,ref37,ref38} (MoE) module, which distinguishes between signal and language tokens to adaptively handle varying signal characteristics, facilitating flexible and specialized computation within a unified framework (Figure~\ref{fig:1}b).
\par
To comprehensively evaluate METIS, we conducted systematic experiments across zero-shot, few-shot, and cross-dataset transfer settings (Figure~\ref{fig:1}c), comparing it against specialized models and general-purpose multimodal foundation models. Experiments covered 17 datasets with varying modalities, acquisition centers, and diverse task types. Across all scenarios, METIS consistently outperformed baseline models. In the zero-shot setting, it not only matched or exceeded the performance of supervised models trained on labeled data, but also outperformed the best general-purpose multimodal foundation model by over 20\% in accuracy. Under few-shot conditions, METIS maintained a clear lead, with an average AUROC more than 16\% above the best pre-trained model. It also delivered comprehensive outperformance in cross-dataset transfer tasks, surpassing the best end-to-end and pretrained models by an average of 19.8\% and 15.9\%, respectively. These results establish METIS as a universal and generalizable foundation model for brain signal analysis, demonstrating its potential to significantly reduce reliance on labeled data and enabling scalable, automated analysis across diverse clinical and research settings.
\par
\section{Results}
\par
\subsection{Zero-Shot Generalization Across Diverse Clinical Tasks}
\par
METIS, a foundation model pretrained on large-scale, heterogeneous brain signal datasets, enables zero-shot brain signal analysis via a unified language-signal alignment framework. Given a brain signal segment and a natural language instruction (e.g., “Which sleep stage does this signal belong to?”), METIS generates answers without any task-specific fine-tuning (Figure~\ref{fig:2}a). This is formulated as a signal-question answering task. Crucially, this capability stems from learning clinically meaningful neural representations, rather than relying on shallow statistical correlations. We examined this in sleep stage classification, where Grad-CAM \cite{ref39} visualizations revealed that METIS identifies the N3 stage \cite{ref40} by attending to its defining clinical hallmark: slow-wave activity. This alignment with expert criteria indicates that METIS learns a semantically grounded and interpretable mapping between neural dynamics and high-level concepts.
\par
Unlike traditional supervised models that require large amounts of labeled data and often overfit to task-specific distributions, METIS is designed to learn task-agnostic, cross-domain representations through large-scale pretraining. This enables plug-and-play generalization to new datasets, guided solely by natural language queries, without any access to training samples from the target domain.
\par
To systematically assess its performance, we evaluated METIS across representative clinical tasks. In sleep stage classification, METIS achieved strong zero-shot AUROC scores on both the ISRUC \cite{ref41} (94.1\%) and Dreams \cite{ref42} (92.9\%) datasets, substantially outperforming competitive baselines trained with 10\% of the labeled data (90.5\% and 76.2\%, respectively). This zero-shot superiority extended to the challenging task of interictal epileptiform discharge detection. On the Mayo \cite{ref43} dataset, its zero-shot performance (AUROC 93.5\%) not only surpassed all 1\%-shot models but also rivaled specialized architectures like SPaRCNet trained with a 10\% data budget. When fine-tuned on the same 10\% subset, METIS further improved to an AUROC of 95.9\%, establishing state-of-the-art performance on this task.
\par

\end{textflow}
 \clearpage\noindent\begin{minipage}{\textwidth}\centering
 \includegraphics[width=134.36mm]{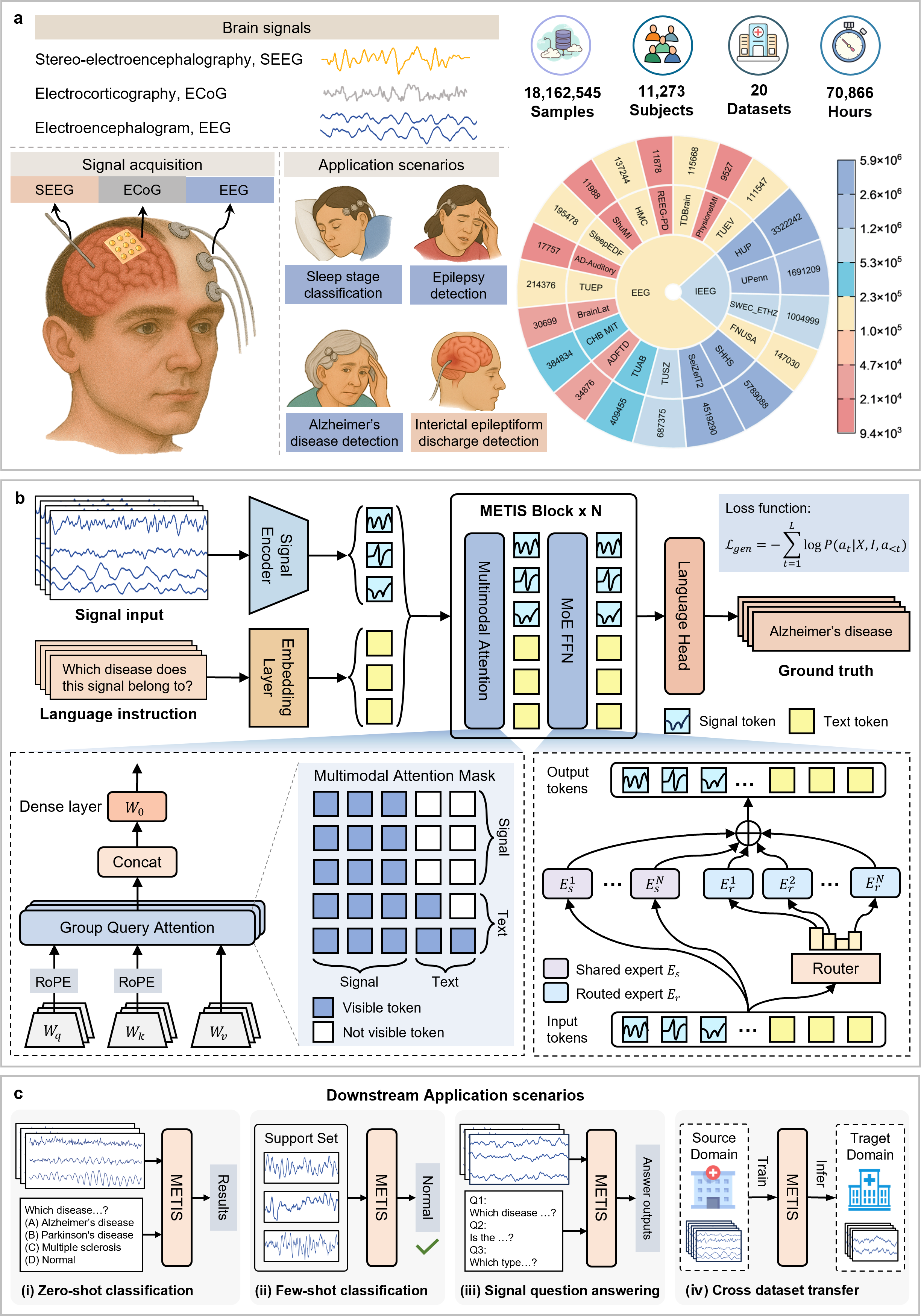}\par
\captionof{figure}{\textbf{Overview of data resources, acquisition modalities, application scenarios, and the METIS model framework.} \textbf{a}, Data resources, modalities, and applications. Brain signals are acquired via multiple modalities, including scalp electroencephalography (scalp EEG), electrocorticography (ECoG), and stereoelectroencephalography (SEEG). Key clinical applications include sleep stage classification, epilepsy detection, Alzheimer's disease detection, and detection of interictal epileptiform discharges (IEDs). The pretraining corpus consists of over 18 million signal segments from 20 datasets, encompassing recordings from more than 11,000 participants and totaling over 70 000 h. The circular chart illustrates the sample distribution across the datasets. \textbf{b}, Core architecture of the METIS model. Brain signals are first transformed into serialized tokens by a universal signal encoder, while natural language prompts are concurrently embedded as text tokens. The model integrates both token streams using a multimodal attention mechanism and stacked METIS blocks. Each block is composed of a Group Query Attention (GQA) layer and a Mixture-of-Experts (MoE) feedforward network. The left inset illustrates the unified multimodal attention mask, which governs the interactions between signal and text tokens. The right inset depicts the MoE layer, where a router directs tokens to specialized experts for targeted processing, while shared experts preserve generalizable representations. \textbf{c}, Evaluation paradigms for METIS. The model supports multiple evaluation paradigms: (i) zero-shot classification on unseen datasets; (ii) few-shot classification using a minimal number of labeled samples; (iii) signal question answering (Signal-QA), where the model processes both a signal and a natural language question to generate a textual answer; and (iv) cross-dataset transfer learning across different recording centers and patient cohorts.}\label{fig:1}

 \end{minipage}\par\clearpage
\begin{textflow}
A critical test for a foundation model is its robustness against major distributional shifts. We challenged METIS with cross-species generalization on the RatEpilepsy \cite{ref44} dataset. Even without prior exposure to rat data, it achieved a zero-shot AUROC of 63.2\%, exceeding untrained baselines by more than 12\% and matching or outperforming supervised models trained with up to 1\% of the target-domain data. When fine-tuned on 10\% of the data, METIS reached an AUROC of 83.6\%, surpassing the best-performing baseline, SPaRCNet, by 6.6\%. This generalization ability also extended to human seizure detection on the SEE \cite{ref45} dataset. When evaluated in the zero-shot setting, METIS achieved an AUROC of 71.5\%—outperforming untrained models by over 17\%, surpassing all 1\%-shot baselines, and exceeding the best-performing model, SPaRCNet, by 6.8\%. With fine-tuning on the same data volume, its advantage further increased to 9.6\%.
\par
We further evaluated METIS’s versatility across a range of challenging psychiatric and neurological disorder detection tasks. For ADHD detection, it achieved zero-shot AUROCs of 72.4\% on ADHD-121 \cite{ref46} and 67.1\% on ADHD-80 \cite{ref47}, exceeding the strongest untrained baselines by 23.7\% and 9.6\%, respectively. When fine-tuned on 1\% or 10\% of labeled data, METIS consistently outperformed all task-specific baselines. Similarly, in schizophrenia detection on the Schizo–Youth \cite{ref48} dataset, METIS achieved a zero-shot AUROC of 72.1\%, surpassing the strongest untrained baseline by 14.4\%, the best 1\%-shot model by 4.5\%, and the best 10\%-shot model by 2.8\%. When fine-tuned on 1\% and 10\% of the labeled data, this margin further expanded to 12.8\% and 18.4\%, respectively. This trend continued for Alzheimer’s disease detection on the ADFSU \cite{ref49} dataset, where METIS achieved a zero-shot AUROC of 74.6\%, outperforming the strongest untrained baseline by 15.3\% and surpassing the best 1\%-shot and 10\%-shot models by 11.6\% and 5.8\%, respectively. On the NMT \cite{ref50} dataset for signal anomaly detection, METIS attained a zero-shot AUROC of 66.2\%, exceeding the best untrained baseline by 16.5\%. While its performance remained slightly below supervised baselines with 1\% and 10\% data, METIS regained a leading advantage when trained under the same data volume.
\par
Aggregated results across all tasks (Figure~\ref{fig:2}b,c) highlight the model’s overall robustness and efficiency. On average, METIS’s zero-shot AUROC (75.5\%) outperformed the 1\%-shot supervised average (73.2\%) and approached the 10\%-shot level (81.6\%). With minimal fine-tuning, METIS’s performance surged to 82.0\% (1\% data) and 87.6\% (10\% data), outperforming all corresponding baselines. Furthermore, the violin plots reveal that METIS not only achieves a higher mean performance but also exhibits a significantly tighter performance distribution, reflecting more stable and reliable generalization across a diverse landscape of clinical tasks and datasets. These results establish METIS as a robust foundation model that consistently delivers strong zero-shot performance and scales effectively with limited supervision, demonstrating the promise of large-scale pretraining for generalizable and interpretable brain signal analysis.
\par
\subsection{Zero-shot signal question answering beyond general-purpose multimodal foundation models}
\par
General-purpose multimodal foundation models are reshaping AI with remarkable performance in language, vision, and reasoning, driving their adoption in visually grounded medical domains such as pathology and radiology \cite{ref26,ref27}. However, it remains an open question whether their capabilities extend to the complex, non-visual data of brain signals. Brain signal analysis presents a unique challenge due to high data variability across clinical settings and the need to ground abstract concepts in temporal patterns. For example, a model must learn to identify specific sleep markers or transient epileptic discharges directly from the signal.
\par
We evaluate METIS against four leading general-purpose multimodal foundation models—ChatGPT, Gemini, Grok, and DeepSeek—on a total of 12 downstream datasets using two zero-shot protocols that share the same prompt and raw input segment. For clarity and conciseness, Figure~\ref{fig:3} presents representative results on four datasets spanning sleep stage classification, Alzheimer’s disease detection, major depressive disorder detection, and interictal epileptiform discharge detection, while comprehensive results on the remaining 8 datasets are provided in Figures~\ref{fig:S18} and~\ref{fig:S19}. The left panels of Figure~\ref{fig:3} report multiple-choice accuracy, whereas the right panels report BERTScore \cite{ref51} for open-ended answers. The open-ended protocol requires the model to generate precise clinical descriptions consistent with the underlying signal evidence.
\par
The multiple-choice results show a consistent trend: METIS leads across all datasets, setting the top performance bar, whereas the strongest general model varies by task and trails behind in average accuracy. METIS achieved an average accuracy of 78.3\% across four datasets, outperforming ChatGPT, the best general model, by 28.6\%. The observed performance gap reflects the varying demands and signal characteristics across tasks. On the sleep stage classification dataset Dreams \cite{ref42}, METIS reached 76.9\% accuracy, establishing a significant 51.7\% lead over ChatGPT, and demonstrating a strong ability to identify complex sleep patterns. This advantage continued on ADFSU \cite{ref49}, an Alzheimer’s disease detection dataset, where METIS (81.7\%) was more than 13\% ahead of ChatGPT (68.4\%). METIS’s superiority was also evident on Mayo \cite{ref43}, a dataset requiring precise detection of interictal discharges in iEEG. Here, its 84.9\% accuracy surpassed that of Grok, the strongest of the generalist models, by over 26\%. Finally, for major depressive disorder detection on MDD \cite{ref52}, METIS’s accuracy of 69.9\% was more than 17\% higher than DeepSeek, the best-performing general model on this dataset.
\par

\end{textflow}
 \clearpage\noindent\begin{minipage}{\textwidth}\centering
 \includegraphics[width=143.65mm]{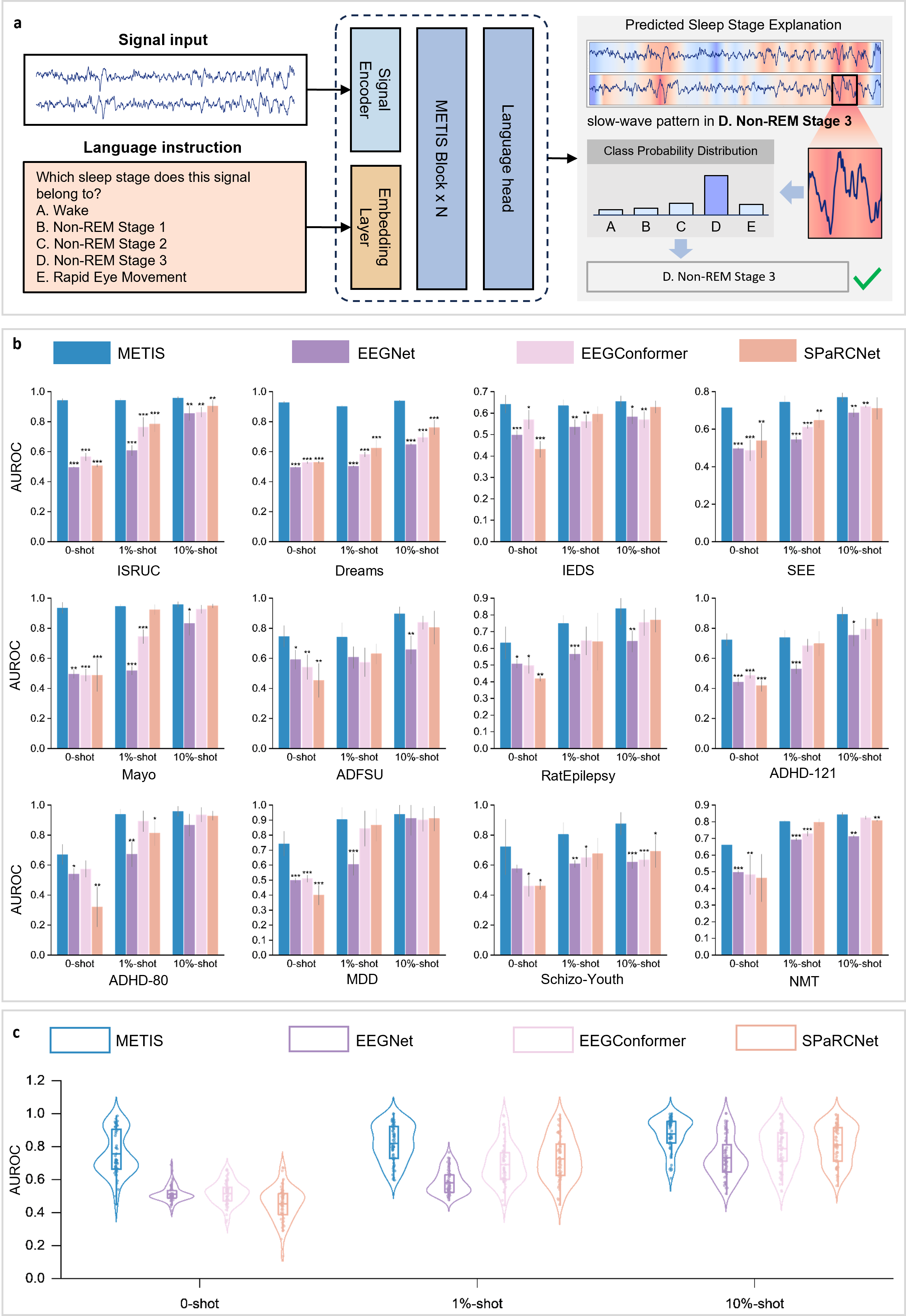}\par
\captionof{figure}{\textbf{Zero-shot performance of METIS in multi-task scenarios.} \textbf{a}, Illustration of zero-shot classification. Using sleep stage classification as an example, the model receives raw brain signals along with a natural language query (``Which sleep stage does this signal belong to?'') and, without any task-specific fine-tuning, outputs the correct answer (``D. Non-REM Stage 3''), while identifying relevant signal patterns (e.g., N3 slow-wave activity). \textbf{b}, Comparison of zero-shot performance with three end-to-end baselines (EEGNet\cite{ref11}, EEGConformer\cite{ref12}, SPaRCNet\cite{ref13}) trained under limited-supervision conditions (1\% and 10\% labeled data), measured by AUROC. Across 12 datasets covering epilepsy detection, sleep stage classification, and psychiatric disorder diagnosis, METIS consistently outperforms task-specific models trained with limited supervision (*p \textless{} 0.05, **p \textless{} 0.01, ***p \textless{} 0.001; two-sided t-test). \textbf{c}, Performance distributions under different data conditions (AUROC, violin plots). The overall mean and median AUROC of METIS under zero-shot settings substantially exceed those of other models trained with 1\% and 10\% of the data, underscoring its cross-task generalization and data efficiency.}\label{fig:2}

 \end{minipage}\par\clearpage
\begin{textflow}
The open-ended setting demonstrates consistent trends, which requires the ability to articulate precise clinical concepts grounded in temporal signal patterns. Averaged across four datasets, METIS achieved a BERTScore of 70.7\%, putting it 17.0\% ahead of the best general baseline, Grok (53.7\%). On the Dreams dataset, METIS demonstrated a substantial improvement over the best-performing baseline, Grok, with a 42.1\% higher BERTScore (88.6\% vs. 46.5\%). It also held a significant 22.6\% advantage on the Mayo dataset (76.7\% vs. 54.1\%). Although the performance gap was narrower on the ADFSU and MDD datasets, METIS remained ahead of the strongest general models.
\par
Qualitative examples in Figure~\ref{fig:3}a reveal the mechanistic failures underlying the quantitative results. A prominent failure mode in several generalist models is a strong thematic bias, where signals from diverse clinical contexts are erroneously mapped to epilepsy. This likely stems from incorrect associations formed during pretraining on vast text corpora, where brain signals are disproportionately discussed in relation to epilepsy, creating a strong but mistaken prior. A more fundamental error is modality confusion, exemplified by Grok, which misinterprets EEG morphology as ECG and consequently provides a cardiovascular diagnosis for a psychiatric condition (major depressive disorder). Furthermore, ChatGPT’s performance highlights a critical imbalance: despite being the strongest generalist model in multiple-choice recognition on Dreams and ADFSU, it ranks last in faithful, open-ended narration for the same tasks. These error patterns collectively point to a disconnect between discriminative pattern matching and true generative understanding. METIS consistently mitigates these failure modes, demonstrating robust performance that stems from its core design of coupling temporal signal structure with semantic concepts through instruction-driven pretraining.
\par
Taken together, these results highlight that the observed performance gap is not simply quantitative, but reflects qualitative differences in model design. Generalist models often exhibit thematic biases, modality confusion, and a disconnect between recognition and generation failures that point to limitations in grounding language to temporal brain signal data. METIS overcomes these challenges through a dedicated architecture that tightly couples semantic interpretation with temporal structure. This demonstrates the necessity of moving beyond general-purpose models and toward specialized, signal-aware architectures to achieve robust performance in brain signal analysis.
\par
\subsection{Few-shot Classification}
\par
A foundation model’s practical utility, especially in data-scarce clinical environments, is critically defined by its ability to generalize from very limited labeled data. This challenge is particularly acute for brain signals, which are highly heterogeneous across individuals and often characterized by low signal-to-noise ratios. To assess this capability, we systematically evaluated METIS on 14 brain signal datasets under 1-shot, 2-shot, 4-shot, and 8-shot settings.
\par
We adopted a linear probing \cite{ref53} protocol to rigorously test the quality of the learned representations. In this setting, all model backbones were frozen, and only a single linear classification layer was fine-tuned. This strategy isolates the contribution of the pretrained features and reflects a practical scenario in which foundation models are used as general-purpose feature extractors without task-specific adaptation. Our evaluation covered nine distinct clinical tasks, such as sleep stage classification, epilepsy detection, psychiatric disorder diagnosis, anesthesia depth estimation, and neurodegenerative disease identification.
\par
As shown in Figure~\ref{fig:4}a, METIS consistently and significantly outperforms all baseline models across the majority of datasets and shot levels. Its advantages are especially evident in high-complexity diagnostic tasks. For example, in sleep stage classification on the Dreams dataset, METIS achieves AUROCs of 95.2\% and 95.7\% under the 2-shot and 8-shot settings. These results surpass the best baseline, BIOT, by 39.8\% and 34.0\%, respectively. Similar gains are observed in interictal epileptiform discharge detection on the Mayo dataset, where METIS scores 93.6\% (2-shot) and 93.7\% (8-shot), outperforming CBraMod by over 22\%.
\par
Even on tasks unseen during pretraining, METIS maintains strong generalization. On the NTUHBIS \cite{ref54} dataset for anesthesia depth monitoring, the model achieves 78.0\% and 83.8\% AUROC under 2-shot and 8-shot settings, respectively. These results confirm its ability to transfer across clinical domains with minimal supervision.
\par
In contrast, existing baselines exhibit pronounced task sensitivity. BIOT performs well on sleep stage classification but struggles on Mayo. CBraMod is relatively stable across datasets, yet fails to reach competitive performance on Dreams. This inconsistency limits their utility in practice, especially in low-resource settings. The ISRUC and ADHD-80 datasets provide further evidence (Figure~\ref{fig:4}a, box plots): METIS exhibits smooth and reliable improvement with more labeled data, while models like LaBraM show minimal gains, with an AUROC increase of only 2\% from 2-shot to 8-shot on ADHD-80.
\par
These trends are confirmed by aggregated performance across all datasets, summarized in Figure~\ref{fig:4}b. METIS achieves average AUROCs of 76.9\%, 77.6\%, 78.9\%, and 81.4\% for the 1-shot, 2-shot, 4-shot, and 8-shot settings. The margins over the best-performing baseline, CBraMod, remain in a narrow band between 16\% and 18\% AUROC, indicating stable advantages across data regimes.
\par

\end{textflow}
 \clearpage\noindent\begin{minipage}{\textwidth}\centering
 \includegraphics[width=142.31mm]{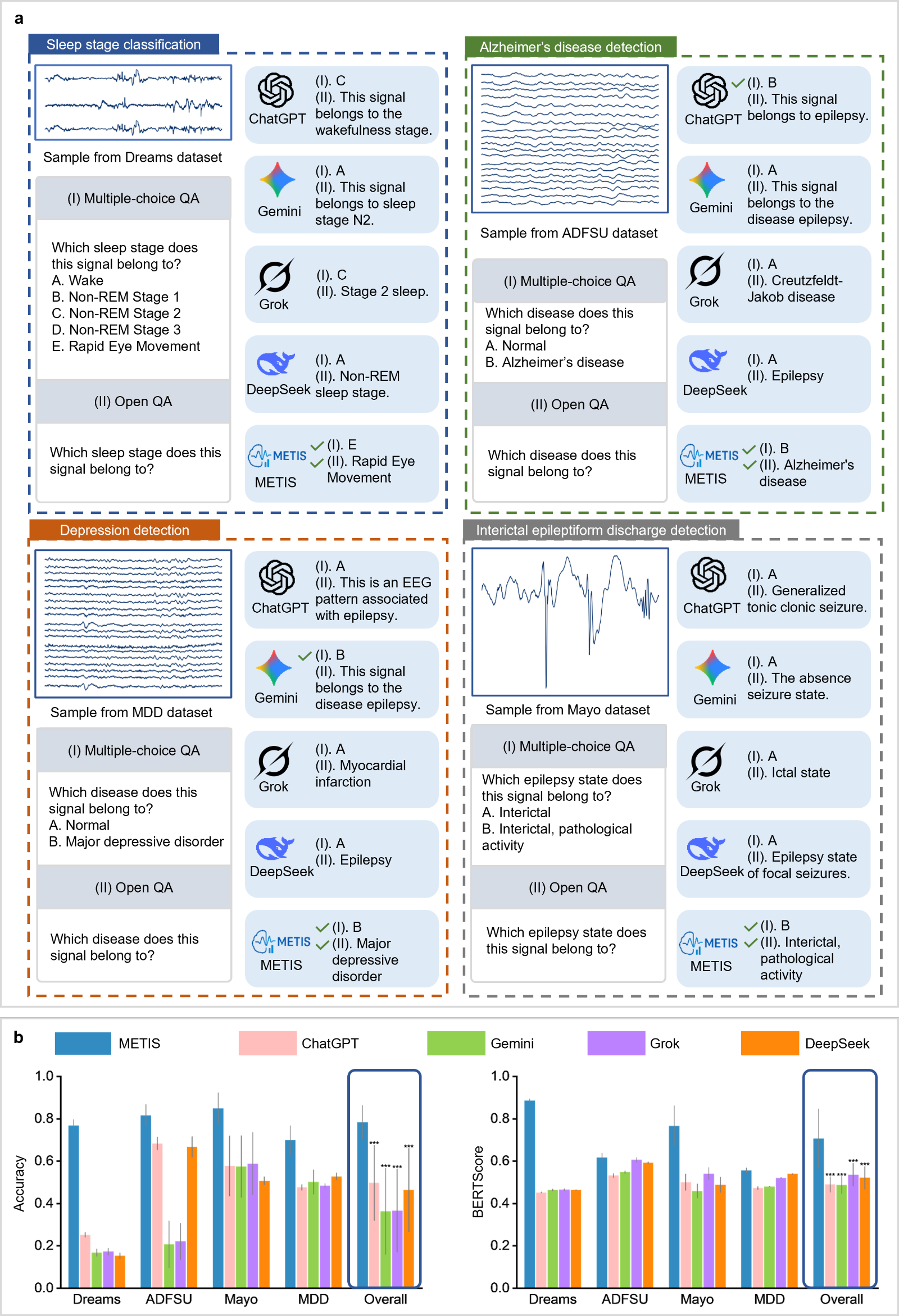}\par
\captionof{figure}{\textbf{Zero shot performance comparison with general-purpose multimodal foundation models.} \textbf{a}, Task examples and model responses on four representative datasets spanning sleep stage classification, Alzheimer's disease detection, major depressive disorder detection, and interictal epileptiform discharge detection. Each example pairs one multiple-choice query and one open ended query for the same signal, with outputs from METIS, ChatGPT\cite{ref21}, Gemini\cite{ref22}, Grok\cite{ref23}, and DeepSeek\cite{ref24}. For readability, only correct answers are marked with a green checkmark; unmarked answers are incorrect. \textbf{b}, Zero-shot results under two protocols. Left panels report multiple-choice accuracy. Right panels report BERTScore\cite{ref51} for open-ended answers. Bars show means over repeated runs, with the rightmost groups giving the overall mean across datasets, where significance markers denote the difference between METIS and the top-performing generalist model (*p \textless{} 0.05, **p \textless{} 0.01, ***p \textless{} 0.001; two-sided t-test). METIS consistently outperforms generalist multimodal models in both protocols.}\label{fig:3}

 \end{minipage}\par\clearpage
\begin{textflow}
Figure~\ref{fig:4}c further illustrates this pattern using normalized radar plots. METIS presents a balanced and broad performance profile across all tasks, indicating strong task-invariant representation learning. In contrast, the baselines show uneven, spiked distributions, suggesting that their success is heavily influenced by task-specific properties rather than generalizable knowledge.
\par
In summary, METIS demonstrates systematic and robust few-shot learning performance across a wide spectrum of clinical brain signal tasks. We attribute this strength to its unified pretraining paradigm, which encourages the alignment of neural signals with semantically meaningful instructions. This results in a shared latent space that supports rapid adaptation with minimal supervision. By consistently outperforming task-specific models with only a handful of labels, METIS offers a scalable and practical solution for real-world brain signal analysis, particularly in data-limited clinical scenarios.
\par
\subsection{Cross Dataset Transfer}
\par
In real-world clinical applications, brain signal models often operate across sites that differ in data acquisition and patient cohorts. Variation in protocols, electrode montages, hardware, sampling rates, and demographics produces substantial distribution shift, which is a major obstacle to practical deployment. To evaluate robustness under these conditions, we conducted 8 cross-dataset transfer experiments spanning four task categories: sleep stage classification, Alzheimer’s disease detection, attention deficit hyperactivity disorder (ADHD) detection, and interictal epileptiform discharge detection. In each experiment, models were trained on one dataset and evaluated directly on another dataset from the same task category without any fine-tuning. Baselines covered both end-to-end models (EEGNet, EEGConformer, SPaRCNet) trained with full supervision on the source data and pretrained models (BIOT, LaBraM, CBraMod) evaluated with frozen encoders and a linear classification head. METIS followed the same linear probing protocol to isolate representational quality.
\par
As shown in Figure~\ref{fig:5}a, METIS consistently outperformed all baselines across 8 transfer settings. The largest gains appeared in sleep stage classification. When transferred from ISRUC to Dreams, METIS achieved an AUROC of 94.0\%, exceeding the best-performing baseline in this setting, EEGConformer (73.1\%), by 20.9\%. In the reverse direction, from Dreams to ISRUC, METIS reached 95.2\%, while the best baseline in this setting, EEGNet, scored 64.0\%, yielding a margin of 31.3\%. These results indicate that METIS retains sleep-relevant structure despite substantial changes in channel configuration between datasets.
\par
Similar patterns held in the remaining tasks. For Alzheimer’s disease detection, METIS attained 72.6\% AUROC when transferred from ADFSU to APAVA \cite{ref56}, outperforming the best-performing baseline in this setting, EEGConformer (57.2\%), by 15.4\%. In the reverse transfer, METIS achieved 80.3\%, surpassing the best baseline, BIOT (64.6\%), by 15.7\%. For ADHD detection, METIS reached 73.9\% from ADHD-80 to ADHD-121 versus 72.6\% for the top baseline in that setting; in the opposite direction it achieved 82.7\%, exceeding BIOT (78.3\%) by 4.4\%. In interictal epileptiform discharge detection, METIS obtained 94.6\% when transferred from IEDS \cite{ref57} to Mayo, outperforming the best-performing baseline (BIOT, 82.8\%) by 11.8\%, and 65.2\% from Mayo to IEDS, ahead of the best baseline (BIOT, 62.1\%) by 3.1\%. Averaged across all transfers, METIS improved AUROC by 15.9\% relative to the best baseline per setting. Together, these results highlight the instability of existing methods under distribution shift, whereas METIS maintains strong and consistent performance.
\par
To examine the representational basis for this robustness, we visualized t-SNE embeddings from the sleep stage classification task (Figure~\ref{fig:5}b). METIS produced compact, well-separated clusters whose spatial arrangement was consistent across ISRUC and Dreams. For example, the relative proximity of wake stage and rapid eye movement stage was preserved across datasets, suggesting that METIS learns task-relevant, domain-invariant features rather than dataset-specific artifacts. In contrast, embeddings from baseline models were often entangled and lacked a stable geometry; LaBraM and CBraMod, in particular, showed overlapping class regions with unclear boundaries, which helps explain their reduced transfer performance.
\par
We also analyzed the behavior of the Mixture-of-Experts (MoE) module in METIS by inspecting expert activation patterns across tasks and datasets (Figure~\ref{fig:5}c). The heatmaps reveal a functional division of roles. Certain experts are preferentially selected for specific task families, such as Expert 19 in sleep stage classification and Expert 73 in interictal epileptiform discharge detection. Other experts, such as Expert 25, are activated broadly across tasks, indicating that they capture common signal attributes like rhythmic structure or artifact suppression. This modular organization allows METIS to combine specialization with generalization: computation can be routed to experts that encode task-specific cues while still leveraging experts that model shared, transferable patterns. Such adaptive routing provides a structural basis for the stable cross-domain performance observed in Figure~\ref{fig:5}a.
\par
In summary, METIS demonstrates reliable domain generalization across four clinical task categories and 8 cross-dataset transfers. The performance advantages are accompanied by interpretable evidence: t-SNE shows topologically consistent, semantically aligned representations across datasets, and MoE analysis shows a complementary mixture of specialized and generalist experts. These findings suggest that coupling strong pretrained representations with modular expert routing is an effective strategy for building brain signal foundation models that remain robust under the distribution shifts encountered in practice.
\par

\end{textflow}
 \clearpage\noindent\begin{minipage}{\textwidth}\centering
 \includegraphics[width=142.05mm]{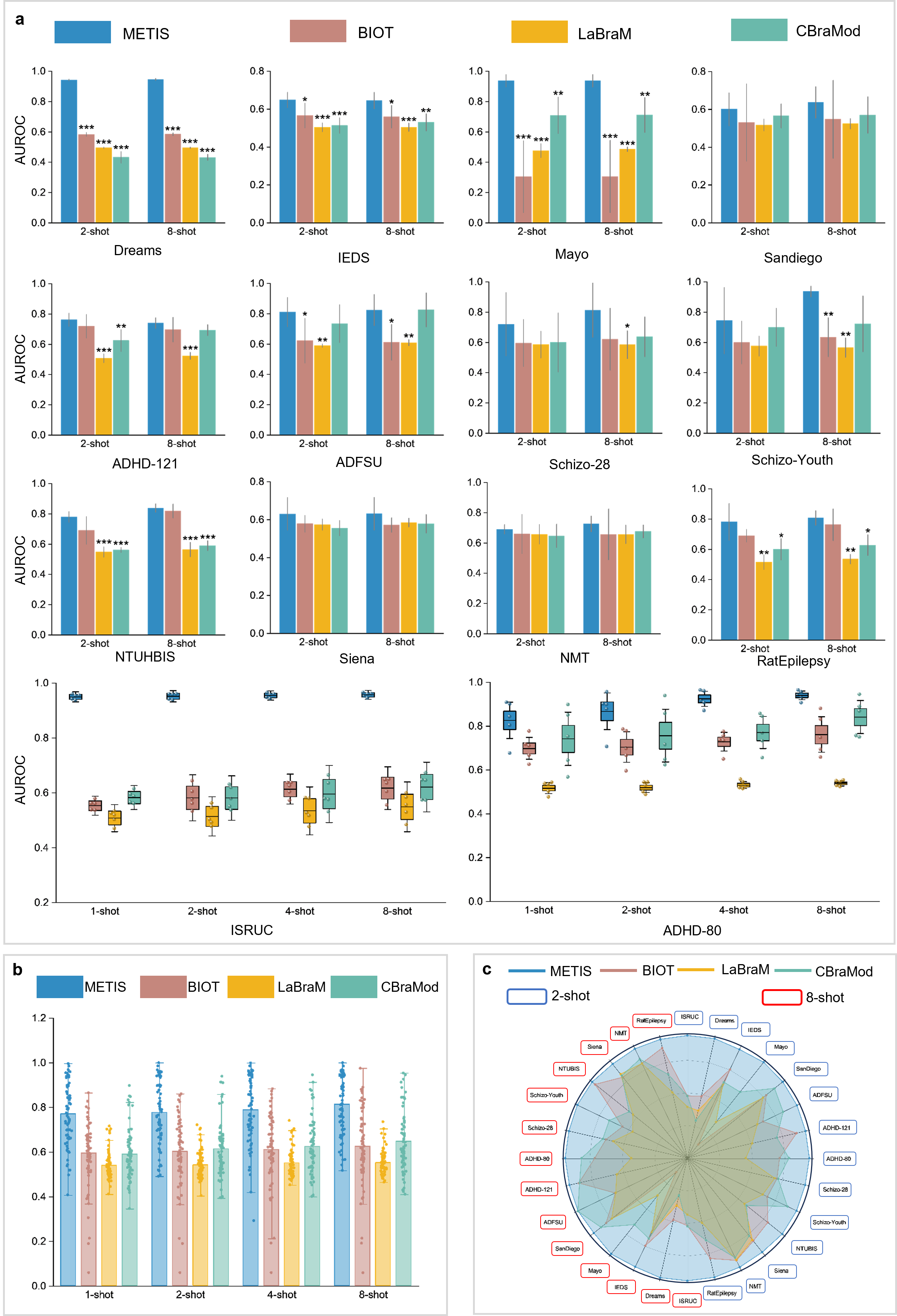}\par
\captionof{figure}{\textbf{Few-shot performance comparison of METIS.} \textbf{a}, Few-shot classification performance (AUROC) across 14 downstream datasets. Bar plots show results under 2-shot and 8-shot conditions, where METIS consistently outperforms existing pretrained baselines (asterisks indicate statistical significance: *p \textless{} 0.05, **p \textless{} 0.01, ***p \textless{} 0.001; two-sided t-test). Box plots below further illustrate performance trends on the ISRUC and ADHD-80 datasets under 1-shot, 2-shot, 4-shot, and 8-shot settings, demonstrating that METIS maintains stable advantages even with extremely limited samples. \textbf{b}, Average performance across all 14 datasets. Bar plots compare mean AUROC under 1-shot, 2-shot, 4-shot, and 8-shot conditions, showing that METIS consistently leads at different data scales and remains robust in low-sample regimes. \textbf{c}, Normalized AUROC radar plots across all tasks. METIS achieves a larger and more uniformly distributed coverage, highlighting its consistency and stability across tasks, whereas baseline models exhibit stronger task dependence.}\label{fig:4}

 \end{minipage}\par\clearpage
\begin{textflow}
\subsection{Ablation Analysis of Key Design Components}
\par
To systematically evaluate the key factors underlying METIS’s performance, we conducted three categories of ablation studies that isolate the effects of model architecture, pretraining modality composition, and data scale. The protocol covered zero-shot inference on 12 downstream datasets and 8-shot linear probing on 14 datasets, with training and evaluation kept consistent across variants to ensure comparability.
\par
First, under structural ablations, removing the mixture-of-experts module (METIS-noMoE) consistently reduced zero-shot performance compared with the full model. Averaged across the 12 datasets, the mean difference between METIS and METIS-noMoE was 10.8\% AUROC, with the largest drops appearing on psychiatric and anomaly-detection benchmarks. On the MDD dataset the reduction was 43.7\% AUROC, and on the Schizo–Youth dataset it was 25.2\%. NMT, ADHD-80, and ADHD-121 datasets also showed clear decreases. These patterns indicate that expert routing creates useful functional subspaces that improve representation quality when class boundaries are subtle or when distributions vary across tasks, which is consistent with the scatter trends visible in Figure~\ref{fig:6}a.
\par
Second, the composition of pretraining modalities had a marked impact on cross-modality transfer. The EEG-only variant maintained reasonable performance on EEG datasets but averaged approximately 50\% AUROC on iEEG tasks such as Mayo and IEDS. The iEEG-only variant performed well on iEEG with an average of 70.4\% AUROC, yet its average on EEG datasets fell to 47.6\%. In contrast, METIS trained jointly on EEG and iEEG was the top performer in 11 of the 12 zero-shot evaluations and was only slightly below the iEEG-only variant on Mayo by 1\% AUROC. Sleep stage classification further illustrates this asymmetry: the iEEG-only variant nearly collapsed on ISRUC and Dreams with AUROCs around 39\%, whereas METIS remained near or above 93\%, in line with the modality-specific clusters seen in Figure~\ref{fig:6}a.
\par
Third, data scale had a systematic effect on both accuracy and stability. As the pretraining corpus increased from 0\% to 100\%, the cross-dataset mean AUROC rose from 49.0\% to 75.5\%. Gains appeared early with only 1\% of the data, where the mean already reached 63.6\%, and continued to accumulate from 10\% to 50\% and from 50\% to full scale. The violin plots in Figure~\ref{fig:6}b reflect this progression as higher central tendencies and tighter spreads at larger scales. The magnitude of improvement was task dependent. Sleep stage classification approached a ceiling quickly, since ISRUC and Dreams were already close to 93\% at 1\% and improved only modestly thereafter. Psychiatric and anomaly-detection benchmarks benefited more strongly from scale, with the Schizo–Youth dataset increasing from about 30\% at 0\% to about 95\% at full scale, and NMT rising from about 33\% to about 66\%. These task-dependent responses explain why scaling reduces variance as well as raises the mean.
\par
Finally, under the 8-shot setting, METIS achieved the highest or joint-highest AUROC on 11 of the 14 datasets. Its overall mean was 81.5\%, compared with 75.4\% for METIS-noMoE, 75.0\% for METIS-EEG, and 76.3\% for METIS-IEEG. Dataset-wise margins were generally small in this low-label regime, but METIS still led by an average of about 5.1\% AUROC over the strongest variant per dataset. Cases where a variant slightly exceeded METIS aligned with that variant’s specialization. METIS-IEEG was marginally higher on Mayo and Schizo-28, and METIS-EEG was marginally higher on ADHD-121. The aggregate bar summaries in Figure~\ref{fig:6}c are consistent with this pattern and show that the full model is the most stable across EEG and iEEG domains.
\par
Taken together, the ablations reveal a coherent picture. The mixture-of-experts architecture supports task-adaptive specialization and helps on domains with weak or overlapping class structure. Joint EEG-iEEG pretraining provides the modality alignment required for transfer in both directions, avoiding the failures observed with unimodal pretraining. Scaling the pretraining corpus improves mean performance and reduces across-task variability, with the largest gains on intrinsically harder or more heterogeneous tasks. These factors explain the robustness of METIS in both zero-shot and few-shot regimes and provide practical guidance for designing transferable foundation models for brain signals.
\par
\section{Discussion}
\par
In this study, we introduce METIS, a novel multimodal language-signal foundation model designed for general-purpose brain signal analysis. Through extensive evaluation across 17 downstream datasets, we demonstrate that METIS surpasses existing end-to-end and pretrained models in zero-shot and multi-task settings, requiring minimal or no additional training. Unlike previous paradigms that rely on single-modality or task-specific models, METIS achieves exceptional cross-task and cross-dataset generalization by aligning heterogeneous brain signals with natural language instructions in a unified space. Specifically, the model shows outstanding performance in critical applications such as sleep stage classification, seizure detection, and psychiatric disorder diagnosis, achieving a zero-shot average accuracy of 70.4\%. This significantly outperforms general-purpose multimodal foundation models (49.5\%), and METIS maintains a consistent advantage in few-shot and cross-dataset transfer tasks.
\par

\end{textflow}
 \clearpage\noindent\begin{minipage}{\textwidth}\centering
 \includegraphics[width=159.00mm]{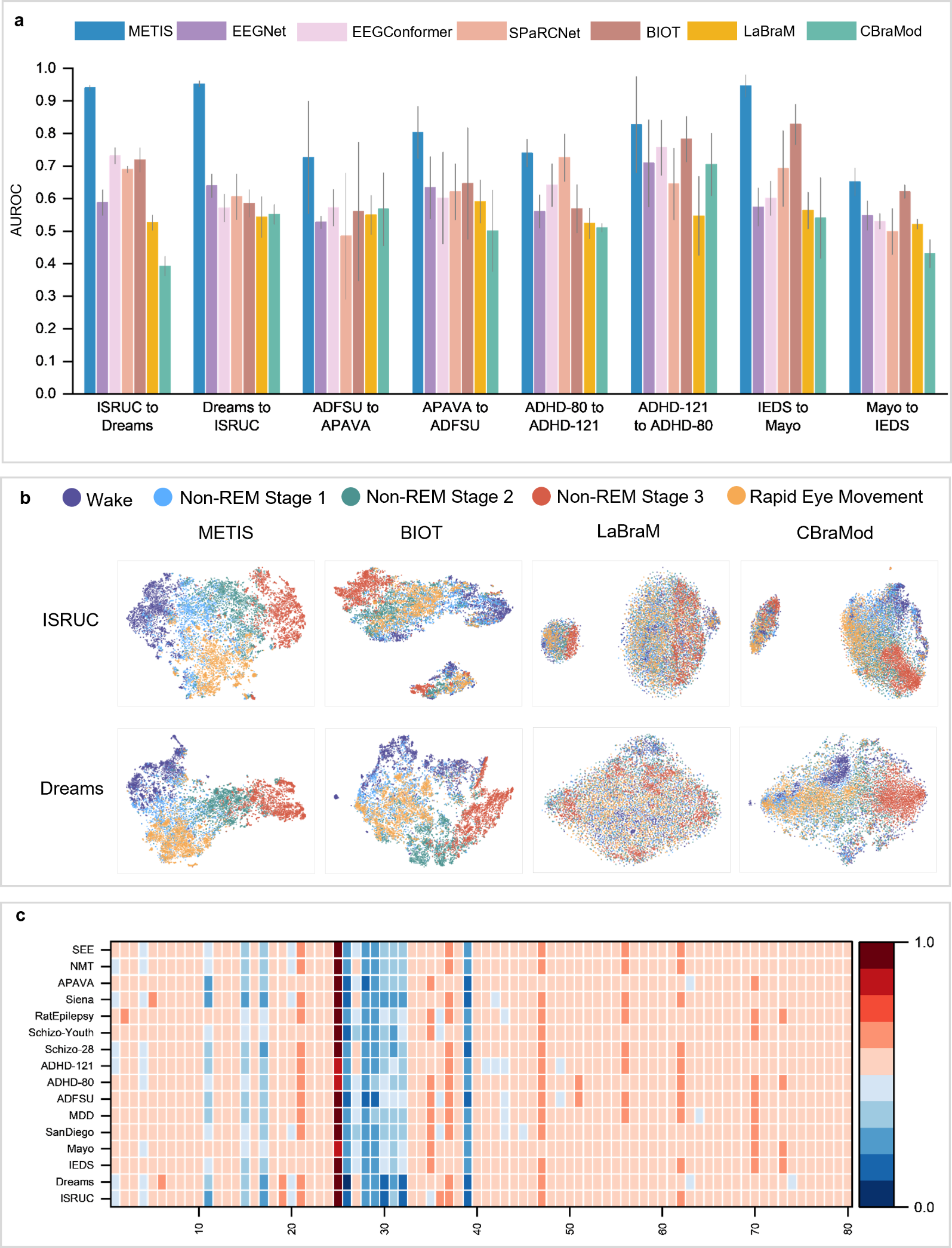}\par
\captionof{figure}{\textbf{Cross dataset transfer, representation geometry, and expert routing in METIS.} \textbf{a}, Cross dataset transfer performance. In each experiment, models are trained on a source dataset and evaluated directly on a different target dataset within the same task category (for example, trained on ISRUC and evaluated on Dreams). Bar plots report AUROC for METIS together with end-to-end baselines (EEGNet, EEGConformer, SPaRCNet) and pretrained baselines (BIOT, LaBraM, CBraMod) across 8 transfer settings. \textbf{b}, Feature visualization for sleep stage classification on ISRUC and Dreams. Two-dimensional t-SNE\cite{ref55} projections are shown for METIS and representative pretrained models; points are colored by sleep stage. The plots illustrate inter-class separability and intra-class compactness, as well as the consistency of relative class layouts across datasets. \textbf{c}, Expert activation heatmap for the METIS mixture-of-experts module. Each row corresponds to a downstream task and each column to an expert. Colors encode routing probability from low (blue) to high (red). The patterns reveal task-dependent expert utilization together with experts that are active across tasks, indicating a balance of specialization and sharing.}\label{fig:5}

 \end{minipage}\par\clearpage
\begin{textflow}
The performance breakthrough of METIS stems from its language-guided causal modeling paradigm combined with large-scale pretraining, which overcomes the fundamental limitations of the two current mainstream approaches. End-to-end models are constrained by task-specific training, resulting in poor generalization. In contrast, the representations learned by unsupervised pretrained models often lack high-level semantic information, which limits their ability to perform zero-shot classification. Meanwhile, general-purpose multimodal foundation models often exhibit modality confusion and degraded performance owing to their lack of domain-specific representations for brain signals. METIS addresses these challenges through an instruction-driven signal– language alignment mechanism that maps brain signal representations to clinically meaningful semantic concepts. This framework enables METIS to leverage the largest and most diverse brain signal corpus to date for pretraining, learning generalizable representations with semantic depth.
\par
The heterogeneity of brain signal data poses a core challenge for training a general brain signal analysis model. Our approach provides an effective solution to this problem by constructing a unified language-signal joint space. METIS maps heterogeneous brain signals from various sources into a shared representation space through its unified signal encoder and employs a MoE module to enable adaptive computation, thereby effectively mitigating the issue of distribution shift. This training paradigm offers a general blueprint for integrating different types of physiological signals and can be extended in the future to the analysis of other time-series data such as electrocardiograms (ECG) and electromyograms (EMG). It even holds the potential to incorporate structured data like genomics to build a multimodal foundation model covering a broader range of clinical scenarios.
\par
The introduction of METIS marks a paradigm shift in brain signal analysis from “specialized tools” to “general-purpose assistants”. Traditionally, each task required customized modeling, leading to redundant processes and high costs. METIS’s zero-shot capabilities fundamentally transform this model: clinical researchers can directly analyze unseen data using natural language instructions, significantly lowering the technical barrier and reducing model deployment time from months to hours. For example, METIS’s average zero-shot performance in this study not only surpassed supervised models trained with 1\% labeled data but also approached those trained with 10\%, indicating strong potential for reducing annotation costs in practical applications.
\par
Although the zero-shot and few-shot classification results of METIS are encouraging, several key issues must be addressed on its path toward broader applicability. While we utilized the largest dataset to date, the data are primarily sourced from existing studies, and there remains room for improvement in covering diverse global populations, rare disease cohorts, and varied acquisition hardware. Future work will require evaluation in broader, multi-center prospective cohorts to further validate robustness and generalizability across clinical settings.
\par
Looking ahead, the principles established by METIS offer a scalable framework for next-generation brain-signal foundation models. We envision future work advancing along three key trajectories. First, a systematic investigation of scaling laws is essential to understand the interplay between model size, data diversity, and emergent capabilities. Expanding pretraining data to the million-hour scale and exploring Mixture-of-Experts architectures with trillions of parameters will unlock unprecedented potential for decoding complex neural states and analyzing rare diseases. Second, advancing these models toward generative and controllable paradigms will enable not only classification but also the simulation of neural activity, thereby suggesting novel intervention strategies. Third, developing lightweight versions via knowledge distillation \cite{ref58} is crucial for deployment on low-power edge devices, such as wearable neural interfaces. This, combined with methods for efficient personalization, will pave the way for truly personalized neurology. Sustained efforts along these directions will transform METIS from a powerful research prototype into an indispensable platform that bridges general AI with the nuanced demands of computational neuroscience and clinical medicine.
\par
\section{Experimental Section}
\par
\subsection{Instruction-Driven Pretraining: Overview and Objective}
\par
We formulate the pretraining of METIS as an instruction-driven generation problem, where the model is required to produce natural language responses grounded in raw EEG or iEEG signals under guidance of a textual prompt. This formulation not only enables a unified sequence-to-sequence modeling paradigm across modalities, but also provides a scalable pretraining strategy that leverages diverse biomedical tasks in a generative way.
\par
As shown in Figure~\ref{fig:1}, the model receives a time-series brain-signal segment along with a natural language instruction (e.g., “Which disease does this signal belong to?”). The raw signal is first transformed into a sequence of spectro-temporal tokens by a universal signal encoder. The instruction is processed by the same model pipeline: it is first tokenized with the Qwen 2.5 \cite{ref59} tokenizer into discrete indices and then passed through the model’s internal embedding layer to form instruction tokens. These two streams of tokens are concatenated into a single unified sequence, which is processed by a decoder-only Transformer equipped with a hybrid attention mask: signal tokens attend bidirectionally, while instruction and answer tokens follow causal masking.
\par

\end{textflow}
 \clearpage\noindent\begin{minipage}{\textwidth}\centering
 \includegraphics[width=167.03mm]{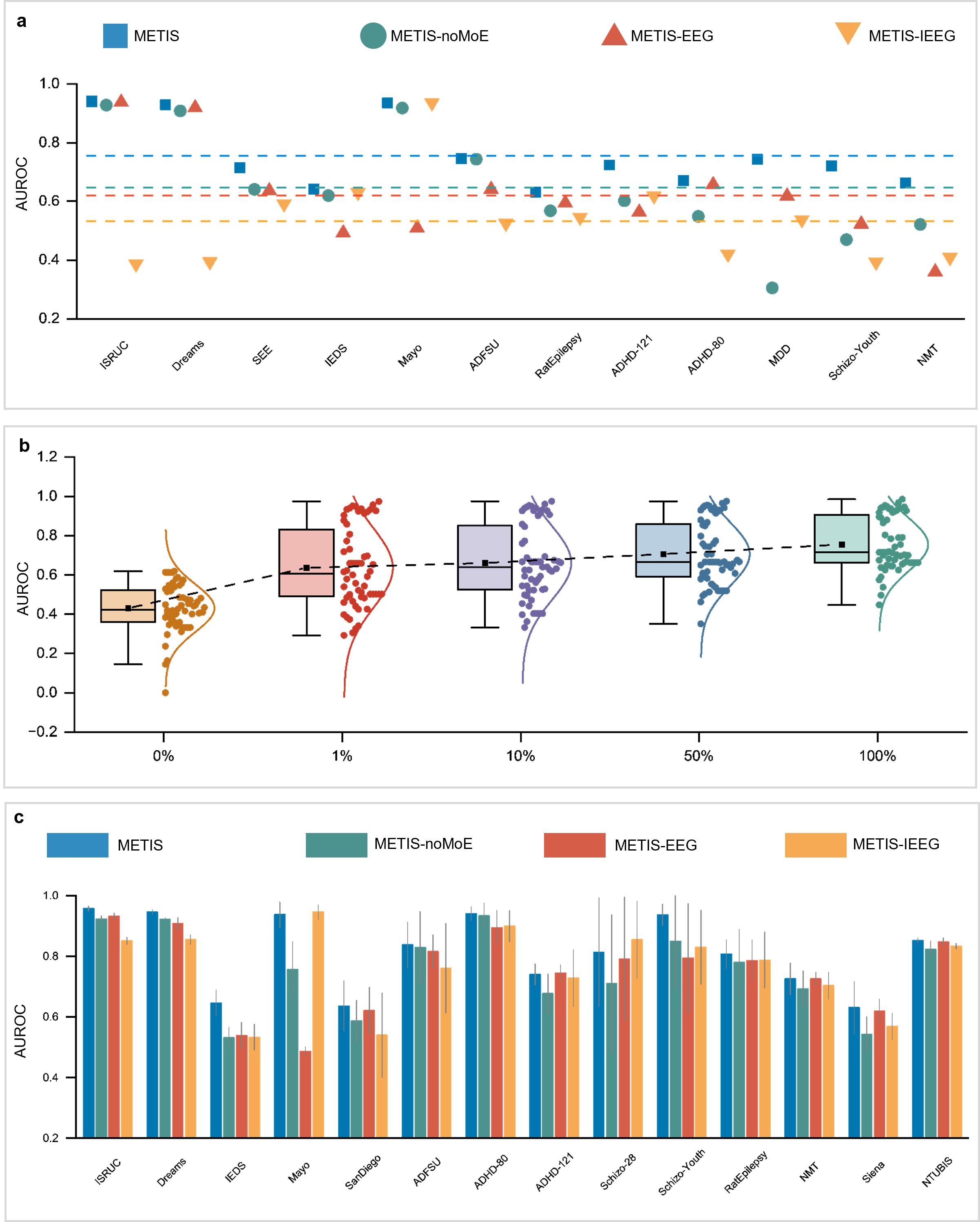}\par
\captionof{figure}{\textbf{Ablation studies on key components of METIS.} \textbf{a}, Zero-shot performance of different model variants across 12 downstream tasks. Scatter plots compare AUROC of the full METIS model with three ablated variants: without the MoE module (METIS-noMoE), pretrained only on EEG (METIS-EEG), and pretrained only on iEEG (METIS-IEEG). Colored dashed lines indicate the average performance of each model across all tasks. \textbf{b}, Impact of pretraining data scale on zero-shot performance. AUROC distributions across 12 tasks are shown for models trained with different proportions of pretraining data (0\%, 1\%, 10\%, 50\%, 100\%). Results illustrate the performance trend from random initialization (0\%) to increasing scales of pretraining. Black dashed lines connect the mean performance at each data scale. \textbf{c}, Few-shot (8-shot) performance of different model variants across 14 downstream tasks. Bar plots compare mean AUROC of the full METIS model and its ablated variants under the 8-shot setting. Error bars denote standard deviations from five-fold cross-validation.}\label{fig:6}

 \end{minipage}\par\clearpage
\begin{textflow}
During pretraining, METIS is optimized in a signal-conditioned autoregressive manner, learning to generate task-aware answer tokens grounded in both neural dynamics and textual instructions. Formally, let \(X\) denote the input brain signal segment, \(I\) the instruction prompt, and \(A = \{ a_{1},a_{2},...,a_{L}\}\ \)the target answer sequence. The model learns a conditional distribution over answer tokens:
\par
\[
p(A\mid X,I)=\prod_{t=1}^{L}p(a_t\mid X,I,a_{<t})
\]
\par
All signal and instruction tokens are fully visible to the answer tokens via a causal attention mechanism, while answer tokens attend only to previous answer tokens through a causal mask. The learning objective is defined as the negative log-likelihood:
\par
\[
\mathcal{L}_{\mathrm{gen}}=-\sum_{t=1}^{L}\log P(a_t\mid X,I,a_{<t})
\]
\par
Unlike vision-language models that align static visual patterns, METIS operates on temporally dynamic neural signals, requiring the model to infer latent neurophysiological states and map them to semantic clinical concepts during generation. This formulation enables METIS to integrate temporal neural representations and linguistic reasoning within a unified autoregressive framework.
\par
Through this signal-conditioned generation objective, METIS learns to associate structured neural dynamics with high-level clinical semantics, supporting zero-shot reasoning across heterogeneous brain signal tasks without task-specific supervision.
\par
In addition to this generation loss, we incorporate an auxiliary sparsity-aware loss \(\mathcal{L}_{balance}\) to encourage expert load balancing in the Mixture-of-Experts (MoE) module. The final training objective is:
\par
\[
\mathcal{L}_{\mathrm{total}}=\mathcal{L}_{\mathrm{gen}}+\lambda\mathcal{L}_{\mathrm{balance}}
\]
\par
Where \(\lambda = 0.001\) is a hyperparameter chosen to balance semantic learning and expert diversity during training.
\par
\subsection{METIS Architecture}
\par
METIS adopts a unified decoder-only Transformer architecture tailored for multimodal reasoning over brain signals and natural-language instructions. Its core design reformulates EEG and iEEG analysis as instruction-driven generation, processing concatenated sequences of raw signals, task prompts, and target answers. The architecture is composed of three synergistic components: a universal signal encoder that transforms heterogeneous recordings into structured representations; a multimodal Transformer backbone integrating Group Query Attention (GQA) \cite{ref60} and a MoE module for efficient, adaptive computation; and modality-aware mechanisms that handle structural and positional distinctions across signal and language tokens.
\par
\subsubsection{Universal Signal Encoder}
The universal signal encoder converts raw EEG and iEEG recordings, which may vary in channel count, sampling frequency, and recording duration, into a tokenized format suitable for integration with textual instructions (Figure~\ref{fig:7}). Each input channel is first Z-score normalized to reduce baseline drift and amplitude variability. To account for the nonstationary nature of brain signals, we apply a Short-Time Fourier Transform (STFT), converting each time-domain waveform \(x_{c}(t)\) into a log-scaled spectrogram \(S_{c}(f,t)\):
\par
\[
S_c(f,t)=\log\!\left(1+\left|\operatorname{STFT}(x_c(t))\right|\right)
\]
\par
Each spectrogram is then processed by a convolutional projection module that transforms the 2D time-frequency representation into a sequence of feature tokens per channel, each with a fixed embedding dimension \(\text{D}\). The resulting token sequence preserves the spectral-temporal structure and serves as the intermediate representation for attention-based modeling. To enable interaction across channels, a multi-head self-attention (MHSA)\cite{ref34} mechanism is applied:
\par
\[
z^{\prime}_{c,t}=\operatorname{MHSA}(z_{1,t},z_{2,t},\ldots,z_{C,t})
\]
\par
The original and attended representations are fused via residual addition and global average pooling across channels:
\par
\[
Z_S=\frac{1}{C}\sum_{c=1}^{C}(z_{c,t}+z^{\prime}_{c,t})
\]
\par
This token sequence \(Z_{S}\) is concatenated with embedded instruction tokens \(Z_{T}\) to form the full input:
\par
\[
Z=[Z_S;Z_T]
\]
\par
This design supports joint modeling of neural dynamics and semantic prompts in a unified generative interface.
\par

\end{textflow}
\clearpage
\noindent\begin{minipage}{\textwidth}\centering\includegraphics[width=134.02mm]{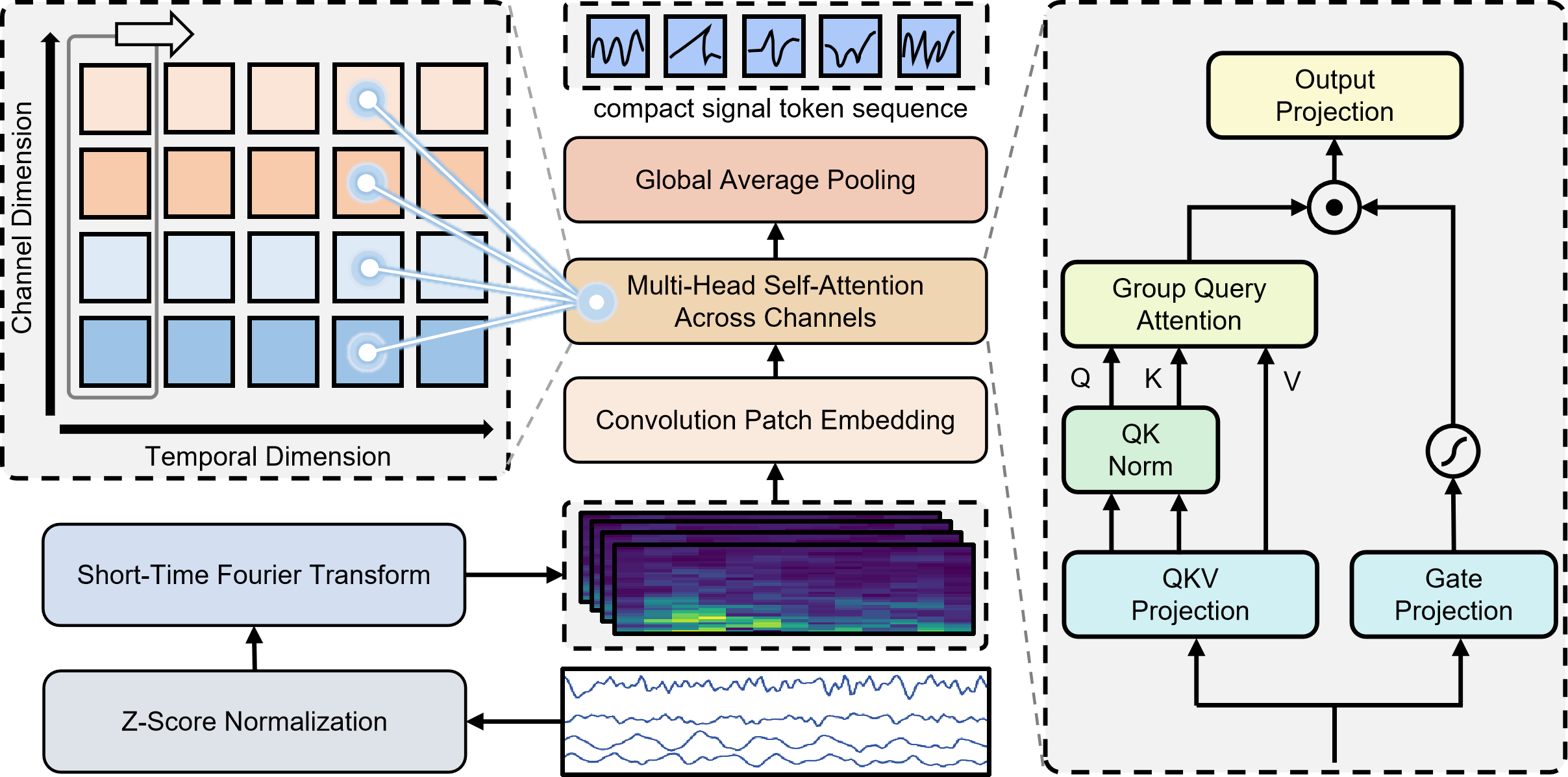}\par
\captionof{figure}{\textbf{Universal signal encoder architecture.} Raw multi-channel brain signals are first Z-score normalized and transformed into log-scaled spectrograms using STFT. A convolutional patch-embedding module converts each time-frequency map into per-channel token sequences. Cross-channel interactions are modeled using multi-head self-attention with QK-Norm for stability and Group Query Attention for parameter efficiency. Residual fusion and global average pooling yield a compact signal-token sequence that serves as a unified representation.}\label{fig:7}
\end{minipage}\par\vspace{7pt}
\begin{textflow}
\subsubsection{Transformer Backbone With Group Query Attention and Mixture-of-Experts}
The concatenated sequence is processed by a stack of Transformer decoder blocks. Each block consists of Group Query Attention (GQA) and a Mixture-of-Experts (MoE) feedforward network. GQA decouples the number of query heads \(h_{q}\) and key-value heads \(h_{k}\), enabling more flexible capacity allocation. For an input tensor \(Z \in \mathbb{R}^{L \times D}\), the projected queries, keys, and values are computed as:
\par
\[
Q=ZW_Q,\quad K=ZW_K,\quad V=ZW_V
\]
\par
The key and value tensors are repeated to match the number of query heads, followed by scaled dot-product attention:
\par
\[
\operatorname{Attention}(Q,K^{\prime},V^{\prime})=\operatorname{SoftMax}\!\left(\frac{QK^{\prime T}}{\sqrt{d_h}}\right)V^{\prime}
\]
\par
To increase model capacity without linearly increasing the parameter count, we adopt a token-level sparse Mixture-of-Experts (MoE) architecture in most feed-forward layers. For each token representation \(x\in \mathbb{R}^{D}\), a router computes gating logits \(g(x) = W_{g}x\), where \(W_{g}\in \mathbb{R}^{\text{N}_{\text{e}} \times D}\) and \(N_{e}\) denotes the number of experts. A learnable bias vector \(b\in \mathbb{R}^{\text{N}_{\text{e}}}\) is added to modulate expert utilization and encourage balanced routing, yielding \(\widetilde{g}(x) = g(x) + b\). The routing probabilities are then obtained via:
\par
\[
p(x)=\operatorname{softmax}(\widetilde{g}(x))
\]
\par
For computational efficiency, only the top-k experts with the highest gate probabilities are activated. Let \(\mathcal{T}_{k}(x)\) denote the index set of the top-k experts, and define a routing mask \(m_{j}(x) \in \{ 0,1\}\). The MoE output is given by:
\par
\[
y(x)=\sum_{j=1}^{N_e}m_j(x)\cdot p_j(x)\cdot E_j(x)+E_s(x)
\]
\par
Where \(E_{j}( \cdot )\) represents the j-th expert network and \(E_{s}( \cdot )\) is a shared expert capturing task-agnostic patterns.
\par
To enforce balanced expert utilization, we introduce an auxiliary load-balancing loss:
\par
\[
\mathcal{L}_{\mathrm{balance}}=\sum_{e=1}^{N_e}p_e\cdot q_e
\]
\par
Here \(p_{e}\) is the fraction of tokens routed to expert \(e\), and \(q_{e}\) is the average gate probability for that expert.
\par
\subsubsection{Multimodal Attention With Hybrid Masking}
We couple brain signal tokens and language tokens using a hybrid attention mask that controls directional dependencies within and across modalities. Let \(N_{s}\) and \(N_{t}\) be the numbers of signal and text tokens. The mask \(M \in \mathbb{R}^{(N_{s} + N_{t}) \times (N_{s} + N_{t})}\) is defined so that signal to signal is bidirectional, text to text is causal, and text to signal is fully visible:
\par
\[
M_{ij}=\begin{cases}0,&i,j\leq N_s,\\0,&i>N_s\text{ and }j\leq i,\\-\infty,&\text{otherwise},\end{cases}
\]
\par
The mask is injected additively into scaled dot-product attention:
\par
\[
\operatorname{Attention}(Q,K^{\prime},V^{\prime})=\operatorname{SoftMax}\!\left(\frac{QK^{\prime T}}{\sqrt{d_h}}+M\right)V^{\prime}
\]
\par
which preserves full bidirectional interactions among signal tokens, enforces causal decoding for text tokens, and grants text tokens access to the entire signal sequence.
\par
\subsubsection{Modality-Aware Rotary Position Encoding}
We use rotary position embeddings\cite{ref61} to inject relative positional information into queries and keys so that attention depends on token distances rather than absolute indices. This is suited to variable-length signal sequences and stabilizes autoregressive decoding when text is appended after the signal. Given a token vector \(x_{m}\) at position \(m\), \(RoPE\) applies a blockwise rotation to queries and keys:
\par
\[
\begin{aligned}\operatorname{RoPE}(x_m,m)={}&x_m\odot\cos(m\theta)\\&+\mathcal{R}(x_m)\odot\sin(m\theta),\end{aligned}
\]
\par
where \(\odot\) denotes elementwise multiplication and \(\mathcal{R}( \cdot )\) swaps the two halves of the vector with a sign flip on the second half. The angular frequencies are defined as \(\theta_{i} = \Theta^{\frac{- 2i}{D}}\), where \(D\) is the feature dimension. We assign modality-specific base frequencies: \(\Theta_{signal} = 10^{4}\) for signal tokens and \(\Theta_{text} = 10^{5}\) for text tokens. Using two bases encodes modality identity directly in the rotational phase and introduces a controlled frequency jump at the signal-text boundary, which discourages spurious adjacency between the last signal token and the first text token, reduces cross-modal attention leakage, and improves alignment of generated language with neural evidence under the hybrid mask.
\par
\par\smallskip\noindent\begin{minipage}{\linewidth}
\captionof{table}{Pretraining datasets and corpus statistics.}\label{tab:1}
\centering\fontsize{8}{10}\selectfont\setlength{\tabcolsep}{2pt}
\begin{tabular}{@{}lcrrr@{}}\toprule
Dataset & \shortstack{Data\\type} & Subjects & Samples & \shortstack{Duration\\(h)}\\\midrule
HUP\cite{ref62} & iEEG & 57 & 3,322,242 & 2768.54\\
UPenn\cite{ref63} & iEEG & 12 & 1,691,209 & 469.78\\
SWEC\_ETHZ\cite{ref64} & iEEG & 16 & 1,004,999 & 837.50\\
FNUSA\cite{ref65} & iEEG & 12 & 147,030 & 122.53\\
SHHS\cite{ref66} & EEG & 6441 & 5,789,088 & 48242.40\\
SeiZelT2\cite{ref67} & EEG & 125 & 4,519,290 & 12553.58\\
TUSZ\cite{ref68} & EEG & 675 & 687,375 & 1909.38\\
TUAB\cite{ref69} & EEG & 2329 & 409,455 & 1137.38\\
CHB MIT\cite{ref70} & EEG & 23 & 384,834 & 1068.98\\
TUEP\cite{ref71} & EEG & 179 & 214,376 & 595.49\\
SleepEDF\cite{ref72} & EEG & 78 & 195,478 & 1628.98\\
HaaglandenSleep\cite{ref73} & EEG & 151 & 137,244 & 1143.70\\
TDBrain\cite{ref74} & EEG & 911 & 115,668 & 64.26\\
TUEV\cite{ref75} & EEG & 370 & 111,547 & 154.93\\
ADFTD\cite{ref76} & EEG & 88 & 34,876 & 19.38\\
BrainLat\cite{ref77} & EEG & 135 & 30,699 & 17.06\\
AD-Auditory\cite{ref78} & EEG & 35 & 17,757 & 9.87\\
ShuMI\cite{ref79} & EEG & 25 & 11,988 & 13.32\\
REEG-PD\cite{ref80} & EEG & 149 & 11,878 & 6.60\\
PhysionetMI\cite{ref81} & EEG & 109 & 9,527 & 7.94\\
\bottomrule\end{tabular}\end{minipage}\par\smallskip

\subsection{Pretraining Corpus Curation and Instruction-Answer Construction}
\par
To train METIS with instruction-driven supervision at scale, we curated a large corpus from 20 EEG and iEEG datasets that span sleep stage classification, epilepsy and interictal epileptiform discharge detection, neurodegenerative and movement disorders, attention-deficit or hyperactivity disorder, mood and psychotic disorders, signal anomaly detection, and motor imagery. The datasets differ in channel configurations, sampling frequencies, subject counts, and annotation conventions. We standardized preprocessing per dataset while respecting its native acquisition and labeling protocols. Signals were minimally cleaned according to source recommendations, and each training instance was paired with a standardized natural-language instruction and a canonical answer string to build instruction-answer pairs for autoregressive pretraining.
\par
We use two QA formats that share the same label sets. In the open ended format the model receives a fixed instruction and generates one canonical answer string drawn from a discrete set. In the multiple-choice format the same candidate set is embedded in the prompt as lettered options and the model selects the corresponding letter. Each dataset is paired with a single instruction sentence and a fixed answer set. No other template families are used. Table~\ref{tab:S1} lists the instruction and answer set for every dataset. For TDBrain the full label inventory is provided in Table~\ref{tab:S2}, and each composite label is treated as a single canonical class string.
\par
\subsection{Evaluation Protocols}
\par
\subsubsection{Zero-Shot Classification}
METIS is uniquely capable of performing zero-shot inference without any task-specific fine-tuning. To evaluate this ability, we reformulate classification problems as multiple-choice question answering tasks. Each instance is presented to the model as a combination of a brain signal segment and a textual instruction prompt. The model processes the concatenated signal and instruction tokens and autoregressively predicts the next token distribution over the vocabulary. As illustrated in Figure~\ref{fig:2}a, we do not decode the full output sequence; instead, we directly extract the logits corresponding to the candidate answer tokens.
\par
Let \(\mathcal{V}\) denote the vocabulary, and let \(\mathcal{C} = \{ c_{1},c_{2},...,c_{K}\} \subset \mathcal{V}\) be the set of candidate answer tokens (e.g., \(\mathcal{C} = \{ A,B,C,D,E\}\) for a five-way classification task). Given input sequence \(X = \{ x_{1},x_{2},...,x_{T}\}\) comprising the signal tokens and the instruction tokens, the model predicts a distribution \(P( \cdot |X) \in \mathbb{R}^{\left| \mathcal{V} \right|}\) over the vocabulary for the next token. We select the logits \(z_{c_{1}},z_{c_{2}},...,z_{c_{K}}\) corresponding to the candidate answers and compute a softmax over this subset:
\par
\[
P(c_i\mid X)=\frac{\exp(z_{c_i})}{\sum_{j=1}^{K}\exp(z_{c_j})},\quad i=1,2,\ldots,K
\]
\par
The predicted label is given by the candidate token with the highest probability. This targeted decoding approach avoids irrelevant tokens and enables efficient classification by leveraging the model’s language-semantic alignment. This zero-shot framework is applied uniformly across all downstream classification tasks, allowing direct comparison of generalization ability across domains.
\par
\subsubsection{Few-Shot Classification}
To assess the label efficiency and adaptability of METIS under limited supervision, we evaluate few-shot classification performance across a range of low-data regimes. Specifically, we consider four settings with \(K = 1,2,4,8\) labeled examples per class.
\par
All few-shot experiments are conducted exclusively on pretrained models, including METIS. We adopt a linear probing protocol, where the backbone encoder remains frozen, and only a task-specific linear classification head is trained using the few-shot training samples. This evaluation strategy focuses on the generalization capacity of the pretrained representations under minimal supervision, allowing us to assess the label efficiency and transferability of learned features across diverse downstream tasks.
\par
To ensure consistency across datasets, we follow two evaluation strategies depending on the presence of predefined splits. For datasets with official train/test splits, we randomly sample \(K\) examples per class from the training set using five independent random seeds, and report the mean and standard deviation of performance on the held-out test set. For datasets without such splits, we perform 5-fold subject-wise cross-validation, where four folds are used for few-shot training and the remaining fold for evaluation, reporting the averaged performance across folds. In all cases, no subjects or recordings are shared across the training, validation, and test partitions; for datasets with official splits, we strictly follow the official partition protocols.
\par
This evaluation framework allows us to compare models in a label-constrained regime while accounting for inter-subject variability and dataset-specific constraints.
\par
\subsubsection{Cross-Dataset Transfer}
To evaluate the generalization ability of each model across different datasets, we design cross-dataset transfer experiments where the model is trained on one dataset (the source) and evaluated on a distinct dataset (the target). Specifically, all available data from the source dataset are used for training, and the target dataset is used exclusively for testing.
\par
We distinguish two cases based on whether the target dataset comes with predefined train/test splits. If no such split is provided, we perform 5-fold subject-wise cross-validation on the target dataset. In each fold, one partition is used as the test set while the remaining four serve as validation-only references (since training occurs solely on the source data). The final results are reported as the average and standard deviation across all five folds. If the target dataset includes predefined splits, we use the official test set directly and conduct five independent transfer experiments using different random seeds, again reporting the mean and standard deviation.
\par
For models trained from scratch, all parameters are fine-tuned on the source dataset before testing on the target. In contrast, all pretrained models, including METIS, are evaluated using a linear probing protocol. In this setting, the backbone encoder remains frozen during training on the source data, and only a task-specific linear classifier is trained. This approach isolates the quality of the learned representations and offers a more rigorous assessment of cross-domain generalization.
\par
\subsection{Downstream Datasets}
\par
To evaluate generalization across heterogeneous clinical contexts, we used 17 datasets spanning sleep stage classification, epilepsy and interictal epileptiform discharge detection, neurodegenerative and movement disorders, ADHD, mood and psychotic disorders, signal anomaly detection, and anesthesia depth monitoring. The datasets vary widely in channel configuration, sampling frequency, cohort size, and segmentation protocol, creating a rigorous test bed for robustness.
\par
\textbf{ISRUC}\cite{ref41}\textbf{:} This sleep stage classification dataset contains overnight scalp EEG from 100 participants sampled at 200 Hz. We followed a two-channel montage (C3-M2 and C4-M1) and segmented recordings into 30-second epochs labeled into the five standard sleep stages, yielding 73,883 labeled segments.
\par
\textbf{Dreams}\cite{ref42}\textbf{:} This sleep stage classification dataset contains overnight EEG from 20 participants sampled at 200 Hz. We used a three-channel montage (FP1-A2, FP2-A1, CZ-A1) and segmented recordings into 30-second epochs labeled into the five standard sleep stages, yielding 20,242 labeled segments.
\par
\textbf{SEE}\cite{ref45}\textbf{:} This epilepsy dataset offers single-channel EEG from 30 subjects at 173.61 Hz with a fixed author-defined split. Signals were cut into 4-second windows and labeled as seizure or non-seizure, giving 4,000 segments in total.
\par
\textbf{Siena}\cite{ref82}\textbf{:} Multichannel scalp EEG from 14 subjects at 256 Hz is annotated into four peri-ictal classes (pre-ictal, ictal, post-ictal, inter-ictal). Using 19 channels and 10-second windows, preprocessing yielded 50,749 segments for evaluating seizure-related dynamics.
\par
\textbf{RatEpilepsy}\cite{ref44}\textbf{:} This intracranial EEG dataset comprises three-channel recordings at 6,000 Hz from five rats. Signals were segmented into 3-second windows labeled as seizure or interictal, yielding 2,530 samples and enabling a cross-species assessment of generalization.
\par
\textbf{IEDS}\cite{ref57}\textbf{:} An iEEG dataset focused on interictal epileptiform discharges, IEDS includes recordings from 25 subjects at 1,000 Hz. After 3-second segmentation, windows were labeled as normal activity or IED, producing 59,615 samples.
\par
\textbf{Mayo}\cite{ref43}\textbf{:} Intracranial EEG from 15 subjects at 500 Hz originally includes four labels. We retained physiological and pathological activity and removed artifacts and power-line noise to form a binary IED detection task. With 3-second windows, the dataset contains 71,957 labeled segments.
\par
\textbf{ADFSU}\cite{ref49}\textbf{:} This Alzheimer's disease dataset comprises nineteen-channel EEG at 128 Hz from 92 participants (80 patients and 12 controls). We created 2-second segments labeled as Alzheimer's disease or control, for a total of 736 samples.
\par
\textbf{APAVA}\cite{ref56}\textbf{:} A complementary Alzheimer's dataset with sixteen-channel EEG at 256 Hz from 23 participants (12 patients and 11 controls). Using 2-second windows, preprocessing yielded 2,652 labeled segments.
\par
\textbf{SanDiego}\cite{ref83}\textbf{:} For Parkinson's disease detection, SanDiego provides thirty-two-channel EEG at 512 Hz from 31 participants (15 patients and 16 controls). We used 2-second segments to obtain 4,521 samples for binary classification.
\par
\textbf{ADHD-121}\cite{ref46}\textbf{:} Nineteen-channel EEG at 128 Hz from 121 children supports ADHD detection. With 5-second windows and binary labels (ADHD or control), the dataset contains 3,322 segments.
\par
\textbf{ADHD-80}\cite{ref47}\textbf{:} This companion ADHD dataset includes two-channel EEG at 256 Hz from 80 participants (38 ADHD, 42 controls). Using 5-second segments, we obtained 5,056 labeled windows.
\par
\textbf{MDD}\cite{ref52}\textbf{:} For major depressive disorder, this dataset contains nineteen-channel EEG at 256 Hz from 61 participants (33 MDD, 28 controls). Segmentation into 5-second windows produced 7,600 labeled samples.
\par
\textbf{Schizo-Youth}\cite{ref48}\textbf{:} Sixteen-channel EEG at 128 Hz from 84 participants (45 with schizophrenia, 39 controls) was segmented into 10-second windows, yielding 504 samples for schizophrenia detection.
\par
\textbf{Schizo-28}\cite{ref84}\textbf{:} Nineteen-channel EEG at 250 Hz from 28 participants evenly split between patients and controls. With 10-second segments, this dataset provides 2,878 samples.
\par
\textbf{NMT}\cite{ref50}\textbf{:} A large-scale EEG anomaly detection benchmark with sixteen channels at 200 Hz. Following the official split and using 10-second windows, we formed 175,091 segments labeled as normal or abnormal to capture broad anomaly patterns.
\par
\textbf{NTUHBIS}\cite{ref54}\textbf{:} This anesthesia depth monitoring dataset contains single-channel EEG at 128 Hz from 23 surgical patients. Labels are derived from the bispectral index and grouped into four levels (deep hypnotic state, general anesthesia, moderate sedation, awake/light sedation). We followed the dataset's segmentation protocol to obtain 5,324 labeled samples.
\par
Across all datasets, we adhered to the original label definitions and harmonized nomenclature where needed. For instruction-driven evaluation, dataset-specific class taxonomies were mapped to consistent prompts while preserving the semantics of each task. This collection spans substantial variability in acquisition protocol, electrode montage, sampling rate, species, and cohort composition, providing a comprehensive basis for assessing task generalization and robustness under distribution shift.
\par
\subsection{Evaluation Metrics}
\par
The primary outcome metric is the area under the receiver operating characteristic curve (AUROC). AUROC is threshold free and less sensitive to class imbalance than accuracy, which is appropriate given the range of class prevalences across the tasks evaluated.
\par
\subsubsection{Binary Tasks}
For two class problems we compute the standard AUROC from the model scores of the positive class. The ROC curve is obtained by varying a threshold on these scores, and the AUROC equals the probability that a randomly chosen positive sample is ranked higher than a randomly chosen negative sample.
\par
\subsubsection{Multiclass Tasks}
For problems with K\textgreater2 classes we report the macro-averaged one-versus-one AUROC. Concretely, we compute AUROC for every unordered pair of classes and then take the unweighted average across all pairs:
\par
\[
\operatorname{AUROC}_{\mathrm{macro\text{-}ovo}}=\frac{2}{K(K-1)}\sum_{1\leq i<j\leq K}\operatorname{AUROC}(i\text{ vs }j)
\]
\par
\subsubsection{Multiple-Choice Question Answering}
For the multiple-choice question answering protocol, each instance provides a brain signal segment and a prompt that enumerates a finite set of candidate answers. The number of candidates varies by dataset and task. For METIS, we obtain a probability for each candidate using the targeted logits described in Section 4.4 and select the candidate with the highest probability. For general-purpose multimodal foundation models, we process input signals into 224\ensuremath{\times}224 images, a widely-used input size in vision-language research\cite{ref30,ref33,ref85}. This image-based input, in contrast to high-dimensional raw signal points, provides a more condensed and feature-rich representation, which we found to be more effective in preliminary studies for capturing the signal's inherent characteristics. We then constrain and parse outputs using a two-stage rule aligned with our implementation. We first request a JSON object validated against a schema whose single field is an uppercase letter drawn from the set of valid option letters. If schema based decoding is unavailable or fails, we fall back to plain text generation and extract the first valid uppercase letter from the allowed set present in the output. Predictions that do not yield a valid option letter are treated as incorrect. Accuracy is the proportion of instances for which the mapped prediction equals the ground truth label. Results are reported per dataset together with an overall mean across datasets.
\par
\subsubsection{Open Ended Question Answering}
In the open ended protocol, the model generates a short textual answer that should match a canonical reference string for the same instance. Answer quality is quantified with BERTScore, which measures token level semantic similarity using contextual embeddings. Precision is the average over candidate tokens of the maximum similarity to any reference token. Recall is the average over reference tokens of the maximum similarity to any candidate token. We report the F1 combination of precision and recall as the main score. Following common practice, reference and candidate strings are lowercased and stripped of punctuation prior to scoring, and we use the default English configuration provided by the official BERTScore package.
\par
\subsection{Implementation Details}
\par
All data preparation, pretraining, and downstream evaluations were conducted on the same Linux server to ensure consistency and reproducibility. The machine uses an Intel Xeon Gold 6342 CPU and a single NVIDIA A100 80 GB GPU. The software stack includes Python 3.11.7, PyTorch 2.0.1 with CUDA 12.2, NumPy 1.26.3, SciPy 1.10.1, Transformers 4.41.2, einops 0.8.0, and pyhealth 1.1.4.
\par
\subsubsection{Pretraining}
Training employed mixed precision in PyTorch. We used the AdamW optimizer with \ensuremath{\beta}=(0.9,0.999), weight decay 0.1, and a base learning rate of 2e-4. The schedule included a warm-up phase covering 10\% of the total pretraining steps. The per-step batch size was 256 with gradient accumulation of 20 steps, giving an effective batch size of 5,120 samples per optimizer update.
\par
\subsubsection{Downstream Evaluation}
Few-shot experiments and cross-dataset transfer were conducted in full precision to ensure numerical stability. All experiments were trained for 200 epochs, allowing sufficient optimization steps for the models to converge. We used AdamW with \ensuremath{\beta}=(0.9,0.999), weight decay 0.1, and a base learning rate of 2e-4, and a batch size of 128. Zero-shot evaluation used the pretrained checkpoint without any fine-tuning, applying the same inference stack across datasets.
\par

\begingroup\fontsize{8.5}{11}\selectfont
\decl{Author Contributions}
This paper was primarily written by Mingzhi Chen and Yiyu Gui. Mingzhi Chen conceived the study, designed the methodology, and conducted all experiments. Mingzhi Chen and Yiyu Gui performed dataset curation and construction. Yiyu Gui contributed to manuscript refinement and figure preparation. Guibo Luo and Yuchao Yang supervised the research and provided critical guidance throughout the project.
\decl{Acknowledgments}
This work has been supported by the National Key R\&D Program of China (2025YFB4507300), Guangdong S\&T Program (2025B0101140001, 2026B0101070006), Guangdong Provincial Key Laboratory of In-Memory Computing Chips (2024B1212020002), Shenzhen Science and Technology Program (ZDCY20250901103401002, JCYJ20241202125907011), and Beijing Natural Science Foundation (L234026, L257010). This work has been supported by the New Cornerstone Science Foundation and Financial Support for Outstanding Scientific and Technological Innovation Talents Training Fund in Shenzhen.
\decl{Use of AI Tools}
The web version of ChatGPT (OpenAI; model: GPT-5) was used solely for English language editing and improving clarity of phrasing. The authors reviewed and edited all AI-assisted text and take full responsibility for the final content.
\decl{Funding}
This work was supported by the Shenzhen Science and Technology Program (JCYJ20241202125907011).
\decl{Conflicts of Interest}
The authors declare no conflicts of interest.
\decl{Data Availability Statement}
The source code, model implementation, and key scripts are available at \url{https://github.com/mingzhi-c/metis-brain-signal-foundation-model}.
\endgroup

\begingroup\interlinepenalty=10000\fontsize{8}{10.5}\selectfont
\endgroup
\end{textflow}
\clearpage
\label{main:end}
\setcounter{figure}{0}\setcounter{table}{0}
\renewcommand{\thefigure}{S\arabic{figure}}\renewcommand{\thetable}{S\arabic{table}}
\section*{Supporting Information}
\addcontentsline{toc}{section}{Supporting Information}
\noindent\textbf{A Language-Guided Multimodal Foundation Model for Zero-Shot and Multi-Task Brain Signal Analysis}\par
\vspace{5pt}\noindent Mingzhi Chen, Yiyu Gui, Guibo Luo, Yuchao Yang\par\vspace{9pt}
Each dataset is associated with one instruction sentence and a fixed answer set. In the multiple-choice format the same answer set is inserted into the prompt as lettered options.\par
\smallskip
\begingroup
\fontsize{8.5}{10.5}\selectfont
\setlength{\tabcolsep}{3pt}\renewcommand{\arraystretch}{1.18}\setlength{\LTpre}{6pt}\setlength{\LTpost}{8pt}
\begin{longtable}{@{}>{\raggedright\arraybackslash}p{29.543mm}>{\raggedright\arraybackslash}p{55.610mm}>{\raggedright\arraybackslash}p{88.629mm}@{}}
\caption{Instruction-answer templates used to build the pretraining QA pairs.}\label{tab:S1}\\
\toprule
\textbf{Dataset} & \textbf{Instruction} & \textbf{Answer}\\\midrule
\endfirsthead
\multicolumn{3}{@{}l}{\fontsize{8}{10}\selectfont\textbf{Table S1.} Continued.}\\\toprule
\textbf{Dataset} & \textbf{Instruction} & \textbf{Answer}\\\midrule
\endhead
\midrule\multicolumn{3}{r@{}}{\fontsize{8}{10}\selectfont Continued on next page}\\\endfoot
\bottomrule\endlastfoot
HUP & Which epilepsy state does this signal belong to? & "Interictal", "Ictal",\par "Preictal", "Postictal"\\
UPenn & Which epilepsy state does this signal belong to? & "Interictal", "Ictal"\\
SWEC\_ETHZ & Which epilepsy state does this signal belong to? & "Preictal", "Ictal",\par "Postictal"\\
FNUSA & Which epilepsy state does this signal belong to? & "Interictal",\par "Interictal, Pathological activity"\\
SHHS & Which sleep stage does this signal belong to? & "Wake", "Non-REM Stage 1",\par "Non-REM Stage 2", "Non-REM Stage 3",\par "Rapid Eye Movement"\\
SeiZelT2 & Which epilepsy state does this signal belong to? & "Interictal", "Ictal",\par "Preictal", "Postictal"\\
TUAB & Is the signal normal or abnormal? & "Normal", "Abnormal"\\
CHB MIT & Which epilepsy state does this signal belong to? & "Interictal", "Ictal",\par "Preictal", "Postictal"\\
SleepEDF & Which sleep stage does this signal belong to? & "Wake", "Non-REM Stage 1",\par "Non-REM Stage 2", "Non-REM Stage 3",\par "Rapid Eye Movement"\\
HaaglandenSleep & Which sleep stage does this signal belong to? & "Wake", "Non-REM Stage 1",\par "Non-REM Stage 2", "Non-REM Stage 3",\par "Rapid Eye Movement"\\
TDBrain & Which disease does this signal belong to? & dataset-specific clinical labels (see Table~\ref{tab:S2})\\
TUEV & Which type does this signal belong to? & "Spike and slow wave", " Generalized periodic epileptiform discharge",\par "Periodic lateralized epileptiform dischage", "Eye movement",\par "Artifact", "Background"\\
ADFTD & Which disease does this signal belong to? & "Normal", "Alzheimer's disease",\par "Frontotemporal dementia"\\
BrainLat & Which disease does this signal belong to? & "Normal", "Alzheimer's disease",\par "Frontotemporal dementia", "Parkinson's disease", "Multiple sclerosis"\\
AD-Auditory & Which disease does this signal belong to? & "Normal", "Alzheimer's disease",\par "Mild cognitive impairment"\\
ShuMI & Which type does this signal belong to? & "Motor imagery left hand",\par "Motor imagery right hand"\\
REEG-PD & Which disease does this signal belong to? & "Normal", "Parkinson's disease"\\
PhysionetMI & Which type does this signal belong to? & "Motor imagery left fist",\par "Motor imagery right fist",\par "Motor imagery both fists",\par "Motor imagery both feet"\\
TUSZ & Which type does this signal belong to? & dataset-specific clinical labels (see Table~\ref{tab:S3})\\
TUEP & Does this signal belong to epilepsy? & "No", "Yes"\\
\end{longtable}\endgroup
\begingroup
\fontsize{9}{11}\selectfont
\setlength{\tabcolsep}{3pt}\renewcommand{\arraystretch}{1.18}\setlength{\LTpre}{6pt}\setlength{\LTpost}{8pt}
\begin{longtable}{@{}>{\raggedright\arraybackslash}p{13.192mm}>{\raggedright\arraybackslash}p{162.699mm}@{}}
\caption{TDBrain label inventory and canonical answer strings}\label{tab:S2}\\
\toprule
\textbf{Class} & \textbf{Canonical answer string}\\\midrule
\endfirsthead
\multicolumn{2}{@{}l}{\fontsize{8}{10}\selectfont\textbf{Table S2.} Continued.}\\\toprule
\textbf{Class} & \textbf{Canonical answer string}\\\midrule
\endhead
\midrule\multicolumn{2}{r@{}}{\fontsize{8}{10}\selectfont Continued on next page}\\\endfoot
\bottomrule\endlastfoot
0 & "Burnout"\\
1 & "Subjective memory complaints"\\
2 & "Normal"\\
3 & "Dyslexia"\\
4 & "Chronic pain"\\
5 & "Major depressive disorder"\\
6 & "Attention deficit hyperactivity disorder"\\
7 & "Attention deficit hyperactivity disorder, asperger syndrome"\\
8 & "Pervasive developmental disorder not otherwise specified, dyslexia"\\
9 & "Pervasive developmental disorder not otherwise specified"\\
10 & "Whiplash"\\
11 & "Anxiety"\\
12 & "Attention deficit hyperactivity disorder, dyslexia"\\
13 & "Autism spectrum disorder"\\
14 & "Tinnitus"\\
15 & "Obsessive-compulsive disorder"\\
16 & "Panic disorder"\\
17 & "Major depressive disorder, anxiety"\\
18 & "Migraine"\\
19 & "Pervasive developmental disorder not otherwise specified, anxiety"\\
20 & "Parkinson's disease"\\
21 & "Bipolar disorder"\\
22 & "Major depressive disorder, bipolar disorder"\\
23 & "Dyspraxia"\\
24 & "Tinnitus, major depressive disorder"\\
25 & "Attention deficit hyperactivity disorder, autism spectrum disorder, anxiety"\\
26 & "Major depressive disorder, attention deficit hyperactivity disorder"\\
27 & "Attention deficit hyperactivity disorder, pervasive developmental disorder not otherwise specified"\\
28 & "Asperger syndrome"\\
29 & "Attention deficit hyperactivity disorder, epilepsy"\\
30 & "Major depressive disorder, pain"\\
31 & "Pervasive developmental disorder not otherwise specified, gilles de la tourette syndrome"\\
32 & "Pervasive developmental disorder not otherwise specified, attention deficit hyperactivity disorder"\\
33 & "Pervasive developmental disorder not otherwise specified, autism spectrum disorder"\\
34 & "Traumatic brain injury"\\
35 & "Attention deficit hyperactivity disorder, anxiety"\\
36 & "Attention deficit hyperactivity disorder, dyslexia, dyscalculia"\\
37 & "Attention deficit hyperactivity disorder, major depressive disorder"\\
38 & "Major depressive disorder, panic disorder"\\
39 & "Depersonalization disorder"\\
40 & "Major depressive disorder, trauma"\\
41 & "Post-traumatic stress disorder, attention deficit hyperactivity disorder"\\
42 & "Obsessive-compulsive disorder, dissociative psychogenic seizures"\\
43 & "Major depressive disorder, obsessive-compulsive disorder"\\
44 & "Major depressive disorder, tumor"\\
45 & "Attention deficit hyperactivity disorder, gilles de la tourette syndrome"\\
46 & "Obsessive-compulsive disorder, major depressive disorder"\\
47 & "Conversion disorder"\\
48 & "Autism spectrum disorder, asperger syndrome"\\
49 & "Major depressive disorder, attention deficit hyperactivity disorder, lyme disease"\\
50 & "Attention deficit hyperactivity disorder, obsessive-compulsive disorder"\\
51 & "Multiple system atrophy - cerebellar type"\\
52 & "Obsessive-compulsive disorder, autism spectrum disorder"\\
53 & "Stroke, pain"\\
54 & "Stroke"\\
55 & "Major depressive disorder, obsessive-compulsive disorder, attention deficit hyperactivity disorder"\\
56 & "Epilepsy, obsessive-compulsive disorder"\\
57 & "Insomnia"\\
58 & "Major depressive disorder, attention deficit hyperactivity disorder, anorexia"\\
59 & "Major depressive disorder, anxiety, tinnitus"\\
\end{longtable}\endgroup
\begingroup
\fontsize{9}{11}\selectfont
\setlength{\tabcolsep}{3pt}\renewcommand{\arraystretch}{1.18}\setlength{\LTpre}{6pt}\setlength{\LTpost}{8pt}
\begin{longtable}{@{}>{\raggedright\arraybackslash}p{13.192mm}>{\raggedright\arraybackslash}p{162.699mm}@{}}
\caption{TUSZ label inventory and canonical answer strings}\label{tab:S3}\\
\toprule
\textbf{Class} & \textbf{Canonical answer string}\\\midrule
\endfirsthead
\multicolumn{2}{@{}l}{\fontsize{8}{10}\selectfont\textbf{Table S3.} Continued.}\\\toprule
\textbf{Class} & \textbf{Canonical answer string}\\\midrule
\endhead
\midrule\multicolumn{2}{r@{}}{\fontsize{8}{10}\selectfont Continued on next page}\\\endfoot
\bottomrule\endlastfoot
0 & "Spike/Sharp and Wave"\\
1 & "Generalized Periodic Epileptiform Discharges"\\
2 & "Periodic Lateralized Epileptiform Discharges"\\
3 & "Eye blink"\\
4 & "Artifacts"\\
5 & "Epilepsy ictal state, Seizure"\\
6 & "Epilepsy ictal state, Focal Non-Specific Seizure"\\
7 & "Epilepsy ictal state, Generalized Non-Specific Seizure"\\
8 & "Epilepsy ictal state, Simple Partial Seizure"\\
9 & "Epilepsy ictal state, Complex Partial Seizure"\\
10 & "Epilepsy ictal state, Absence Seizure"\\
11 & "Epilepsy ictal state, Tonic Seizure"\\
12 & "Epilepsy ictal state, Clonic Seizure"\\
13 & "Epilepsy ictal state, Tonic Clonic Seizure"\\
14 & "Epilepsy ictal state, Atonic Seizure"\\
15 & "Epilepsy ictal state, Myoclonic Seizure"\\
16 & "Panic disorder"\\
17 & "Major depressive disorder, anxiety"\\
18 & "Epilepsy ictal state, Non-Epileptic Seizure"\\
19 & "Interesting Patterns"\\
20 & "Slowing"\\
21 & "Eye Movement Artifact"\\
22 & "Chewing Artifact"\\
23 & "Shivering Artifact"\\
24 & "Muscle Artifact"\\
25 & "Electrode Pop Artifact"\\
26 & "Electrostatic Artifact"\\
27 & "Calibration Artifact"\\
28 & "Hypnagogic Hypersynchrony"\\
29 & "Triphasic Wave"\\
\end{longtable}\endgroup
\clearpage
\noindent\begin{minipage}{\textwidth}\centering\includegraphics[width=172mm]{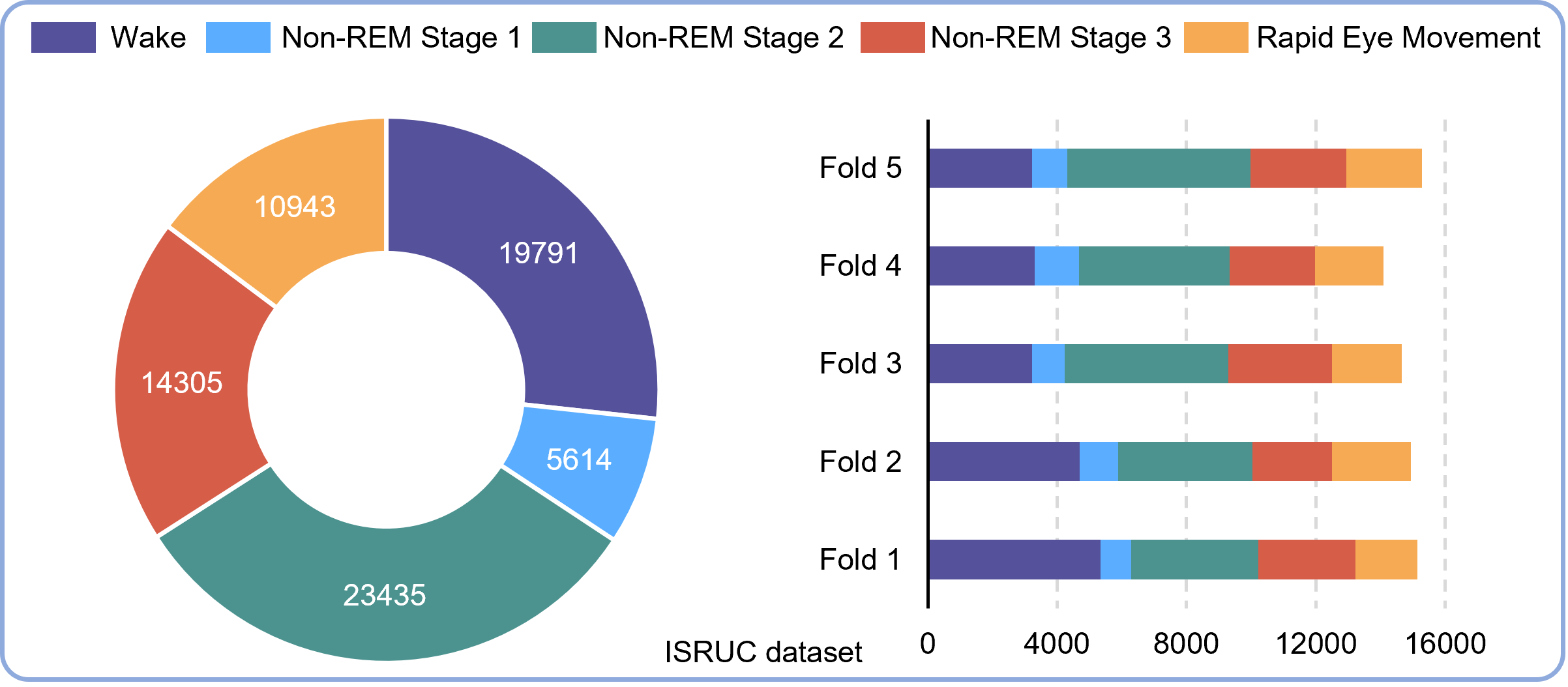}\par
\captionof{figure}{\textbf{ISRUC dataset class composition: overall counts and per fold proportions.}}\label{fig:S1}
\end{minipage}\par
\vspace{4pt}
\noindent\begin{minipage}{\textwidth}\centering\includegraphics[width=172mm]{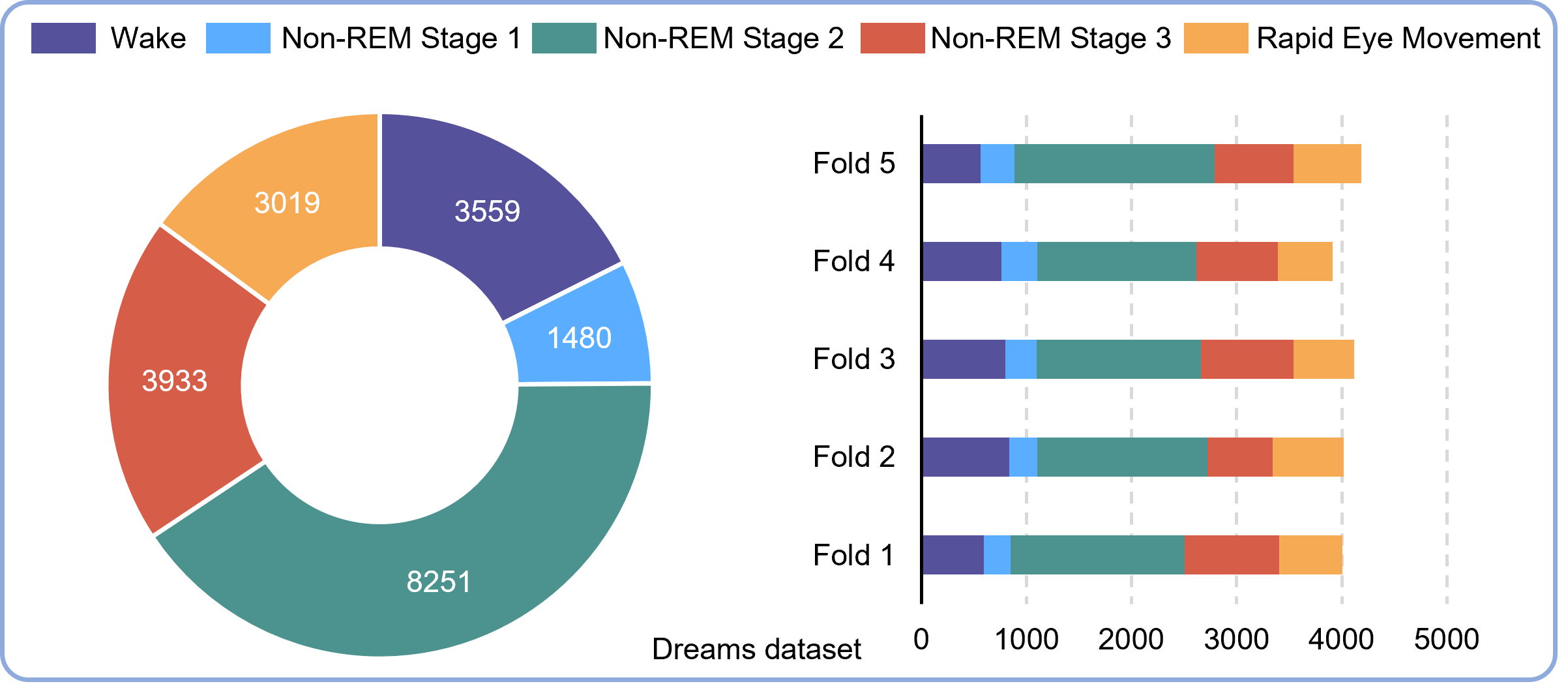}\par
\captionof{figure}{\textbf{Dreams dataset class composition: overall counts and per fold proportions.}}\label{fig:S2}
\end{minipage}\par
\vspace{4pt}
\noindent\begin{minipage}{\textwidth}\centering\includegraphics[width=172mm]{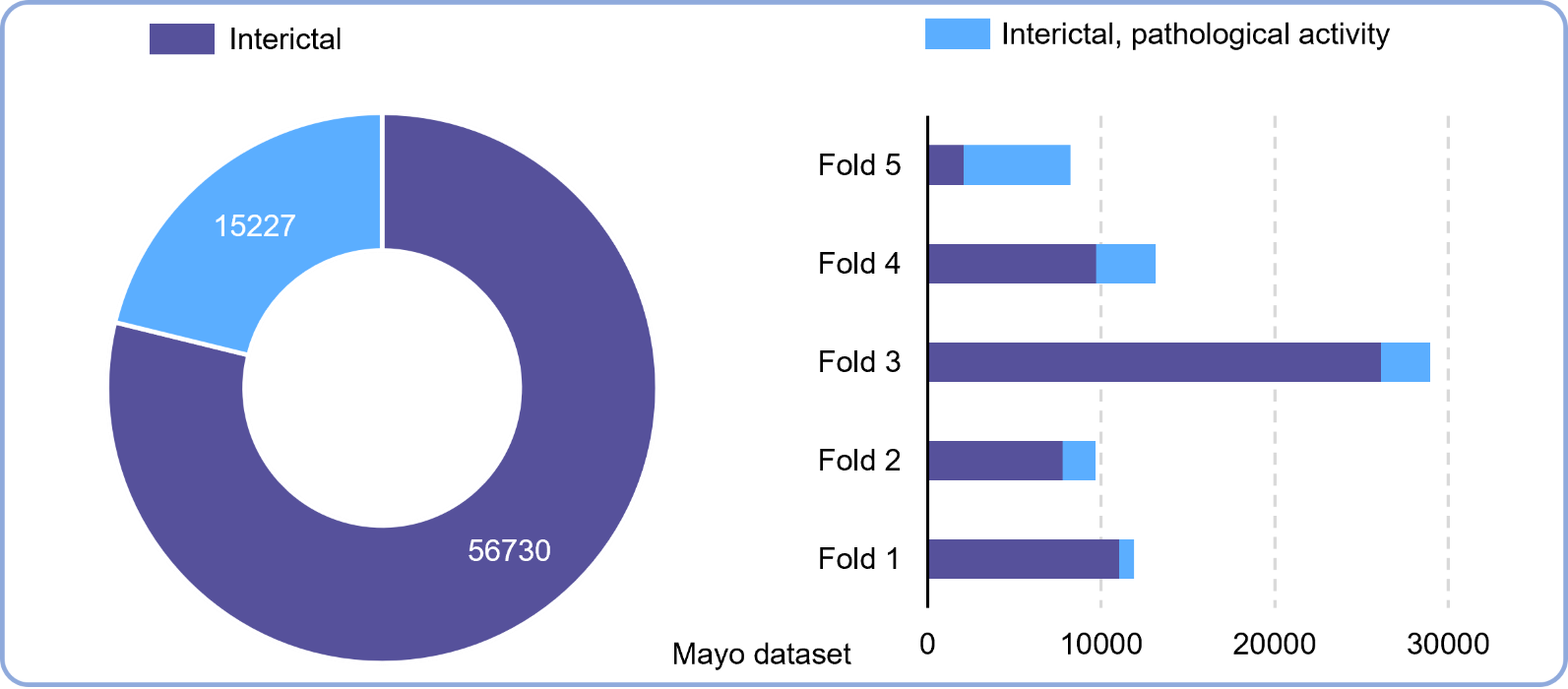}\par
\captionof{figure}{\textbf{Mayo dataset class composition: overall counts and per fold proportions.}}\label{fig:S3}
\end{minipage}\par
\vspace{4pt}
\clearpage
\noindent\begin{minipage}{\textwidth}\centering\includegraphics[width=172mm]{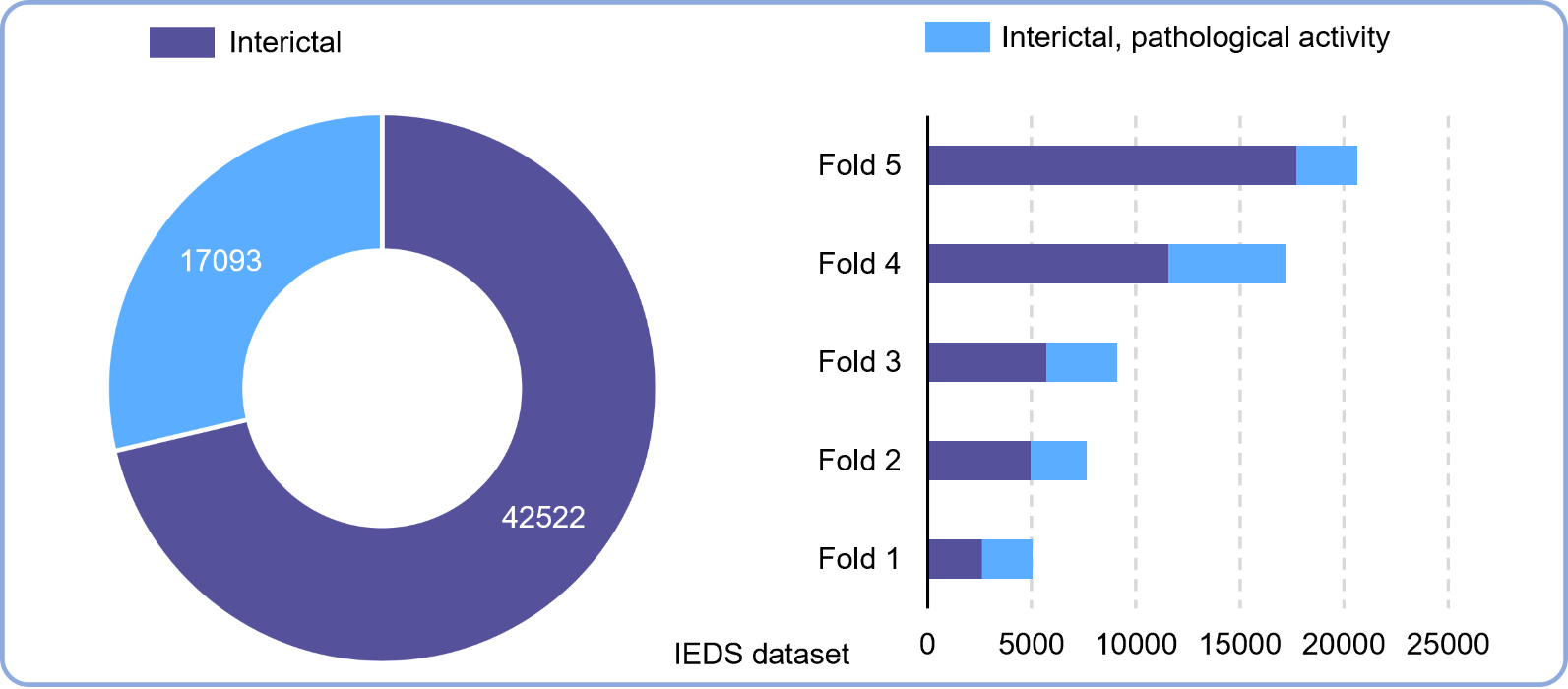}\par
\captionof{figure}{\textbf{IEDS dataset class composition: overall counts and per fold proportions.}}\label{fig:S4}
\end{minipage}\par
\vspace{4pt}
\noindent\begin{minipage}{\textwidth}\centering\includegraphics[width=172mm]{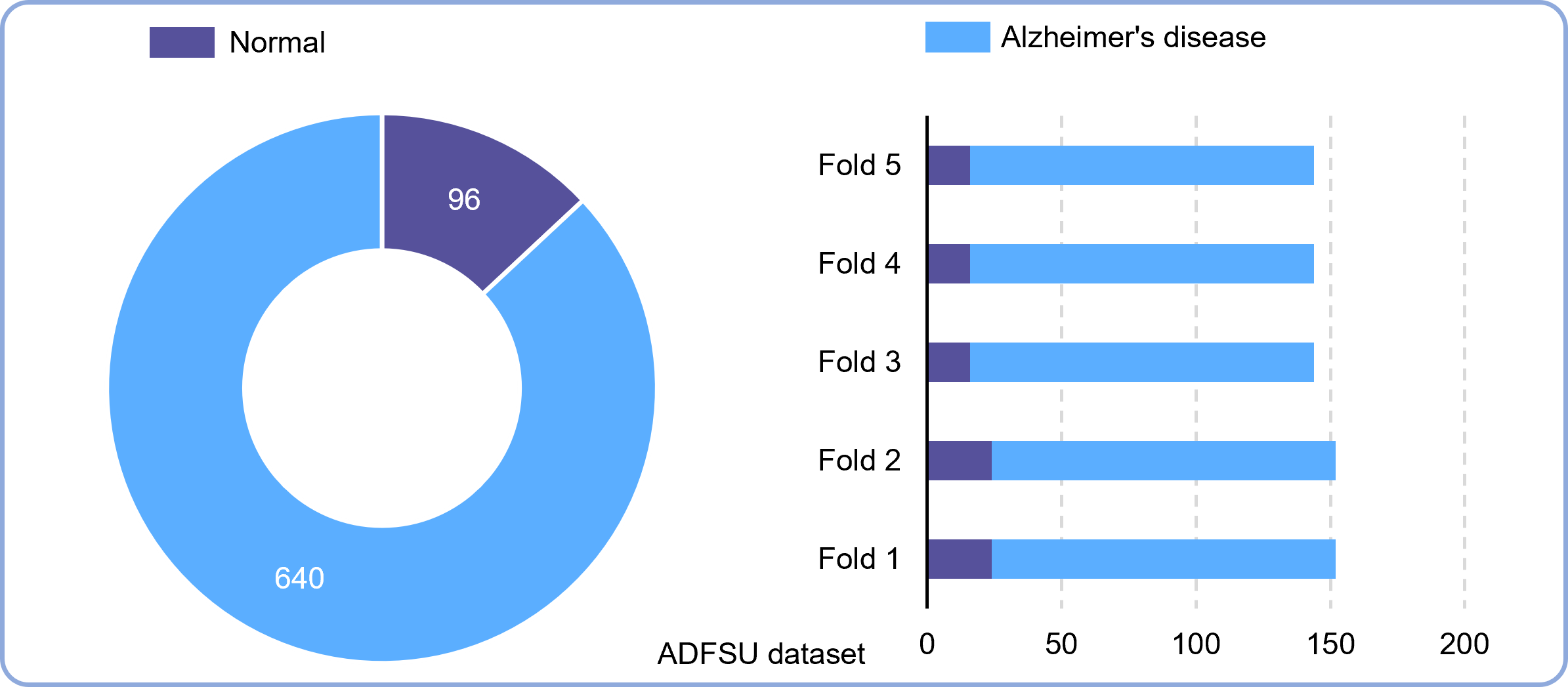}\par
\captionof{figure}{\textbf{ADFSU dataset class composition: overall counts and per fold proportions.}}\label{fig:S5}
\end{minipage}\par
\vspace{4pt}
\noindent\begin{minipage}{\textwidth}\centering\includegraphics[width=172mm]{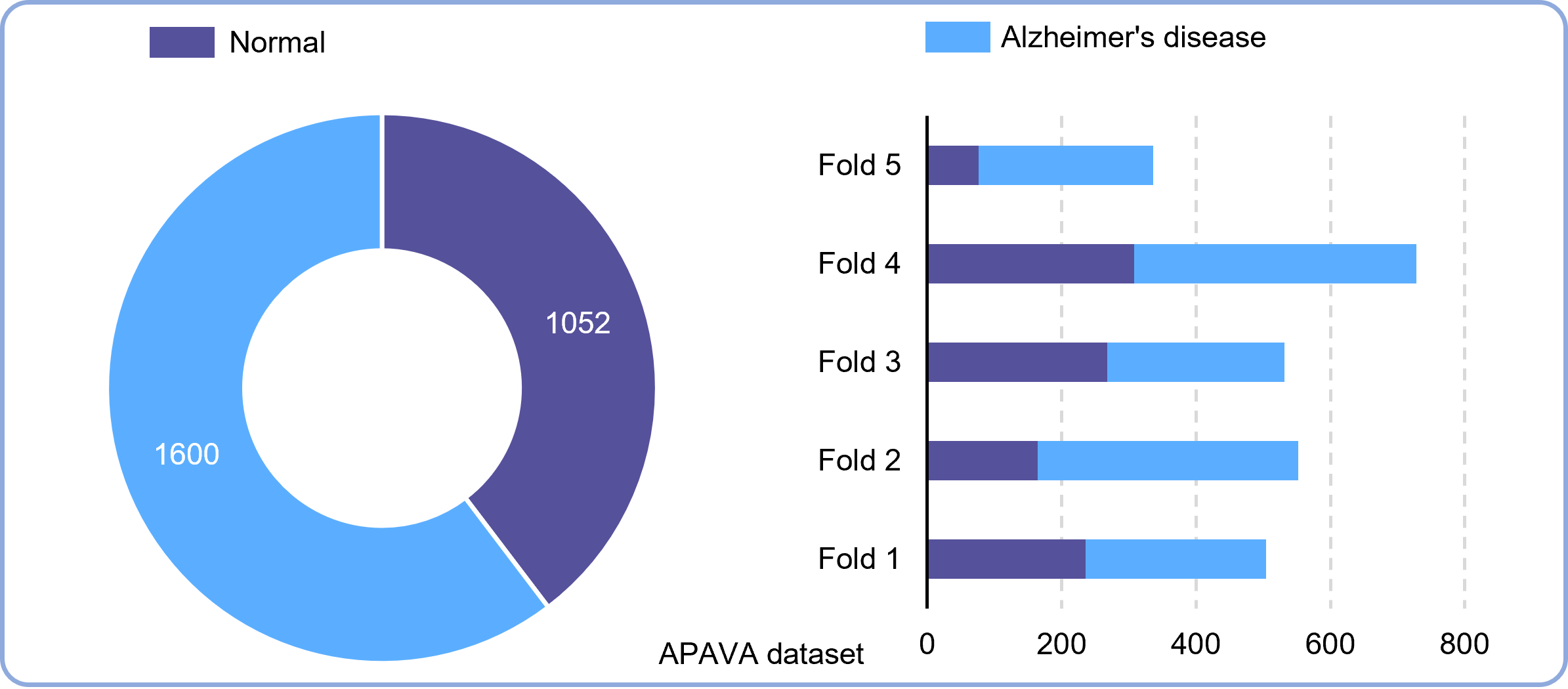}\par
\captionof{figure}{\textbf{APAVA dataset class composition: overall counts and per fold proportions.}}\label{fig:S6}
\end{minipage}\par
\vspace{4pt}
\clearpage
\noindent\begin{minipage}{\textwidth}\centering\includegraphics[width=172mm]{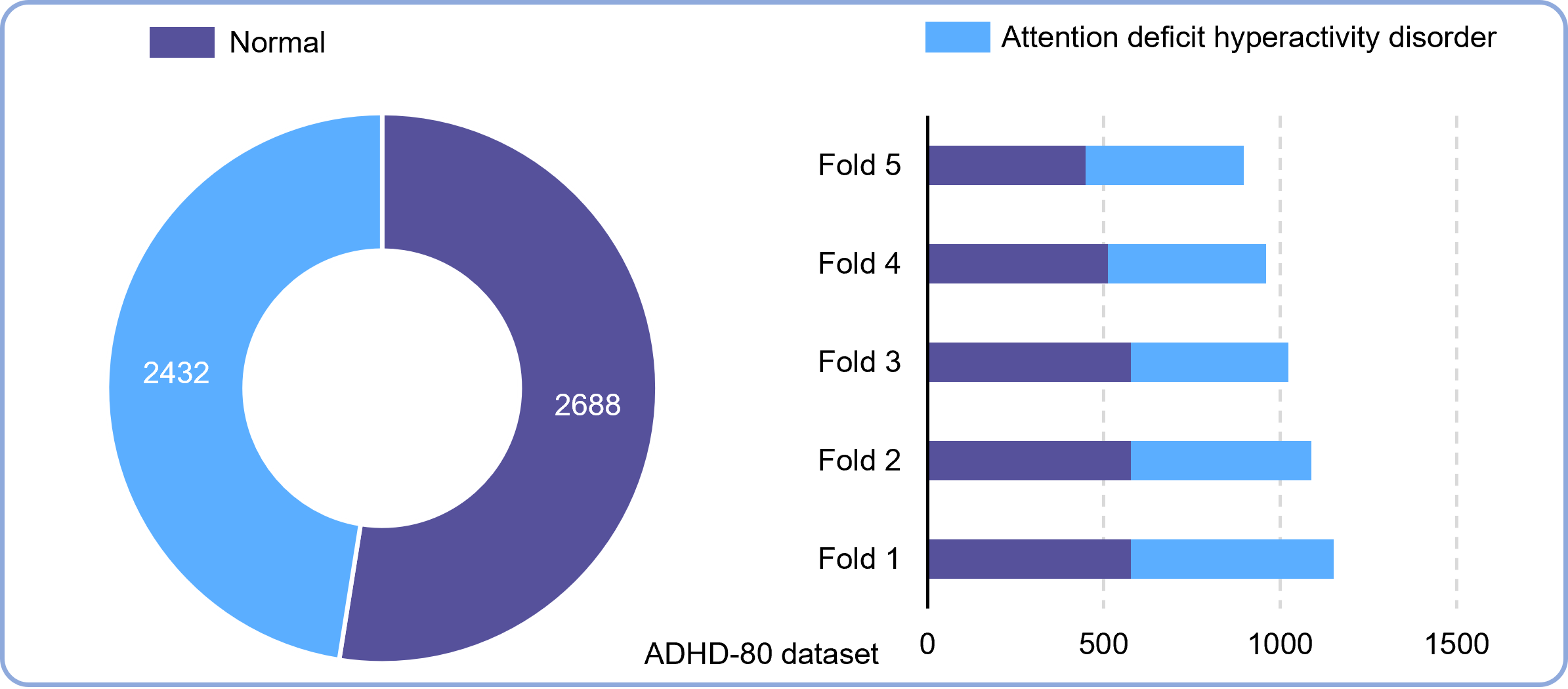}\par
\captionof{figure}{\textbf{ADHD-80 dataset class composition: overall counts and per fold proportions.}}\label{fig:S7}
\end{minipage}\par
\vspace{4pt}
\noindent\begin{minipage}{\textwidth}\centering\includegraphics[width=172mm]{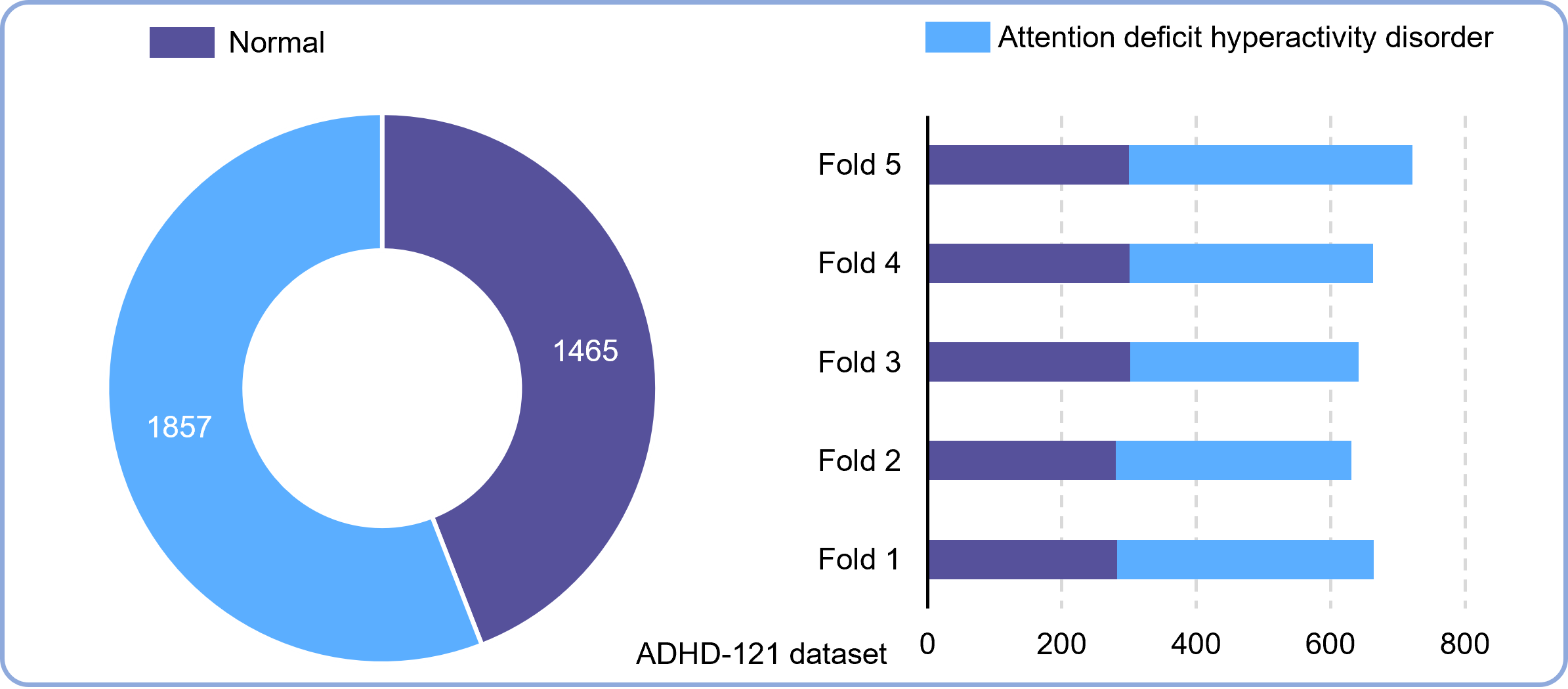}\par
\captionof{figure}{\textbf{ADHD-121 dataset class composition: overall counts and per fold proportions.}}\label{fig:S8}
\end{minipage}\par
\vspace{4pt}
\noindent\begin{minipage}{\textwidth}\centering\includegraphics[width=172mm]{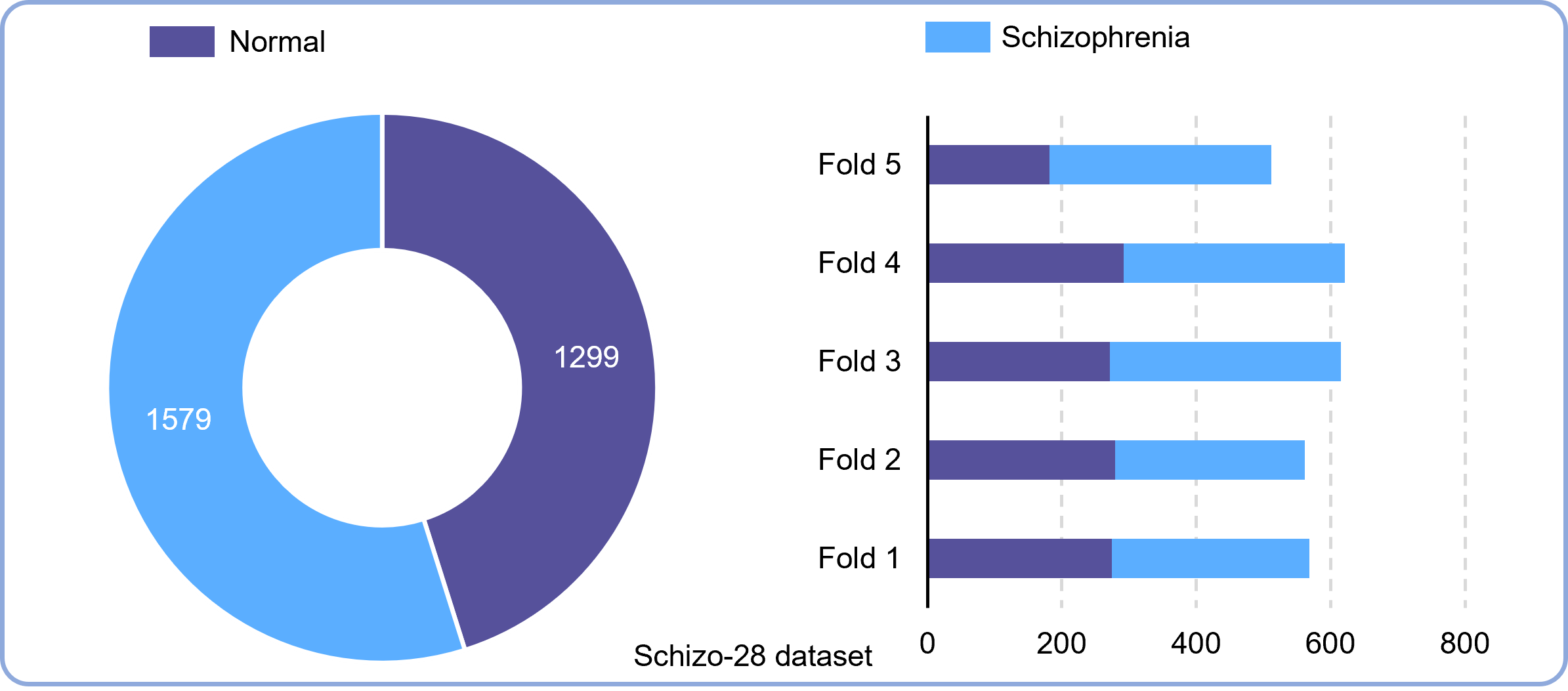}\par
\captionof{figure}{\textbf{Schizo-28 dataset class composition: overall counts and per fold proportions.}}\label{fig:S9}
\end{minipage}\par
\vspace{4pt}
\clearpage
\noindent\begin{minipage}{\textwidth}\centering\includegraphics[width=172mm]{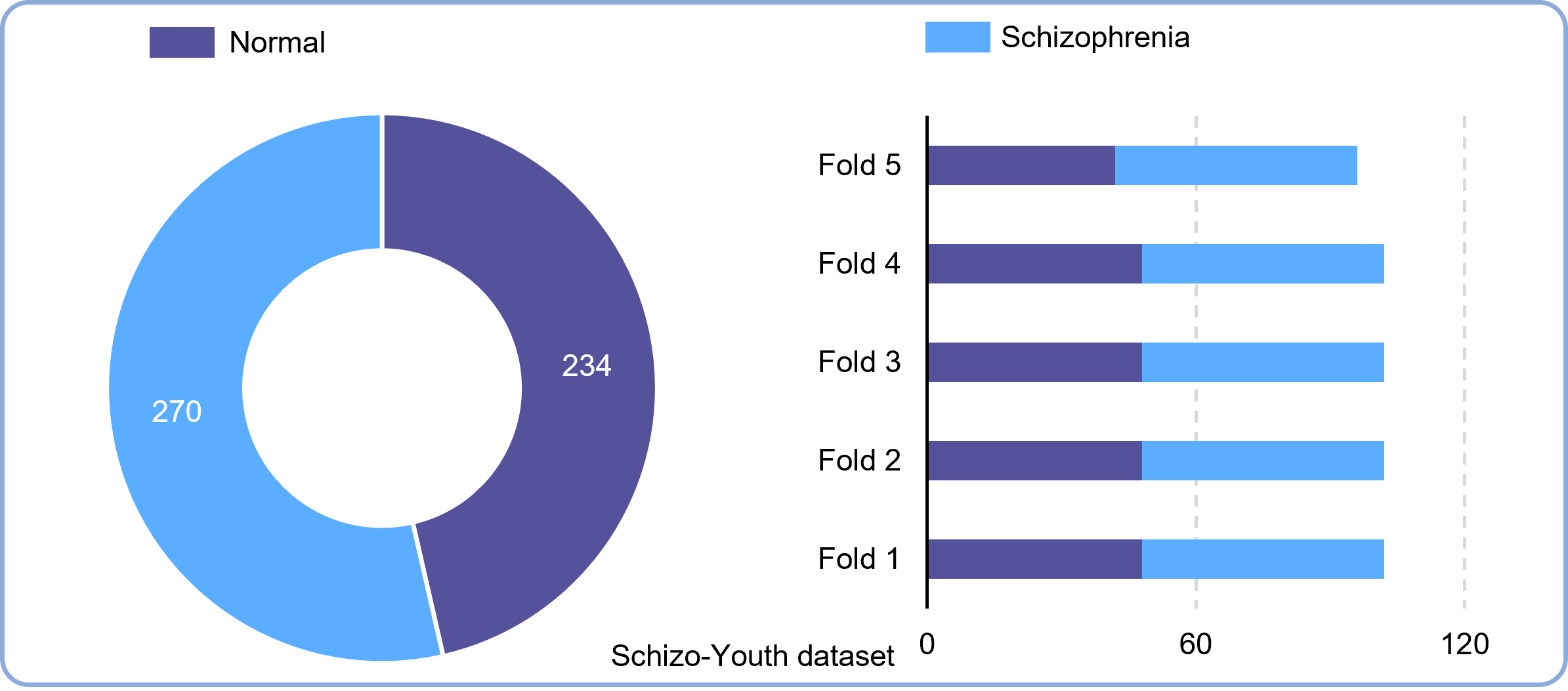}\par
\captionof{figure}{\textbf{Schizo-Youth dataset class composition: overall counts and per fold proportions.}}\label{fig:S10}
\end{minipage}\par
\vspace{4pt}
\noindent\begin{minipage}{\textwidth}\centering\includegraphics[width=172mm]{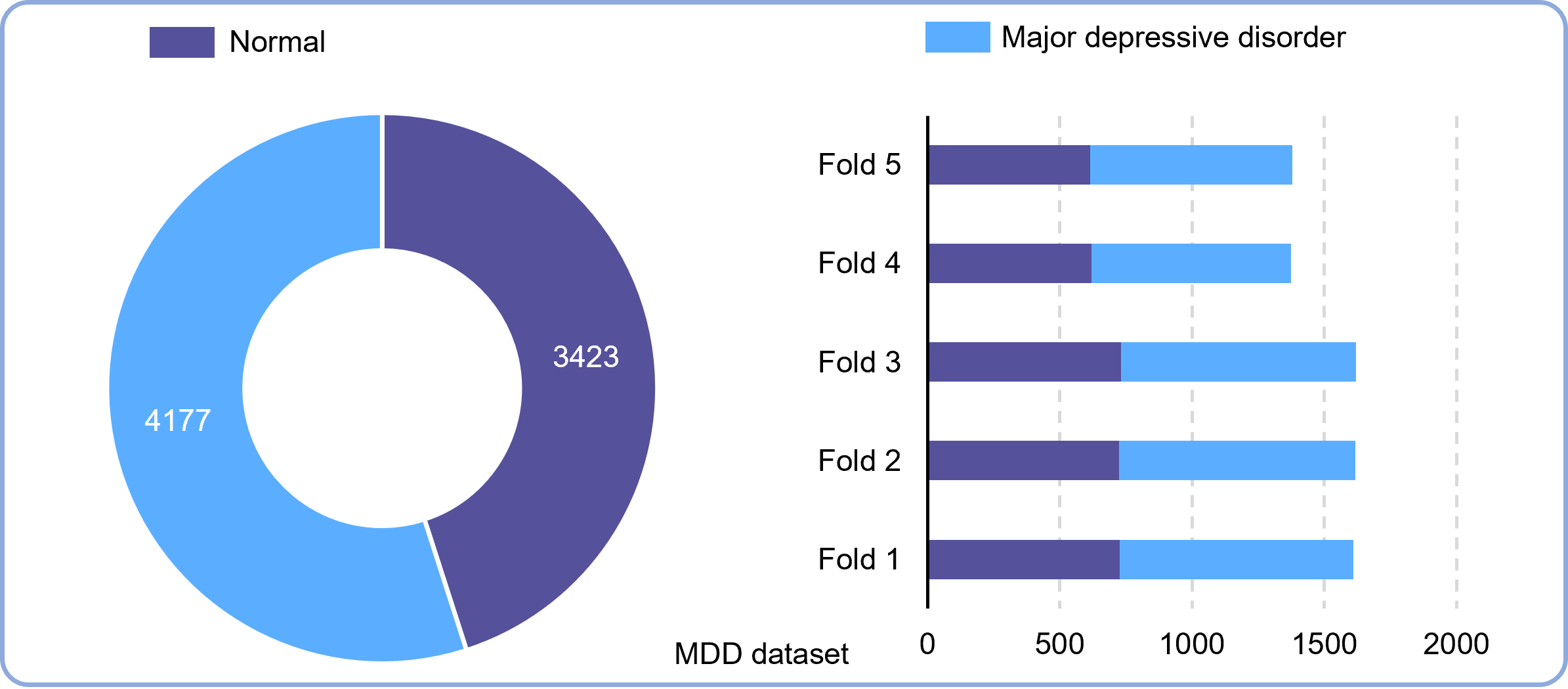}\par
\captionof{figure}{\textbf{MDD dataset class composition: overall counts and per fold proportions.}}\label{fig:S11}
\end{minipage}\par
\vspace{4pt}
\noindent\begin{minipage}{\textwidth}\centering\includegraphics[width=172mm]{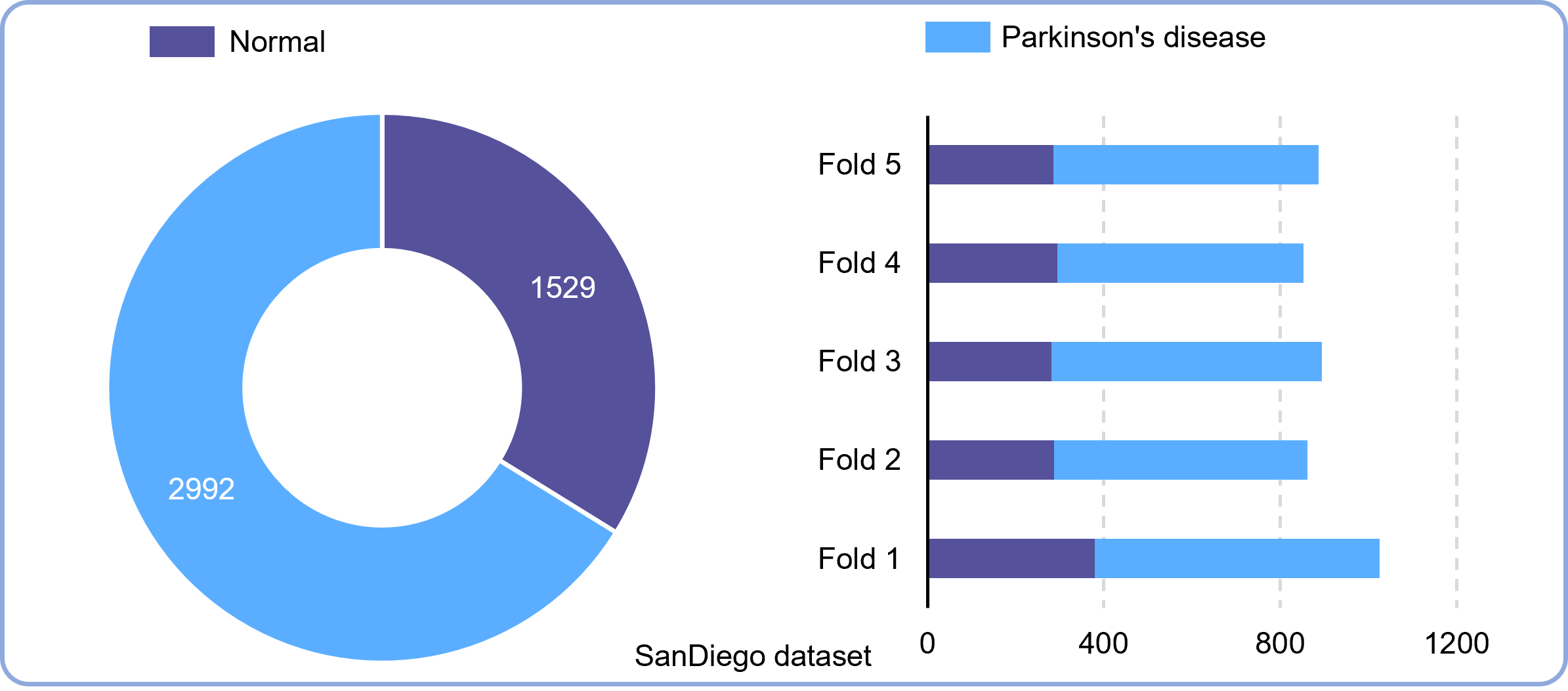}\par
\captionof{figure}{\textbf{SanDiego dataset class composition: overall counts and per fold proportions.}}\label{fig:S12}
\end{minipage}\par
\vspace{4pt}
\clearpage
\noindent\begin{minipage}{\textwidth}\centering\includegraphics[width=172mm]{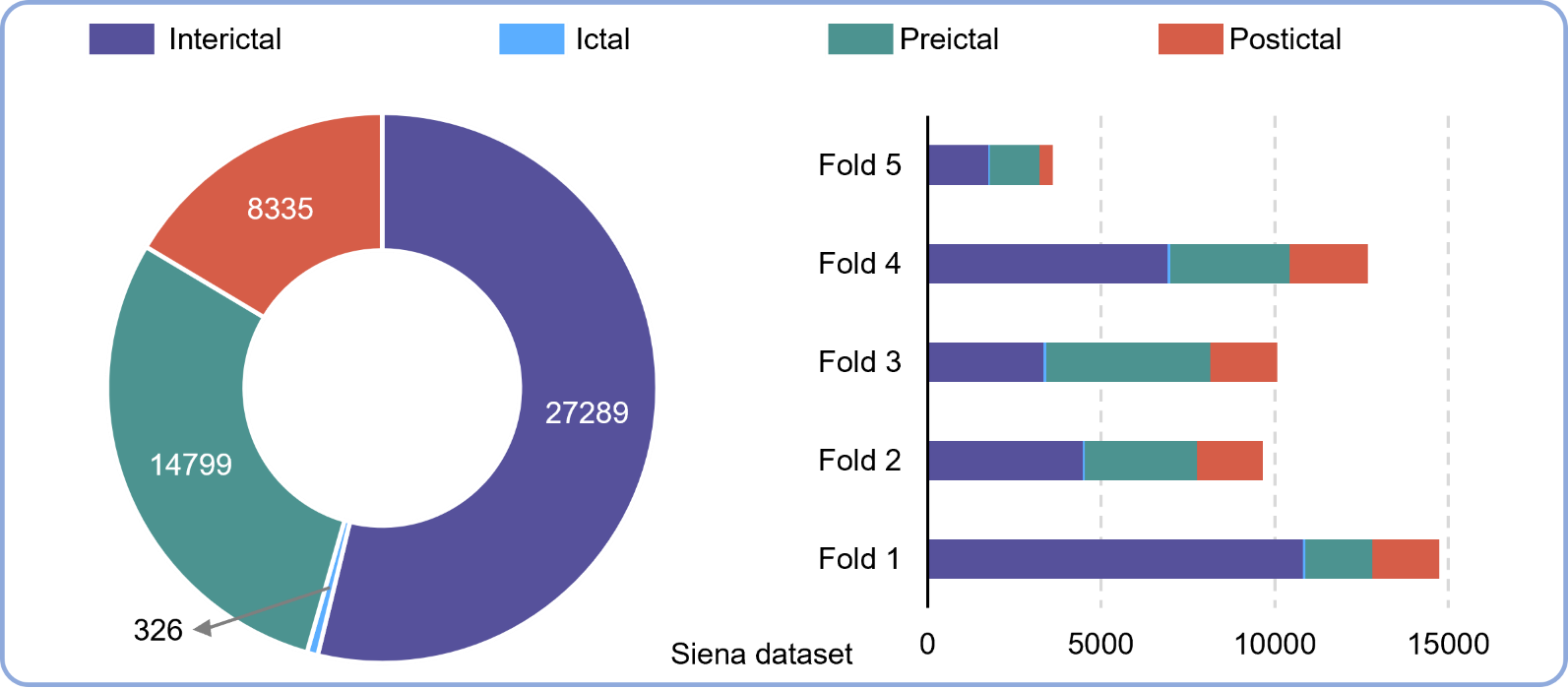}\par
\captionof{figure}{\textbf{Siena dataset class composition: overall counts and per fold proportions.}}\label{fig:S13}
\end{minipage}\par
\vspace{4pt}
\noindent\begin{minipage}{\textwidth}\centering\includegraphics[width=172mm]{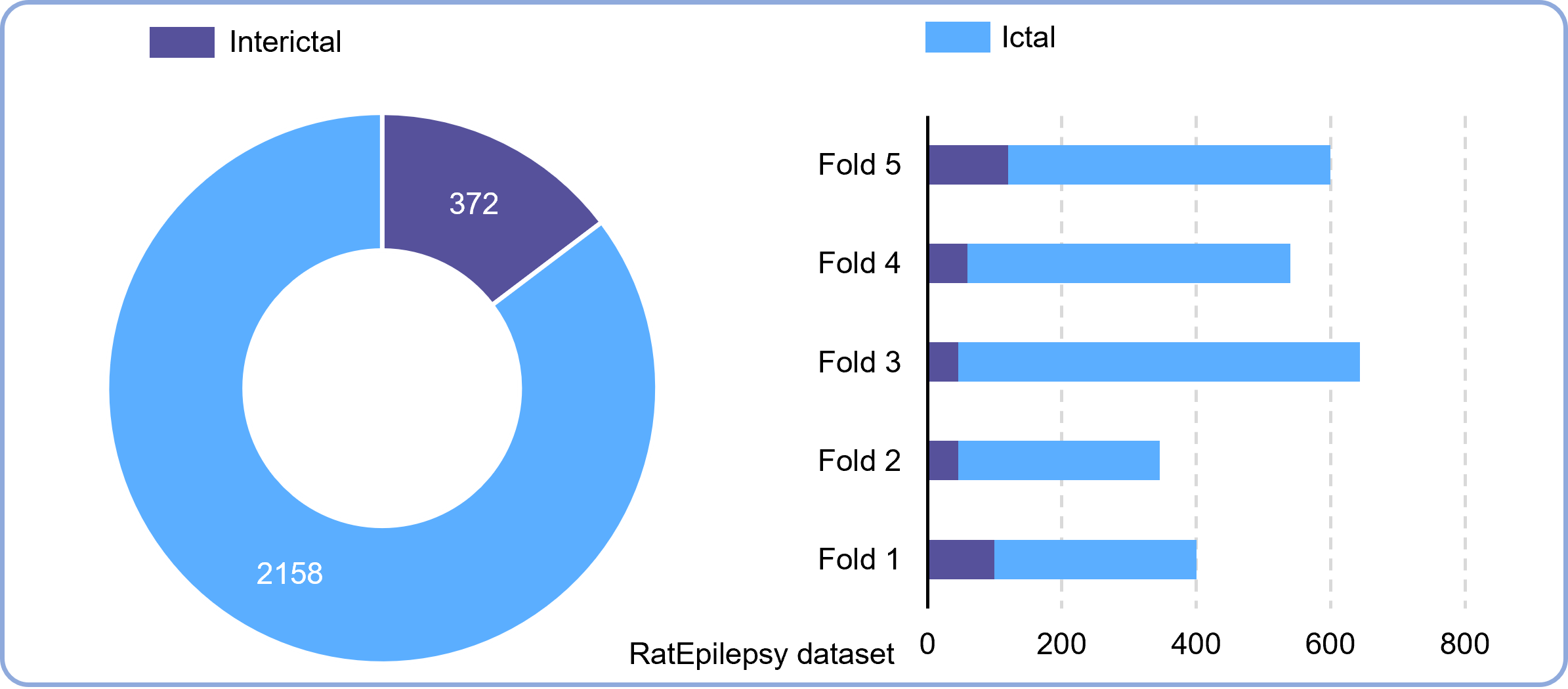}\par
\captionof{figure}{\textbf{RatEpilepsy dataset class composition: overall counts and per fold proportions.}}\label{fig:S14}
\end{minipage}\par
\vspace{4pt}
\noindent\begin{minipage}{\textwidth}\centering\includegraphics[width=172mm]{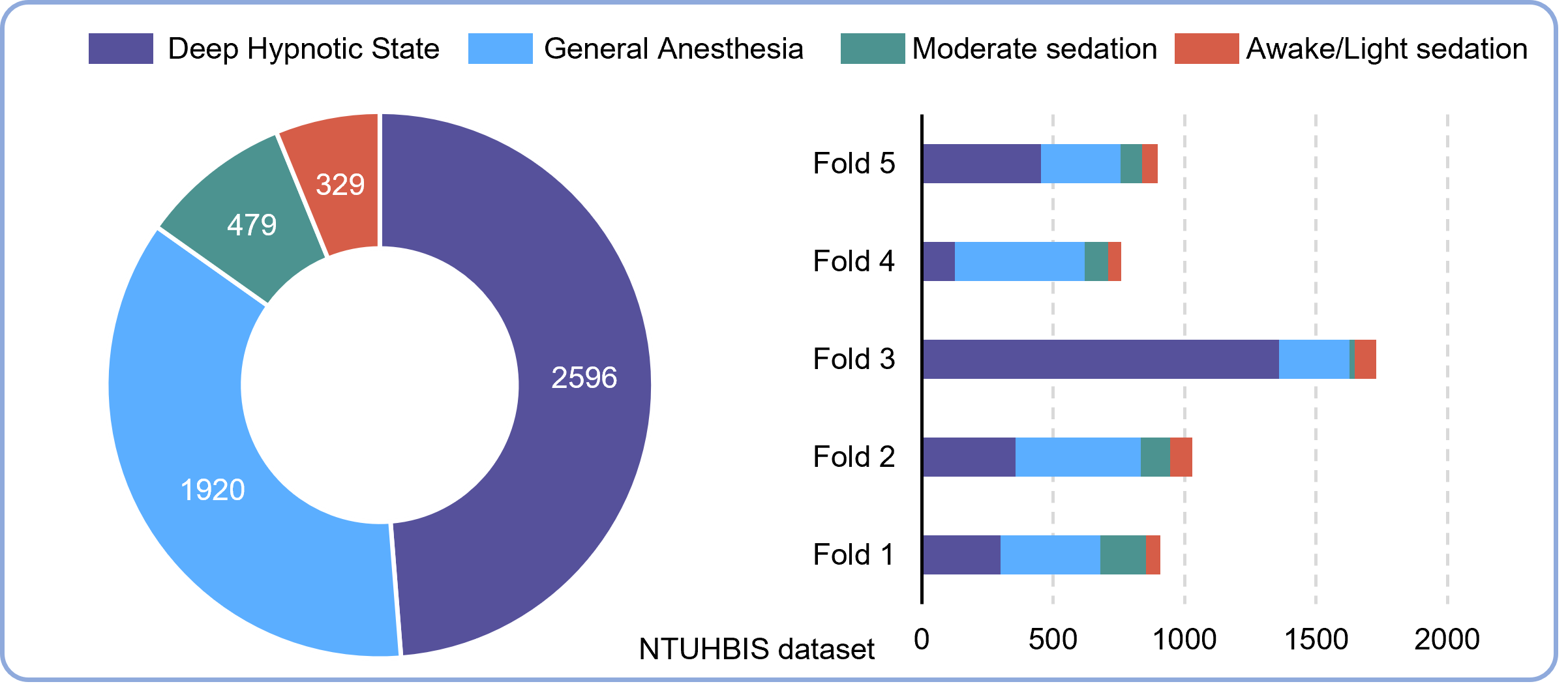}\par
\captionof{figure}{\textbf{NTUHBIS dataset class composition: overall counts and per fold proportions.}}\label{fig:S15}
\end{minipage}\par
\vspace{4pt}
\clearpage
\noindent\begin{minipage}{\textwidth}\centering\includegraphics[width=172mm]{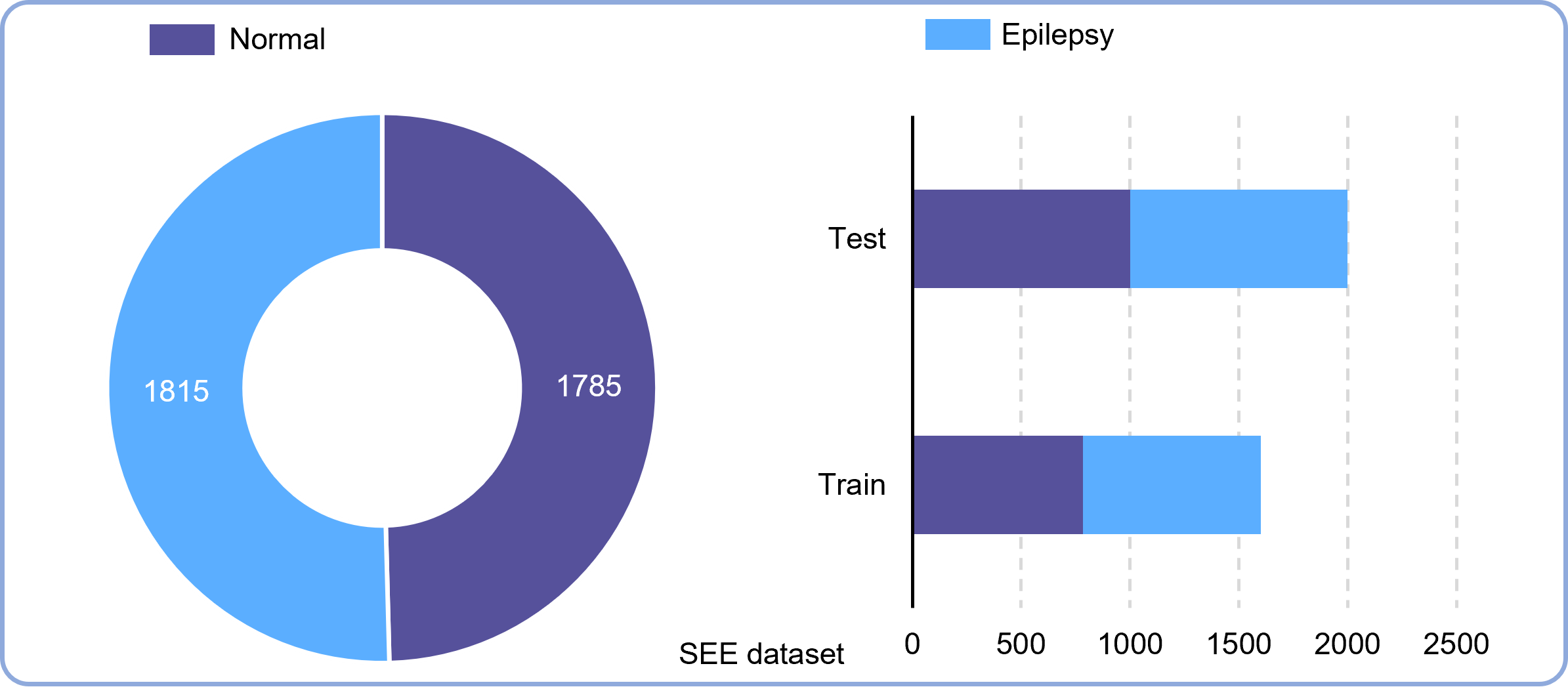}\par
\captionof{figure}{\textbf{SEE dataset class composition: train, test, and total.}}\label{fig:S16}
\end{minipage}\par
\vspace{4pt}
\noindent\begin{minipage}{\textwidth}\centering\includegraphics[width=172mm]{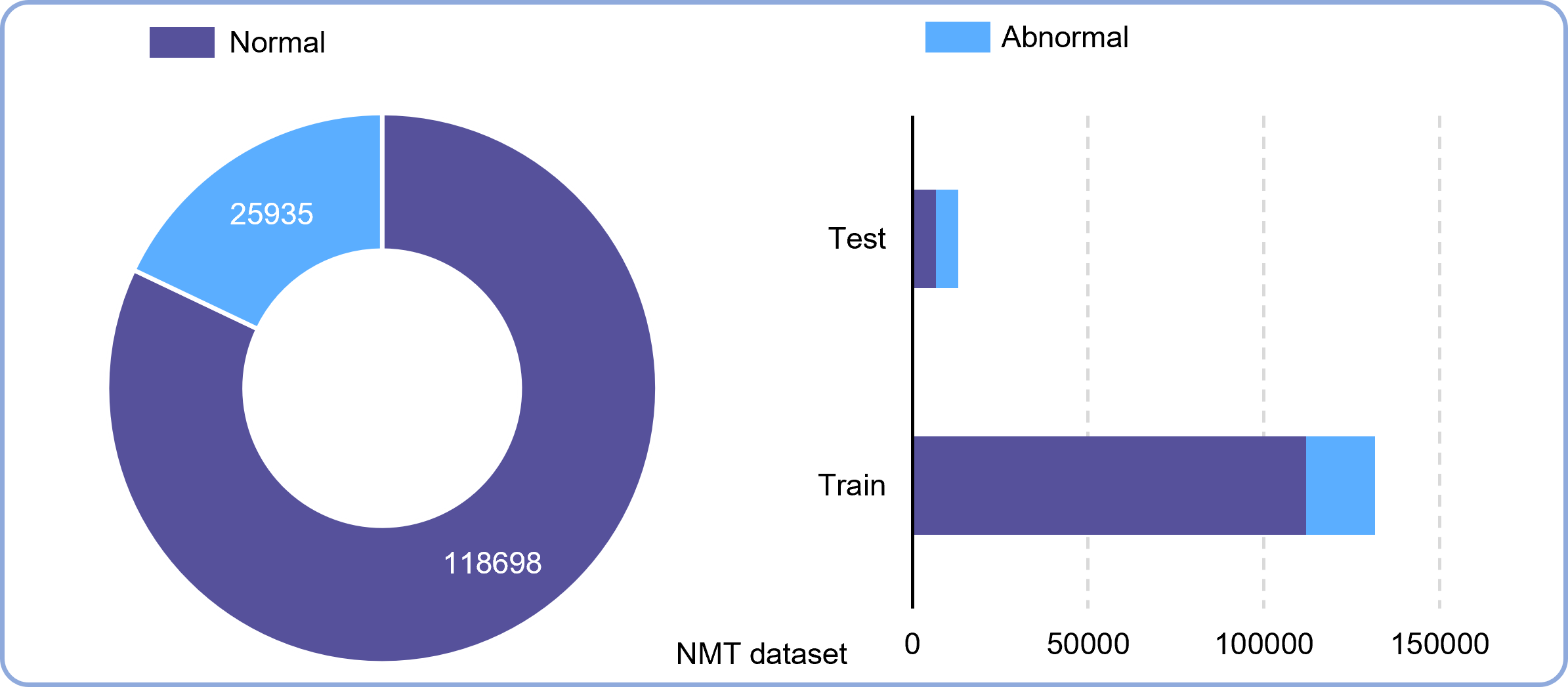}\par
\captionof{figure}{\textbf{NMT dataset class composition: train, test, and total.}}\label{fig:S17}
\end{minipage}\par
\vspace{4pt}
\clearpage
\noindent\begin{minipage}{\textwidth}\centering\includegraphics[width=176mm]{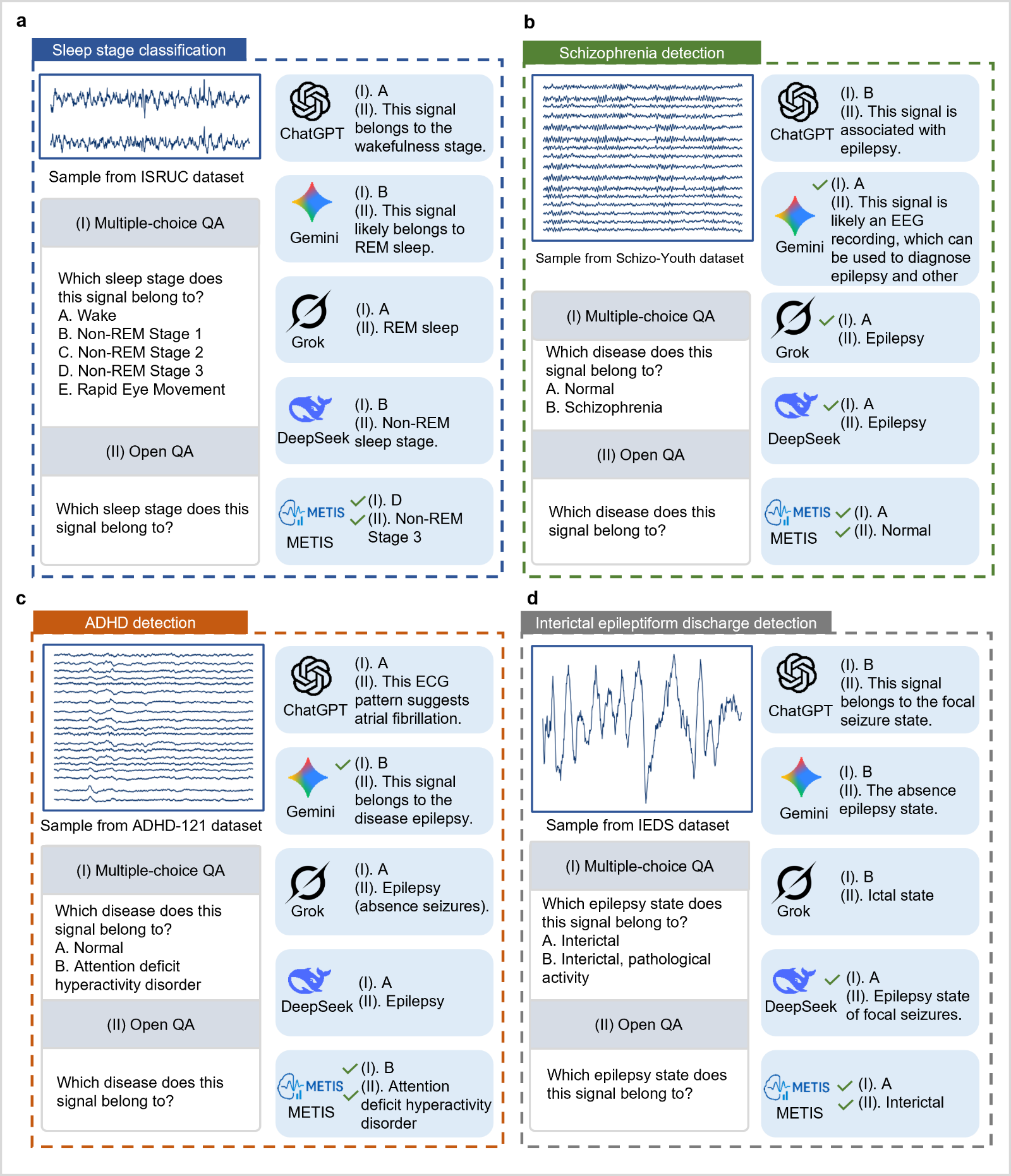}\par
\captionof{figure}{\textbf{Additional zero shot task examples on ISRUC, Schizo-Youth, ADHD-121, and IEDS.} \textbf{a}, Sleep stage classification on ISRUC. Each example uses the same signal and pairs one multiple-choice query with one open ended query. Model outputs are shown for METIS, ChatGPT, Gemini, Grok, and DeepSeek. Correct answers are marked with a green checkmark for readability. \textbf{b}, Schizophrenia detection on Schizo-Youth. Binary disease detection with options Normal and Schizophrenia under the multiple-choice format, together with an open ended query that expects a canonical label. Green checkmarks indicate correct responses. \textbf{c}, ADHD detection on ADHD-121. Binary disease detection with options Normal and Attention deficit hyperactivity disorder in the multiple-choice query and the matching open ended query. Green checkmarks mark correct outputs. \textbf{d}, Interictal epileptiform discharge detection on IEDS. State detection with dataset specific options Interictal and Interictal pathological activity in the multiple-choice query and an open ended counterpart. All panels follow the same presentation: one signal, two query formats, five models, green checkmarks for correct answers and no marks for incorrect ones.}\label{fig:S18}
\end{minipage}\par
\clearpage
\noindent\begin{minipage}{\textwidth}\centering\includegraphics[width=176mm]{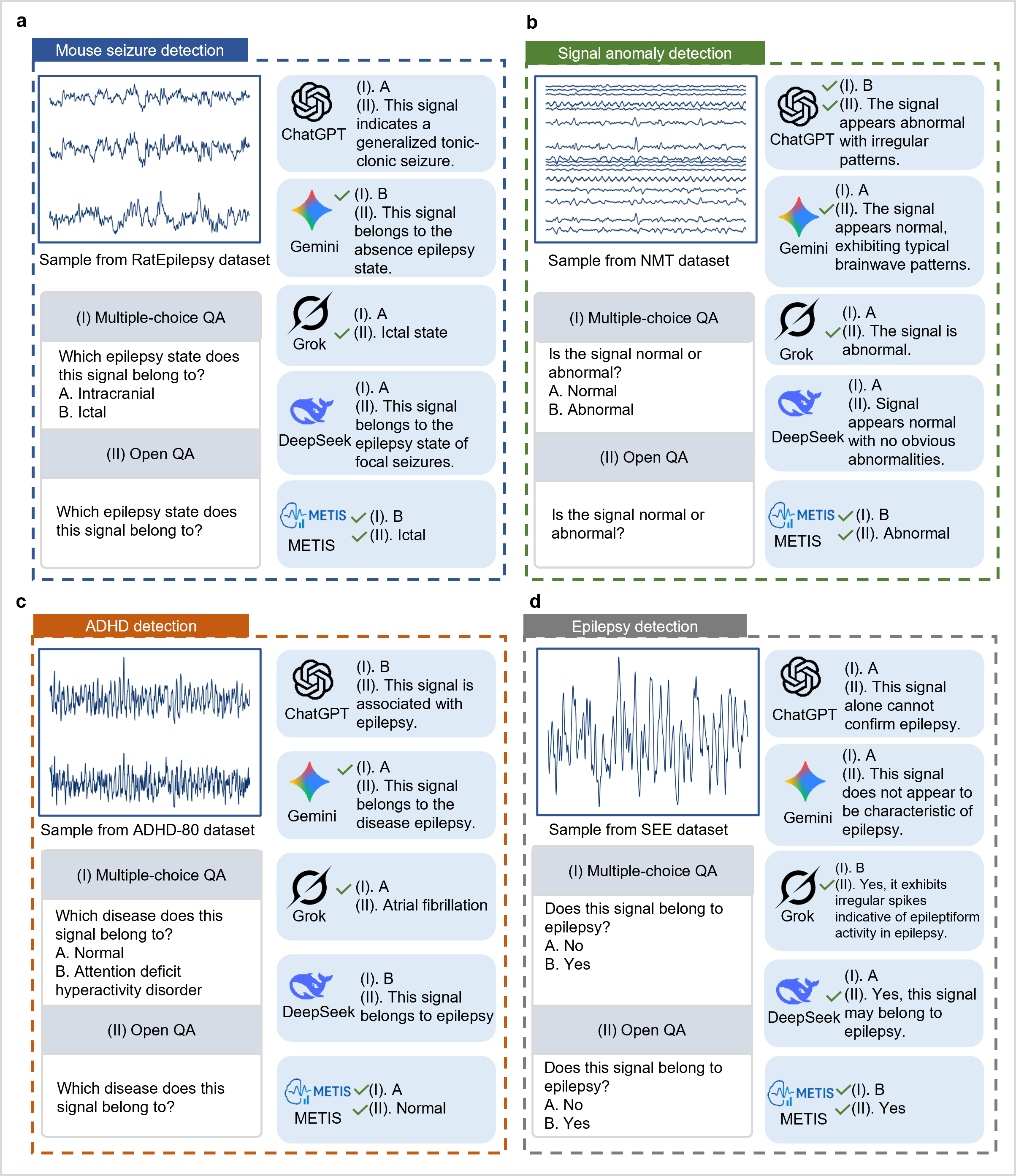}\par
\captionof{figure}{\textbf{Additional zero shot task examples on RatEpilepsy, NMT, ADHD-80, and SEE.} \textbf{a}, Mouse seizure detection on RatEpilepsy. Ictal versus non-ictal state detection shown with a multiple-choice query that lists the dataset options and an open ended query that expects a canonical state label. Outputs are reported for METIS, ChatGPT, Gemini, Grok, and DeepSeek. Correct answers carry a green checkmark. \textbf{b}, Signal anomaly detection on NMT. Binary abnormality screening with Normal and Abnormal as the candidate set in the multiple-choice query and an aligned open ended query. Green checkmarks indicate correct predictions. \textbf{c}, ADHD detection on ADHD-80. Binary disease detection with options Normal and Attention deficit hyperactivity disorder under the multiple-choice format, plus an open ended query that expects the same label space. Correct answers are marked with green checkmarks. \textbf{d}, Epilepsy detection on SEE. Binary epilepsy screening with options No and Yes for the multiple-choice query and the matched open ended query. Across all panels a single signal is queried in both formats, the five models are evaluated under identical prompts, correct responses are highlighted with a green checkmark, and unmarked responses are incorrect.}\label{fig:S19}
\end{minipage}\par
\clearpage
\noindent\begin{minipage}{\textwidth}\centering\includegraphics[width=174mm]{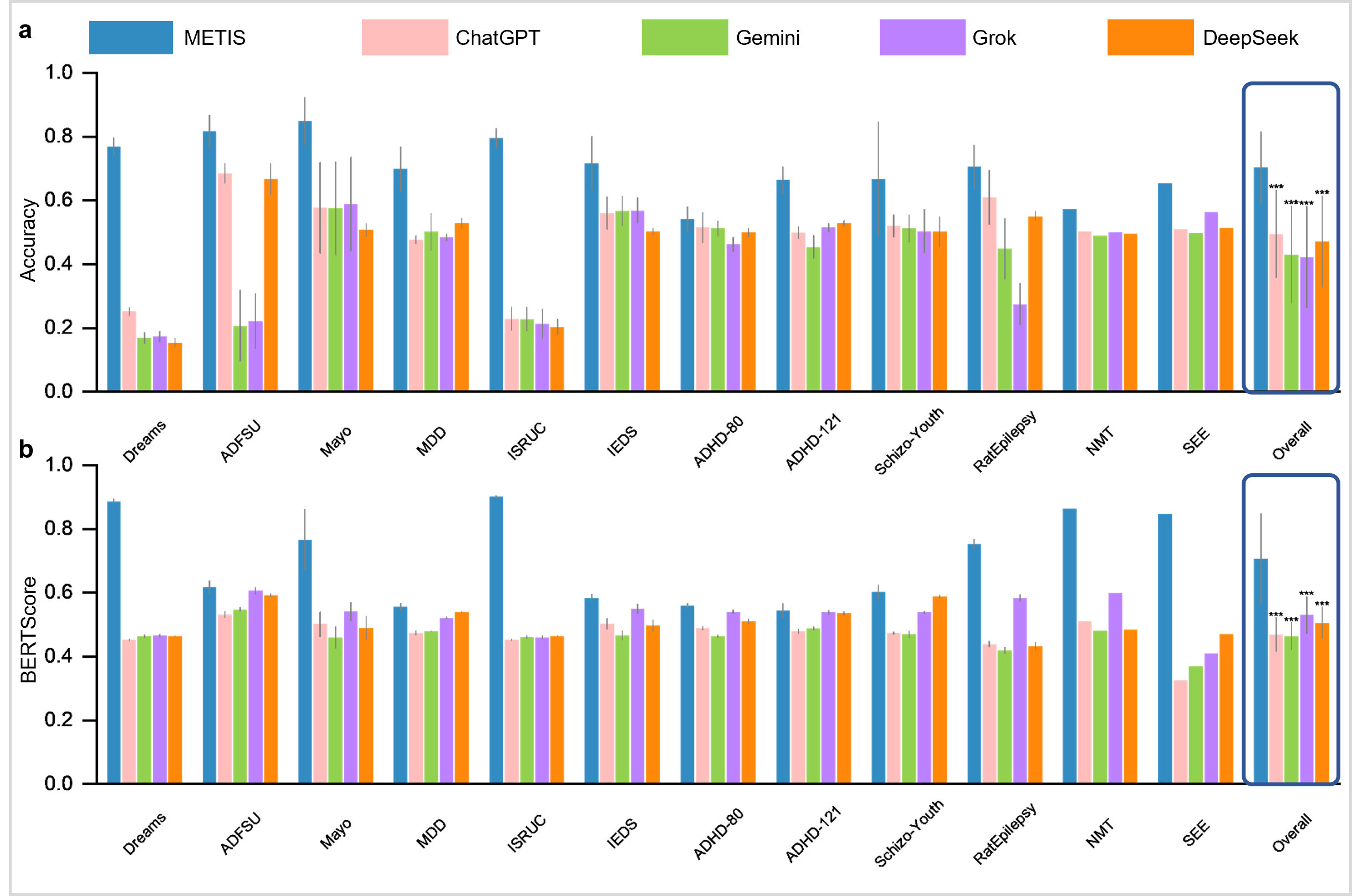}\par
\captionof{figure}{\textbf{Zero-shot performance comparison between METIS and generalist multimodal models across two QA protocols.} Panel \textbf{a} presents accuracy scores for multiple-choice question answering tasks, where the model selects the most plausible answer among a limited set of options. Panel \textbf{b} shows BERTScore for open-ended question answering, evaluating the semantic similarity between generated answers and ground-truth references. Each bar represents the average performance across repeated runs on 12 diverse brain signal datasets. The rightmost group in each panel reports the overall mean across all datasets, using the same averaging method. Significance markers (*, **, ***) indicate the difference between METIS and the best-performing generalist baseline, based on two-sided t-tests (*p \textless{} 0.05, **p \textless{} 0.01, ***p \textless{} 0.001). Across both protocols, METIS consistently achieves superior zero-shot performance compared to state-of-the-art generalist models such as ChatGPT, Gemini, Grok, and DeepSeek.}\label{fig:S20}
\end{minipage}\par
\Needspace{7\baselineskip}\section*{Model Configuration and Architectural Details}
METIS is implemented with a hidden dimension of 512 across all transformer layers. Query-key normalization is applied in the attention mechanism to improve training stability. For standard feed-forward layers, the intermediate dimensionality is set to 2048. For mixture-of-experts (MoE) layers, each routed expert adopts an intermediate dimension of 512, and the shared expert adopts an intermediate dimension of 1024. The backbone consists of 12 transformer layers in total. To ensure stable optimization during early training, the first two layers are implemented as standard transformer layers without expert routing. The remaining layers are configured as MoE layers. Each MoE layer contains 8 routed experts and 1 shared expert, following a top-k routing strategy with \emph{k = 2}. Multi-head attention is employed with 8 query heads, while key and value representations are shared across 2 heads to reduce computational overhead. All other architectural components follow standard transformer implementations unless otherwise specified.\par
\smallskip
\Needspace{7\baselineskip}\section*{MoE Contribution Under Matched Model Scale}
\begingroup
\fontsize{9}{11}\selectfont
\setlength{\tabcolsep}{3pt}\renewcommand{\arraystretch}{1.18}\setlength{\LTpre}{6pt}\setlength{\LTpost}{8pt}
\begin{longtable}{@{}>{\raggedright\arraybackslash}p{31.416mm}>{\raggedright\arraybackslash}p{11.574mm}>{\raggedright\arraybackslash}p{13.228mm}>{\raggedright\arraybackslash}p{28.109mm}>{\raggedright\arraybackslash}p{21.495mm}>{\raggedright\arraybackslash}p{29.763mm}>{\raggedright\arraybackslash}p{29.763mm}@{}}
\caption{Model scale comparison between METIS-MoE and METIS-Dense.}\label{tab:S4}\\
\toprule
\textbf{Model} & \textbf{MoE} & \textbf{Layers} & \textbf{Attention heads} & \textbf{Hidden dim} & \textbf{Total params} & \textbf{Active params}\\\midrule
\endfirsthead
\multicolumn{7}{@{}l}{\fontsize{8}{10}\selectfont\textbf{Table S4.} Continued.}\\\toprule
\textbf{Model} & \textbf{MoE} & \textbf{Layers} & \textbf{Attention heads} & \textbf{Hidden dim} & \textbf{Total params} & \textbf{Active params}\\\midrule
\endhead
\midrule\multicolumn{7}{r@{}}{\fontsize{8}{10}\selectfont Continued on next page}\\\endfoot
\bottomrule\endlastfoot
METIS-MoE & Yes & 12 & 8 & 512 & 170.70M & 124.21M\\
METIS-Dense & No & 12 & 10 & 640 & 168.70M & 168.70M\\
\end{longtable}\endgroup
\begingroup
\fontsize{9}{11}\selectfont
\setlength{\tabcolsep}{3pt}\renewcommand{\arraystretch}{1.18}\setlength{\LTpre}{6pt}\setlength{\LTpost}{8pt}
\begin{longtable}{@{}>{\raggedright\arraybackslash}p{42.918mm}>{\raggedright\arraybackslash}p{46.352mm}>{\raggedright\arraybackslash}p{46.352mm}>{\raggedright\arraybackslash}p{36.051mm}@{}}
\caption{Parameter-matched dense Transformer baseline under zero-shot evaluation. Gain = METIS-MoE \ensuremath{-} METIS-Dense, reported in percentage points.}\label{tab:S5}\\
\toprule
\textbf{Dataset} & \textbf{METIS-Dense AUROC} & \textbf{METIS-MoE AUROC} & \textbf{Gain (pp)}\\\midrule
\endfirsthead
\multicolumn{4}{@{}l}{\fontsize{8}{10}\selectfont\textbf{Table S5.} Continued.}\\\toprule
\textbf{Dataset} & \textbf{METIS-Dense AUROC} & \textbf{METIS-MoE AUROC} & \textbf{Gain (pp)}\\\midrule
\endhead
\midrule\multicolumn{4}{r@{}}{\fontsize{8}{10}\selectfont Continued on next page}\\\endfoot
\bottomrule\endlastfoot
ISRUC & 0.9525 & 0.9411 & -1.14\\
Dreams & 0.9404 & 0.9289 & -1.15\\
Mayo & 0.8970 & 0.9352 & +3.82\\
IEDS & 0.6308 & 0.6419 & +1.11\\
SEE & 0.6518 & 0.7145 & +6.27\\
RatEpilepsy & 0.5270 & 0.6319 & +10.49\\
ADHD-80 & 0.4000 & 0.6707 & +27.07\\
ADHD-121 & 0.6365 & 0.7240 & +8.75\\
MDD & 0.4833 & 0.7425 & +25.92\\
Schizo-Youth & 0.6302 & 0.7208 & +9.06\\
ADFSU & 0.6958 & 0.7458 & +5.00\\
NMT & 0.7153 & 0.6620 & -5.33\\
Mean & 0.6801 & 0.7550 & +7.49\\
\end{longtable}\endgroup
To examine whether the gain of METIS comes from the MoE architecture rather than total model scale, we constructed a parameter-matched dense Transformer baseline, denoted METIS-Dense. As shown in Table~\ref{tab:S4}, METIS-Dense replaces the sparse expert feed-forward layers with standard dense feed-forward layers, while maintaining a comparable total parameter count to METIS-MoE.\par
\smallskip
METIS-Dense follows the same multimodal signal-language formulation and the same zero-shot evaluation protocol as METIS-MoE. It also uses the same pretraining data, objectives, optimizer, optimization settings, batch size, and training steps. Specifically, METIS-Dense contains 12 Transformer layers with a hidden dimension of 640 and 168.70M parameters, and its feed-forward intermediate dimension is 2560. METIS-MoE contains 12 Transformer layers with a hidden dimension of 512 and 170.70M total parameters. In METIS-MoE, the first two layers are standard Transformer layers, and the remaining ten layers are MoE layers. Each MoE layer contains eight routed experts and one shared expert, with top-k routing using k = 2. Each routed expert has an intermediate dimension of 512, and the shared expert has an intermediate dimension of 1024. METIS-MoE has 124.21M active parameters, corresponding to 73.6\% of the active parameters of METIS-Dense. We report active parameters as an input-independent measure, since approximate FLOPs depend on the input length and the number of signal tokens across datasets.\par
\smallskip
As shown in Table~\ref{tab:S5}, METIS-MoE achieved a mean zero-shot AUROC of 0.7550 across 12 datasets, compared with 0.6801 for METIS-Dense, yielding an average gain of 7.49 percentage points. METIS-MoE outperformed METIS-Dense on 9 of the 12 datasets, with especially large gains on ADHD-121, MDD, RatEpilepsy, Schizo-Youth, ADHD-80, SEE, ADFSU, and Mayo. These results suggest that sparse expert routing is beneficial for heterogeneous clinical brain-signal tasks, where diverse signal patterns and task semantics must be handled within a unified model.\par
\smallskip
METIS-Dense performed slightly better on ISRUC, Dreams, and NMT, indicating that dense feed-forward capacity can remain competitive for some structured tasks. Nevertheless, the overall zero-shot benchmark supports that the MoE architecture contributes to METIS's generalization ability beyond merely increasing total parameter count.\par
\smallskip
\Needspace{7\baselineskip}\section*{Dataset Usage and Overlap Control}
\begingroup
\fontsize{8.5}{10.5}\selectfont
\setlength{\tabcolsep}{3pt}\renewcommand{\arraystretch}{1.18}\setlength{\LTpre}{6pt}\setlength{\LTpost}{8pt}
\begin{longtable}{@{}>{\raggedright\arraybackslash}p{36.051mm}>{\raggedright\arraybackslash}p{22.318mm}>{\raggedright\arraybackslash}p{41.202mm}>{\raggedright\arraybackslash}p{72.103mm}@{}}
\caption{Dataset usage for pretraining and downstream evaluation.}\label{tab:S6}\\
\toprule
\textbf{Dataset} & \textbf{Modality} & \textbf{Usage} & \textbf{Downstream setting}\\\midrule
\endfirsthead
\multicolumn{4}{@{}l}{\fontsize{8}{10}\selectfont\textbf{Table S6.} Continued.}\\\toprule
\textbf{Dataset} & \textbf{Modality} & \textbf{Usage} & \textbf{Downstream setting}\\\midrule
\endhead
\midrule\multicolumn{4}{r@{}}{\fontsize{8}{10}\selectfont Continued on next page}\\\endfoot
\bottomrule\endlastfoot
HUP & iEEG & Pretraining & Not applicable\\
UPenn & iEEG & Pretraining & Not applicable\\
SWEC\_ETHZ & iEEG & Pretraining & Not applicable\\
FNUSA & iEEG & Pretraining & Not applicable\\
SHHS & EEG & Pretraining & Not applicable\\
SeizeIT2 & EEG & Pretraining & Not applicable\\
TUSZ & EEG & Pretraining & Not applicable\\
TUAB & EEG & Pretraining & Not applicable\\
CHB-MIT & EEG & Pretraining & Not applicable\\
TUEP & EEG & Pretraining & Not applicable\\
SleepEDF & EEG & Pretraining & Not applicable\\
HaaglandenSleep & EEG & Pretraining & Not applicable\\
TDBrain & EEG & Pretraining & Not applicable\\
TUEV & EEG & Pretraining & Not applicable\\
ADFTD & EEG & Pretraining & Not applicable\\
BrainLat & EEG & Pretraining & Not applicable\\
AD-Auditory & EEG & Pretraining & Not applicable\\
ShuMI & EEG & Pretraining & Not applicable\\
REEG-PD & EEG & Pretraining & Not applicable\\
PhysioNetMI & EEG & Pretraining & Not applicable\\
ISRUC & EEG & Downstream evaluation & Zero-shot; few-shot; transfer\\
Dreams & EEG & Downstream evaluation & Zero-shot; few-shot; transfer\\
SEE & EEG & Downstream evaluation & Zero-shot\\
Siena & EEG & Downstream evaluation & Few-shot\\
RatEpilepsy & iEEG & Downstream evaluation & Zero-shot; few-shot\\
IEDS & iEEG & Downstream evaluation & Zero-shot; few-shot; transfer\\
Mayo & iEEG & Downstream evaluation & Zero-shot; few-shot; transfer\\
ADFSU & EEG & Downstream evaluation & Zero-shot; few-shot; transfer\\
APAVA & EEG & Downstream evaluation & Transfer\\
SanDiego & EEG & Downstream evaluation & Few-shot\\
ADHD-121 & EEG & Downstream evaluation & Zero-shot; few-shot; transfer\\
ADHD-80 & EEG & Downstream evaluation & Zero-shot; few-shot; transfer\\
MDD & EEG & Downstream evaluation & Zero-shot\\
Schizo-Youth & EEG & Downstream evaluation & Zero-shot; few-shot\\
Schizo-28 & EEG & Downstream evaluation & Few-shot\\
NMT & EEG & Downstream evaluation & Zero-shot; few-shot\\
NTUHBIS & EEG & Downstream evaluation & Few-shot\\
\end{longtable}\endgroup
To clarify dataset usage, Table~\ref{tab:S6} maps each dataset to its role in the study. The pretraining corpus and downstream evaluation datasets are disjoint. We further confirmed that there was no subject-level overlap or inadvertent reuse of recordings between pretraining and downstream evaluation. For downstream evaluations, no subjects or recordings were shared across the training, validation, and test partitions. For datasets with official splits, we strictly followed the official partition protocols; for datasets without official splits, subject-wise partitioning was used to prevent subject-level leakage across folds.\par
\smallskip
\Needspace{7\baselineskip}\section*{Effect of Pretraining Instruction Quality}
To examine the impact of language-instruction quality during pretraining, we conducted a weak-instruction ablation. In this experiment, the model architecture, pretraining datasets, training protocol, and zero-shot evaluation protocol were kept unchanged. The only modification was that the original task-specific natural-language instructions were replaced with a generic weak instruction, ``Classify this signal.'' The label options were preserved. Therefore, this ablation weakens the task-level linguistic context while keeping the answer space available.\par
\smallskip
Table~\ref{tab:S7} shows the zero-shot performance of the full-instruction model and the weak-instruction model across the 12 datasets used in the main zero-shot benchmark. The full-instruction model achieved a mean AUROC of 0.7550, whereas the weak-instruction model achieved 0.6600, corresponding to an average gain of 9.50 percentage points. The full-instruction model outperformed the weak-instruction model on 11 of the 12 datasets.\par
\smallskip
The effect of instruction quality was task-dependent. The performance gap was relatively small on ISRUC, Dreams, Mayo, and IEDS, which are closely related to high-frequency pretraining task types such as sleep-stage classification and epilepsy-related classification. In these settings, the weak-instruction model can still partly rely on strong signal-label associations and the provided label options. In contrast, the degradation was much larger on clinically heterogeneous or semantically subtle tasks, including MDD, RatEpilepsy, ADFSU, Schizo-Youth, ADHD-121, ADHD-80, SEE, and NMT. This pattern suggests that high-quality task-specific instructions are especially important in heterogeneous multi-task pretraining, where they help condition the model on the appropriate semantic decision space rather than relying mainly on dominant task distributions or label-option associations.\par
\smallskip
These results indicate that the language instructions used during pretraining are not merely superficial prompt templates. Instead, they contribute to task-aware signal-language alignment and improve zero-shot generalization across diverse clinical brain-signal tasks.\par
\smallskip
\begingroup
\fontsize{9}{11}\selectfont
\setlength{\tabcolsep}{3pt}\renewcommand{\arraystretch}{1.18}\setlength{\LTpre}{6pt}\setlength{\LTpost}{8pt}
\begin{longtable}{@{}>{\raggedright\arraybackslash}p{46.352mm}>{\raggedright\arraybackslash}p{44.635mm}>{\raggedright\arraybackslash}p{44.635mm}>{\raggedright\arraybackslash}p{36.051mm}@{}}
\caption{Weak-instruction ablation under zero-shot evaluation. Gain = Full instruction \ensuremath{-} Weak instruction, reported in percentage points.}\label{tab:S7}\\
\toprule
\textbf{Dataset} & \textbf{Full instruction} & \textbf{Weak instruction} & \textbf{Gain (pp)}\\\midrule
\endfirsthead
\multicolumn{4}{@{}l}{\fontsize{8}{10}\selectfont\textbf{Table S7.} Continued.}\\\toprule
\textbf{Dataset} & \textbf{Full instruction} & \textbf{Weak instruction} & \textbf{Gain (pp)}\\\midrule
\endhead
\midrule\multicolumn{4}{r@{}}{\fontsize{8}{10}\selectfont Continued on next page}\\\endfoot
\bottomrule\endlastfoot
ISRUC & 0.9411 & 0.9408 & +0.03\\
Mayo & 0.9352 & 0.9389 & -0.37\\
Dreams & 0.9289 & 0.9157 & +1.32\\
ADFSU & 0.7458 & 0.5916 & +15.42\\
MDD & 0.7425 & 0.5482 & +19.43\\
ADHD-121 & 0.7240 & 0.6392 & +8.48\\
Schizo-Youth & 0.7208 & 0.5691 & +15.17\\
SEE & 0.7145 & 0.6500 & +6.45\\
ADHD-80 & 0.6707 & 0.5226 & +14.81\\
NMT & 0.6620 & 0.5334 & +12.86\\
IEDS & 0.6419 & 0.6273 & +1.46\\
RatEpilepsy & 0.6319 & 0.4431 & +18.88\\
Mean & 0.7550 & 0.6600 & +9.50\\
\end{longtable}\endgroup
\Needspace{7\baselineskip}\section*{Prompt and Label Phrasing Robustness}
To examine whether METIS relies on exact prompt templates or label surface forms, we evaluated the original METIS checkpoint under controlled prompt and label phrasing variations. The model parameters, datasets, evaluation protocol, answer space, and ground-truth annotations were kept unchanged. Only the surface form of the task instructions or answer labels was modified.\par
\smallskip
The prompt and label variants were constructed based on the original dataset annotation protocols and standard clinical terminology. For prompt paraphrasing, we rewrote the task questions while preserving the original task intent. For label phrasing, we used clinically equivalent terms, standard abbreviations, or spelling variants where applicable, such as ``Rapid Eye Movement'' and ``REM,'' or ``Alzheimer's disease'' and ``AD.'' Each variant was checked against the original label definitions to ensure that the task meaning, class boundaries, and ground-truth annotations remained unchanged.\par
\smallskip
We evaluated two types of robustness. In the prompt paraphrasing test, the original label set was kept unchanged, while the canonical task instruction was replaced by three semantically equivalent paraphrased prompts. In the label phrasing test, the canonical prompt was kept unchanged, while the answer labels were replaced by three alternate but semantically equivalent label phrasings. Table~\ref{tab:S8} summarizes the mean results, and Tables~\ref{tab:S9} and~\ref{tab:S10} provide the per-variant results. The exact prompt paraphrases and alternate label phrasings used for each dataset are listed in Tables~\ref{tab:S11} and~\ref{tab:S12}.\par
\smallskip
Across the 12 zero-shot datasets, METIS achieved a mean AUROC of 0.7550 under the canonical setting. Under paraphrased prompts, the mean AUROC was 0.7301, corresponding to an average decrease of 2.49 percentage points. Under alternate label phrasings, the mean AUROC was 0.7235, corresponding to an average decrease of 3.15 percentage points. These results indicate that METIS maintains overall stable zero-shot performance under controlled prompt and label surface-form changes.\par
\smallskip
The sensitivity to linguistic variation was task-dependent. Performance remained highly stable on ISRUC, Dreams, MDD, Schizo-Youth, and ADHD-80, while larger drops were observed on Mayo, IEDS, ADHD-121, ADFSU, and NMT. Overall, these results support that METIS is not solely dependent on exact template matching, while also acknowledging that some datasets show substantial sensitivity to prompt or label phrasing.\par
\smallskip
\begingroup
\fontsize{9}{11}\selectfont
\setlength{\tabcolsep}{3pt}\renewcommand{\arraystretch}{1.18}\setlength{\LTpre}{6pt}\setlength{\LTpost}{8pt}
\begin{longtable}{@{}>{\raggedright\arraybackslash}p{31.817mm}>{\raggedright\arraybackslash}p{23.444mm}>{\raggedright\arraybackslash}p{30.142mm}>{\raggedright\arraybackslash}p{26.793mm}>{\raggedright\arraybackslash}p{28.468mm}>{\raggedright\arraybackslash}p{26.793mm}@{}}
\caption{Summary of prompt and label phrasing robustness under zero-shot evaluation. Change is computed relative to the canonical setting and reported in percentage points.}\label{tab:S8}\\
\toprule
\textbf{Dataset} & \textbf{Canonical AUROC} & \textbf{Prompt mean AUROC} & \textbf{Prompt change (pp)} & \textbf{Label mean AUROC} & \textbf{Label change (pp)}\\\midrule
\endfirsthead
\multicolumn{6}{@{}l}{\fontsize{8}{10}\selectfont\textbf{Table S8.} Continued.}\\\toprule
\textbf{Dataset} & \textbf{Canonical AUROC} & \textbf{Prompt mean AUROC} & \textbf{Prompt change (pp)} & \textbf{Label mean AUROC} & \textbf{Label change (pp)}\\\midrule
\endhead
\midrule\multicolumn{6}{r@{}}{\fontsize{8}{10}\selectfont Continued on next page}\\\endfoot
\bottomrule\endlastfoot
ISRUC & 0.9411 & 0.9365 & -0.46 & 0.9387 & -0.24\\
Dreams & 0.9289 & 0.9274 & -0.15 & 0.9285 & -0.04\\
Mayo & 0.9352 & 0.8645 & -7.07 & 0.8541 & -8.11\\
IEDS & 0.6419 & 0.5631 & -7.88 & 0.5583 & -8.36\\
SEE & 0.7145 & 0.7210 & +0.65 & 0.6771 & -3.74\\
RatEpilepsy & 0.6319 & 0.6254 & -0.65 & 0.6034 & -2.85\\
ADHD-80 & 0.6707 & 0.7205 & +4.98 & 0.7072 & +3.65\\
ADHD-121 & 0.7240 & 0.6449 & -7.91 & 0.6153 & -10.87\\
MDD & 0.7425 & 0.7138 & -2.87 & 0.7418 & -0.07\\
Schizo-Youth & 0.7208 & 0.7242 & +0.34 & 0.7637 & +4.29\\
ADFSU & 0.7458 & 0.7590 & +1.32 & 0.6845 & -6.13\\
NMT & 0.6620 & 0.5611 & -10.09 & 0.6098 & -5.22\\
Mean & 0.7550 & 0.7301 & -2.49 & 0.7235 & -3.15\\
\end{longtable}\endgroup
\begingroup
\fontsize{9}{11}\selectfont
\setlength{\tabcolsep}{3pt}\renewcommand{\arraystretch}{1.18}\setlength{\LTpre}{6pt}\setlength{\LTpost}{8pt}
\begin{longtable}{@{}>{\raggedright\arraybackslash}p{31.416mm}>{\raggedright\arraybackslash}p{21.495mm}>{\raggedright\arraybackslash}p{21.495mm}>{\raggedright\arraybackslash}p{21.495mm}>{\raggedright\arraybackslash}p{21.495mm}>{\raggedright\arraybackslash}p{24.802mm}>{\raggedright\arraybackslash}p{23.149mm}@{}}
\caption{Prompt paraphrasing robustness under zero-shot evaluation. Prompt mean denotes the average AUROC across three semantically equivalent paraphrased prompts. Max drop denotes the largest decrease from the canonical prompt among the three paraphrased prompts.}\label{tab:S9}\\
\toprule
\textbf{Dataset} & \textbf{Canonical} & \textbf{Prompt 1} & \textbf{Prompt 2} & \textbf{Prompt 3} & \textbf{Prompt mean} & \textbf{Max drop (pp)}\\\midrule
\endfirsthead
\multicolumn{7}{@{}l}{\fontsize{8}{10}\selectfont\textbf{Table S9.} Continued.}\\\toprule
\textbf{Dataset} & \textbf{Canonical} & \textbf{Prompt 1} & \textbf{Prompt 2} & \textbf{Prompt 3} & \textbf{Prompt mean} & \textbf{Max drop (pp)}\\\midrule
\endhead
\midrule\multicolumn{7}{r@{}}{\fontsize{8}{10}\selectfont Continued on next page}\\\endfoot
\bottomrule\endlastfoot
ISRUC & 0.9411 & 0.9358 & 0.9373 & 0.9365 & 0.9365 & 0.53\\
Dreams & 0.9289 & 0.9274 & 0.9277 & 0.9270 & 0.9274 & 0.19\\
Mayo & 0.9352 & 0.8696 & 0.8329 & 0.8910 & 0.8645 & 10.23\\
IEDS & 0.6419 & 0.5686 & 0.5455 & 0.5752 & 0.5631 & 9.64\\
SEE & 0.7145 & 0.7438 & 0.7377 & 0.6814 & 0.7210 & 3.31\\
RatEpilepsy & 0.6319 & 0.6403 & 0.5401 & 0.6958 & 0.6254 & 9.18\\
ADHD-80 & 0.6707 & 0.7213 & 0.7187 & 0.7216 & 0.7205 & 0.00\\
ADHD-121 & 0.7240 & 0.6460 & 0.6674 & 0.6211 & 0.6449 & 10.29\\
MDD & 0.7425 & 0.7078 & 0.7097 & 0.7239 & 0.7138 & 3.47\\
Schizo-Youth & 0.7208 & 0.7188 & 0.7230 & 0.7307 & 0.7242 & 0.20\\
ADFSU & 0.7458 & 0.7952 & 0.7404 & 0.7415 & 0.7590 & 0.54\\
NMT & 0.6620 & 0.5744 & 0.5606 & 0.5483 & 0.5611 & 11.37\\
\end{longtable}\endgroup
\begingroup
\fontsize{9}{11}\selectfont
\setlength{\tabcolsep}{3pt}\renewcommand{\arraystretch}{1.18}\setlength{\LTpre}{6pt}\setlength{\LTpost}{8pt}
\begin{longtable}{@{}>{\raggedright\arraybackslash}p{31.416mm}>{\raggedright\arraybackslash}p{21.495mm}>{\raggedright\arraybackslash}p{21.495mm}>{\raggedright\arraybackslash}p{21.495mm}>{\raggedright\arraybackslash}p{21.495mm}>{\raggedright\arraybackslash}p{24.802mm}>{\raggedright\arraybackslash}p{23.149mm}@{}}
\caption{Label phrasing robustness under zero-shot evaluation. Label mean denotes the average AUROC across three semantically equivalent alternate label phrasings. Max drop denotes the largest decrease from the canonical label phrasing among the three label variants.}\label{tab:S10}\\
\toprule
\textbf{Dataset} & \textbf{Canonical} & \textbf{Label 1} & \textbf{Label 2} & \textbf{Label 3} & \textbf{Label mean} & \textbf{Max drop (pp)}\\\midrule
\endfirsthead
\multicolumn{7}{@{}l}{\fontsize{8}{10}\selectfont\textbf{Table S10.} Continued.}\\\toprule
\textbf{Dataset} & \textbf{Canonical} & \textbf{Label 1} & \textbf{Label 2} & \textbf{Label 3} & \textbf{Label mean} & \textbf{Max drop (pp)}\\\midrule
\endhead
\midrule\multicolumn{7}{r@{}}{\fontsize{8}{10}\selectfont Continued on next page}\\\endfoot
\bottomrule\endlastfoot
ISRUC & 0.9411 & 0.9394 & 0.9382 & 0.9387 & 0.9387 & 0.30\\
Dreams & 0.9289 & 0.9280 & 0.9276 & 0.9299 & 0.9285 & 0.13\\
Mayo & 0.9352 & 0.8831 & 0.8310 & 0.8484 & 0.8541 & 10.42\\
IEDS & 0.6419 & 0.5681 & 0.5532 & 0.5536 & 0.5583 & 8.87\\
SEE & 0.7145 & 0.6715 & 0.6787 & 0.6811 & 0.6771 & 4.30\\
RatEpilepsy & 0.6319 & 0.6389 & 0.6000 & 0.5714 & 0.6034 & 6.05\\
ADHD-80 & 0.6707 & 0.6985 & 0.7270 & 0.6962 & 0.7072 & 0.00\\
ADHD-121 & 0.7240 & 0.6980 & 0.5936 & 0.5543 & 0.6153 & 16.97\\
MDD & 0.7425 & 0.7271 & 0.7442 & 0.7542 & 0.7418 & 1.54\\
Schizo-Youth & 0.7208 & 0.7647 & 0.7526 & 0.7738 & 0.7637 & 0.00\\
ADFSU & 0.7458 & 0.7263 & 0.6325 & 0.6947 & 0.6845 & 11.33\\
NMT & 0.6620 & 0.6008 & 0.6068 & 0.6218 & 0.6098 & 6.12\\
\end{longtable}\endgroup
\begingroup
\fontsize{8}{10}\selectfont
\setlength{\tabcolsep}{3pt}\renewcommand{\arraystretch}{1.18}\setlength{\LTpre}{6pt}\setlength{\LTpost}{8pt}
\renewcommand{\arraystretch}{1.05}
\begin{longtable}{@{}>{\raggedright\arraybackslash}p{22.043mm}>{\raggedright\arraybackslash}p{36.880mm}>{\raggedright\arraybackslash}p{36.880mm}>{\raggedright\arraybackslash}p{36.880mm}>{\raggedright\arraybackslash}p{36.880mm}@{}}
\caption{Exact prompt paraphrases used in the prompt robustness evaluation.}\label{tab:S11}\\
\toprule
\textbf{Dataset} & \textbf{Canonical prompt} & \textbf{Prompt 1} & \textbf{Prompt 2} & \textbf{Prompt 3}\\\midrule
\endfirsthead
\multicolumn{5}{@{}l}{\fontsize{8}{10}\selectfont\textbf{Table S11.} Continued.}\\\toprule
\textbf{Dataset} & \textbf{Canonical prompt} & \textbf{Prompt 1} & \textbf{Prompt 2} & \textbf{Prompt 3}\\\midrule
\endhead
\midrule\multicolumn{5}{r@{}}{\fontsize{8}{10}\selectfont Continued on next page}\\\endfoot
\bottomrule\endlastfoot
ISRUC & Which sleep stage does this signal belong to? & Which sleep stage best describes this signal? & Identify the sleep stage of this signal. & Classify the sleep stage for this recording segment.\\
Dreams & Which sleep stage does this signal belong to? & Which sleep stage best describes this signal? & Identify the sleep stage of this signal. & Classify the sleep stage for this recording segment.\\
Mayo & Which epilepsy state does this signal belong to? & Identify the epilepsy state of this recording segment. & Classify the epilepsy state for the given neural signal. & What epilepsy state best describes this signal?\\
IEDS & Which epilepsy state does this signal belong to? & Identify the epilepsy state of this recording segment. & Classify the epilepsy state for the given neural signal. & What epilepsy state best describes this signal?\\
SEE & Does this signal belong to epilepsy? & Is this EEG segment indicative of epilepsy? & Determine whether this signal shows epileptic activity. & Does this signal show evidence of epilepsy?\\
RatEpilepsy & Which epilepsy state does this signal belong to? & Identify the epilepsy state of this recording segment. & Classify the epilepsy state for the given neural signal. & What epilepsy state best describes this signal?\\
ADHD-121 & What clinical diagnosis is associated with this signal? & Which clinical condition label applies to this signal? & Assign the clinical condition for this recording segment. & Determine the clinical condition indicated by this signal.\\
ADHD-80 & What clinical diagnosis is associated with this signal? & Which clinical condition label applies to this signal? & Assign the clinical condition for this recording segment. & Determine the clinical condition indicated by this signal.\\
MDD & What clinical diagnosis is associated with this signal? & Which clinical condition label applies to this signal? & Assign the clinical condition for this recording segment. & Determine the clinical condition indicated by this signal.\\
Schizo-Youth & What clinical diagnosis is associated with this signal? & Which clinical condition label applies to this signal? & Assign the clinical condition for this recording segment. & Determine the clinical condition indicated by this signal.\\
ADFSU & What clinical diagnosis is associated with this signal? & Which clinical condition label applies to this signal? & Assign the clinical condition for this recording segment. & Determine the clinical condition indicated by this signal.\\
NMT & Is the signal normal or abnormal? & Determine whether this EEG segment is normal or abnormal. & Classify this EEG recording by its clinical status. & Does this signal indicate an abnormal EEG pattern?\\
\end{longtable}\endgroup
\begingroup
\fontsize{8}{10}\selectfont
\setlength{\tabcolsep}{3pt}\renewcommand{\arraystretch}{1.18}\setlength{\LTpre}{6pt}\setlength{\LTpost}{8pt}
\renewcommand{\arraystretch}{1.05}
\begin{longtable}{@{}>{\raggedright\arraybackslash}p{22.043mm}>{\raggedright\arraybackslash}p{36.880mm}>{\raggedright\arraybackslash}p{36.880mm}>{\raggedright\arraybackslash}p{36.880mm}>{\raggedright\arraybackslash}p{36.880mm}@{}}
\caption{Exact alternate label phrasings used in the label robustness evaluation.}\label{tab:S12}\\
\toprule
\textbf{Dataset} & \textbf{Canonical labels} & \textbf{Label phrasing 1} & \textbf{Label phrasing 2} & \textbf{Label phrasing 3}\\\midrule
\endfirsthead
\multicolumn{5}{@{}l}{\fontsize{8}{10}\selectfont\textbf{Table S12.} Continued.}\\\toprule
\textbf{Dataset} & \textbf{Canonical labels} & \textbf{Label phrasing 1} & \textbf{Label phrasing 2} & \textbf{Label phrasing 3}\\\midrule
\endhead
\midrule\multicolumn{5}{r@{}}{\fontsize{8}{10}\selectfont Continued on next page}\\\endfoot
\bottomrule\endlastfoot
ISRUC & Wake; Non-REM Stage 1; Non-REM Stage 2; Non-REM Stage 3; Rapid Eye Movement & Wakefulness; N1; N2; N3; REM sleep & Wake stage; N1 sleep; N2 sleep; N3 sleep; Stage REM & Awake; NREM Stage 1; NREM Stage 2; NREM Stage 3; REM\\
Dreams & Wake; Non-REM Stage 1; Non-REM Stage 2; Non-REM Stage 3; Rapid Eye Movement & Wakefulness; N1; N2; N3; REM sleep & Wake stage; N1 sleep; N2 sleep; N3 sleep; Stage REM & Awake; NREM Stage 1; NREM Stage 2; NREM Stage 3; REM\\
Mayo & Intracranial; Intracranial, pathological activity & Intracranial without pathological activity; Intracranial with pathological activity & Non-pathological intracranial segment; Pathological intracranial segment & Intracranial non-pathological segment; Intracranial pathological segment\\
IEDS & Intracranial; Intracranial, pathological activity & Intracranial without pathological activity; Intracranial with pathological activity & Non-pathological intracranial segment; Pathological intracranial segment & Intracranial non-pathological segment; Intracranial pathological segment\\
SEE & No; Yes & No seizure; Seizure & Non-seizure; Seizure event & Without seizure; With seizure\\
RatEpilepsy & Intracranial; Ictal; Preictal; Postictal & Interictal; Ictal state; Preictal state; Postictal state & Non-ictal intracranial state; Seizure state; Pre-seizure state; Post-seizure state & Between-seizure state; Seizure period; Before-seizure state; After-seizure state\\
ADHD-121 & Normal; Attention deficit hyperactivity disorder & Healthy control; ADHD & Clinically normal control; Attention-deficit/hyperactivity disorder & Normal control; Attention-deficit hyperactivity disorder\\
ADHD-80 & Normal; Attention deficit hyperactivity disorder & Healthy control; ADHD & Clinically normal control; Attention-deficit/hyperactivity disorder & Normal control; Attention-deficit hyperactivity disorder\\
MDD & Normal; Major depressive disorder & Healthy control; MDD & Clinically normal control; Major depression & Normal control; Depressive disorder\\
Schizo-Youth & Normal; Schizophrenia & Healthy control; Schizophrenia diagnosis & Clinically normal control; Schizophrenia spectrum disorder & Normal control; Schizophrenic disorder\\
ADFSU & Normal; Alzheimer's disease & Healthy control; Alzheimer disease & Cognitively normal; AD dementia & Normal control; Alzheimer-type dementia\\
NMT & Normal; Abnormal & Clinically normal; Clinically abnormal & Normal recording; Abnormal recording & Normal signal; Abnormal signal\\
\end{longtable}\endgroup

\end{document}